%% file: SELMA_SEmantic_Largescale_Multimodal_Acquisitions_in_Variable_Weather_Daytime_and_Viewpoints.tex
\documentclass[journal]{IEEEtran}
\usepackage{amsmath,amsfonts,mathtools}
\usepackage{pifont}
\newcommand{\cmark}{\ding{51}}%
\newcommand{\xmark}{\ding{55}}%
\usepackage{algorithmic}
\usepackage{algorithm}
\usepackage{array}
\usepackage{textcomp}
\usepackage{stfloats}
\usepackage{url}
\usepackage{verbatim}
\usepackage{graphicx}
\usepackage{cite}
\usepackage{glossaries}
\usepackage{multirow}
\usepackage{makecell}
\usepackage{color, colortbl}
\definecolor{mixed}{gray}{0.9}
\definecolor{synth}{gray}{0.8}

\usepackage[tableposition=top]{caption}
\usepackage{subcaption}

\usepackage{colortbl}
\usepackage{pgfplots}
\usepackage{pgfplotstable}
\usepackage{longtable}
\usepackage{xspace}
\usepackage{arydshln}

\usepackage[font=scriptsize]{subcaption}
\usepackage[font=footnotesize]{caption}

\usepackage{tikz}
\pgfplotsset{compat=newest}
\pgfplotsset{plot coordinates/math parser=false}
\newlength\fheight
\newlength\fwidth
\usetikzlibrary{plotmarks,shapes,patterns,decorations.pathreplacing,backgrounds,calc,arrows,arrows.meta,spy,matrix}
\usepgfplotslibrary{patchplots,groupplots,}
\usepackage{tikzscale}

\usepackage[capitalise]{cleveref}
\crefname{section}{Sec.}{Secs.}
\crefname{figure}{Fig.}{Figs.}

\DeclareCaptionLabelFormat{andtable}{#1~#2  \&  \tablename~\thetable}

\IEEEoverridecommandlockouts

\input{./acronymsSorted.tex}

\begin{document}

\title{\fontsize{22}{22}\selectfont SELMA: SEmantic Large-scale Multimodal Acquisitions\\in Variable Weather, Daytime and Viewpoints}

\author{Paolo~Testolina*,~\IEEEmembership{Student~Member,~IEEE}, Francesco~Barbato*,~\IEEEmembership{Student~Member,~IEEE}, Umberto~Michieli,~\IEEEmembership{Graduate~Student~Member,~IEEE}, Marco~Giordani,~\IEEEmembership{Member,~IEEE}, Pietro~Zanuttigh,~\IEEEmembership{Member,~IEEE}, Michele~Zorzi,~\IEEEmembership{Fellow,~IEEE}
\thanks{*: P. Testolina and F. Barbato are primary co-authors.}
\thanks{All authors are with the University of Padova. The work of F.\ Barbato, U.\ Michieli and P.\ Zanuttigh was in part supported by the SID 2020 project ``Semantic Segmentation in the Wild.'' 
The work of P.\ Testolina was supported by Fondazione CaRiPaRo under grants ``Dottorati di Ricerca'' 2019.}
\thanks{Manuscript received xxx, 2022; revised xxx, 2022.}}

\markboth{Journal of \LaTeX\ Class Files,~Vol.~14, No.~8, August~2022}%
{Shell \MakeLowercase{\textit{et al.}}: A Sample Article Using IEEEtran.cls for IEEE Journals}


\maketitle

\begin{abstract}
Accurate scene understanding from multiple sensors mounted on cars is a key requirement for autonomous driving systems. Nowadays, this task is mainly performed through data-hungry deep learning techniques that need very large amounts of data to be trained. Due to the high cost of performing segmentation labeling, many synthetic datasets have been proposed. However, most of them miss the multi-sensor nature of the data, and do not capture the significant changes introduced by the variation of daytime and weather conditions. To fill these gaps, we introduce SELMA, a novel synthetic dataset for semantic segmentation that contains more than 30K unique waypoints acquired from 24 different sensors including RGB, depth, semantic cameras and LiDARs, in 27 different weather and daytime conditions, for a total of more than 20M samples.
SELMA is based on CARLA, an open-source simulator for generating synthetic data in autonomous driving scenarios, that we modified to increase the variability and the diversity in the scenes and class sets, and to align it with other benchmark datasets.
As shown by the experimental evaluation, SELMA allows the efficient training of standard and multi-modal deep learning architectures, and achieves remarkable results on real-world data. SELMA is free and publicly available, thus supporting open science and research.
\end{abstract}


\begin{tikzpicture}[remember picture,overlay]
\node[anchor=north,yshift=-10pt] at (current page.north) {\parbox{\dimexpr\textwidth-\fboxsep-\fboxrule\relax}{
\centering\footnotesize This paper has been submitted to IEEE. Copyright may change without notice. 
}};
\end{tikzpicture}

\section{Introduction}
\label{sec:intro}
Recent advances in the automotive sector have paved the way toward \glspl{c-its} to achieve safer and more efficient driving. 
Not only can \glspl{c-its} reduce the number of traffic accidents (up to $90\%$, according to some estimates~\cite{fagnant2015preparing}) or improve traffic management via smart  platooning, cruise control and/or traffic light coordination, but it holds the promise to improve fuel economy and contribute to a $60\%$ fall in carbon emissions~\cite{silberg2012self}. Overall, \glspl{c-its} represent a huge market of more than 7 trillion USD~\cite{clements2017economic},  hence stimulating significant research efforts.

To these goals, future connected vehicles will be equipped with heterogeneous sensors, including \gls{lidar} and RGB camera sensors, able to provide an accurate perception of the environment.
In particular, \glspl{lidar} generate a 3D omnidirectional representation of the environment in the form of a point cloud, and stand out as the most accurate sensors for geometry acquisition under several weather and lighting conditions~\cite{li2020lidar}.
On the other side, RGB cameras offer advantages like cheaper price, higher resolution 
and higher frame rate than LiDARs, even though they suffer from severe sensitivity to illumination and visibility conditions~\cite{secci2020failures}.
In this sense, sensor fusion appears as a promising solution to provide more robust scene understanding, at the expense of the additional processing overhead for collecting and combining observations from multiple sensors~\cite{rossi2021role}.

However, autonomous driving tasks, in particular \gls{ss} and \gls{v2x} communication, raise several challenges~\cite{giordani2018performance,giordani2018feasibility}, also in view of the complex and dynamic environment in which autonomous vehicles move and operate. 
In these regards, \gls{ml} and \gls{dl} represent valuable tools to address these issues and optimize driving decisions~\cite{zhang2018vehicular}.
However, these techniques require the availability of massive  amounts of labeled data for proper training, whose acquisition and labeling is extremely expensive and time consuming.
Hence, existing open-source datasets, like Waymo~\cite{sun2020scalability}, Cityscapes \cite{Cordts2016},
and KITTI~\cite{geiger2013vision}, are scarce and generally lack diversity. 
Moreover, many datasets are too small to capture the many challenges of the urban scenario, do not encompass multiple (and diverse) sensors, and come with unlabeled scenes, undermining the training of ML models~\cite{mao2021one}.

To fill these gaps, the scientific community has been investigating the usage of synthetic (computer-generated) datasets, where the full control of the data generation pipeline is~delegated to simulations, hence ensuring lower costs, greater flexibility and larger quantity than real-world data~\cite{gaidon2016virtual,toldo2020unsupervised,barbato2021latent,barbato2021road,michieli2020adversarial}. 
Notably, simulations facilitate data acquisition in different conditions and scenarios, and considering diverse sets of sensors.
An open-source simulator to generate synthetic data is \gls{carla}~\cite{Dosovitskiy17}, which includes urban layouts, a wide range of environmental conditions, vehicles, buildings and pedestrians models, and supports a flexible setup of sensors. 
At the time of writing, several synthetic datasets exist for \gls{ss} in autonomous driving~\cite{Richter2016,ros2016synthia,gaidon2016virtual,cabon2020vkitti2,alberti2020idda}, where only~\cite{alberti2020idda} is based on \gls{carla}.
These datasets, however, present limitations. 
In particular, samples are generally captured in a limited number of settings, in similar viewpoints, weather, lighting, and daytime conditions, and often from a single sensor. Moreover, they do not provide end-users with fine-grained control over the weather setup or even the same semantic class set as common benchmarks, like Cityscapes~\cite{Cordts2016}.

To overcome these limitations, in this paper we present \gls{selma}, a new multimodal synthetic dataset for autonomous driving, built using a modified version of the \gls{carla} simulator.
Our dataset stands out as one of the largest, most complete and diverse datasets to ensure adequate design, prototyping, and validation of autonomous driving models, in particular to
solve complex tasks like \gls{ss}.
Specifically, our \gls{selma} dataset consists of:
    \begin{itemize}
        \item Data acquired in 30\,909 independent locations from 7 RGB cameras, 7 depth cameras, 7 semantic cameras, and 3 LiDARs coupled with semantic information.
        The multimodal setup of \gls{selma} promotes complementary and diversity of data, and permits higher accuracy and performance of learning tasks~\cite{lahat2015multimodal}.
        \item Acquisitions generated in variable weather, daytime and viewpoint conditions, and across 8 maps, for a total of 216 unique settings. 
        To this aim, the CARLA simulator has been modified to increase 
        the photo-realism of the weather conditions, and to maximize the visual variability, e.g., by adding parked bikes or traffic lights and~signs.
        \item Semantic labeling for both camera and LiDAR data into $36$ distinct classes, with complete overlap with the training set of common benchmarks like Cityscapes~\cite{Cordts2016}, obtained by modifying the source code of the simulator.
    \end{itemize}

We validate the accuracy and realism of our dataset starting from a set of baseline experiments, and show that \gls{dl} models for semantic understanding trained on our dataset outperform the same models trained on competing synthetic datasets when tested on a real (i.e., non-simulated) domain.

The dataset is freely available for download\footnote{The \gls{selma} dataset is available at \url{https://scanlab.dei.unipd.it/app/dataset}.}\hspace{-3pt}, thus supporting open science and stimulating further research in the field of autonomous driving.

The remainder of this paper is organized as follows. In \Cref{sec:related} we describe the existing real and synthetic datasets related to our work. \Cref{sec:setup} presents the CARLA simulator and the additions we introduced to acquire the data.
The \gls{selma} dataset is then described in detail in \cref{sec:design}, while Secs. \ref{sec:experiments} and \ref{sec:UDA} show some numerical results validating the accuracy of models trained on our dataset. Finally, in \cref{subsec:conclusions} we provide the conclusions and some future research directions.

\section{Related Work}
\label{sec:related}
The development of \gls{dl} architectures seen in recent years goes along with the design of extensive datasets, needed for their optimization. 
One \gls{cv} task where such advancements have been particularly significant is scene understanding, which evolved in several sub-tasks, each requiring appropriate data for training. Among them, three tasks are worth mentioning, given the strong push they provided to datasets design: {image classification}, {object detection} and {semantic segmentation}. 
The first sparked the generation of widely used datasets, e.g.,  ImageNet~\cite{deng2009imagenet}.
The second and third tasks have been widely applied to many problems, and especially to autonomous driving systems. Here, vehicles require accurate recognition of the surrounding environment to appropriately plan driving actions. 
This translated into a wide range of real and synthetic datasets to support the training of autonomous driving applications~\cite{9000872}.
In this work, we focus on the semantic segmentation task, generally recognized as the most challenging of the three. The most popular \gls{ss} datasets existing in the literature and their characteristics are reported in \Cref{tab:related}.

\paragraph{Real datasets}
Given the high complexity and cost of labeling, most wide-scale real datasets tend not to provide the ground truth, e.g.{, \gls{ss} labels or bounding boxes}, thus limiting their use in  tasks like semantic segmentation and {object detection}. 
One of the first works to introduce labeled \gls{ss} samples in the context of autonomous driving was CamVid~\cite{BROSTOW200988}, which consists of over $700$ images labeled in 32 classes. 
Based on this, a huge effort was made by the creators of KITTI~\cite{Geiger2012CVPR,Geiger2013IJRR,Fritsch2013ITSC,Menze2015CVPR} to provide the first multimodal (stereo RGB and LiDAR) dataset for road scenes. 
This dataset, unfortunately, consists of only a small subset of $200$ \gls{ss} training images. The next fundamental step was the acquisition of the Cityscapes~\cite{Cordts2016} dataset, which was the first collection of labeled samples large enough to support training of deep architectures to a satisfactory level.
It includes $5\,000$ finely-labeled samples and $20\,000$ coarsely-labeled samples captured in several German cities, and has become an important benchmark for the segmentation task.
Recent works focus more on the volume~\cite{neuhold2017mapillary,yu2018bdd100k,varma2019idd} and variability~\cite{caesar2020nuscenes,lyft2019,wang2019apolloscape, ma2019trafficpredict} of data.
Even more recently, researchers are supporting the advent of LiDAR sensors, and some datasets have been generated accordingly \cite{behley2019iccv,hackel2017isprs,pan2020semanticposs,jiang2021lidarnet}.

\begin{table}
\centering
\setlength\tabcolsep{1.1pt} 
\caption{Comparison among the most popular SS datasets. White rows refer to real datasets, dark grey refers to synthetic datasets, and light grey refers to a combination of the two. Best in \textbf{bold}, runner-up \underline{underlined}.  T: Type, R: Real, M: Mixed, S: Synthetic, BB: Bounding Boxes, W: Weathers, ToD: daytime, AA: Anti-Aliasing, $\dagger$: estimated depth, *: random.}
\resizebox{\columnwidth}{!}{
\begin{tabular}{|c|cccccccccccccc|}
\hline
\multirow{2}{*}{Name} & \multirow{2}{*}{T} & \multirow{2}{*}{Cams} & \multirow{2}{*}{Depth} & \multirow{2}{*}{LiDAR} & \multirow{2}{*}{BB} & \multicolumn{2}{c}{Labels} & \multicolumn{3}{c}{SS/BB CS} & \multirow{2}{*}{W} & \multirow{2}{*}{ToD} & \multirow{2}{*}{AA} & \multirow{2}{*}{Positions} \\ 
 &  &  &  &  &  & RGB & LiDAR & all & train & classes &  &  &  & \\\hline
A2D2~\cite{geyer2020a2d2} & R & 6 & - & \textbf{5} & \cmark & 1 & 1 & 38 & \underline{38} & 12 & - & - & - & 41280\\
ACDC~\cite{sakaridis2021acdc}  & R & 1 & - & \xmark & \xmark & 1 & - & 19 & 19 & \textbf{19} & \textbf{3} & \underline{2} & - & 4006\\
Apolloscape~\cite{wang2019apolloscape}  & R & 6 & 2\textsuperscript{$\dagger$} & 2 & \cmark & 1 & 1 & 36 & 22 & 12 & * & * & - & 143906\\
Argo~\cite{Chang_2019_CVPR}  & R & \textbf{9} & 2\textsuperscript{$\dagger$} & 2 & \cmark & - & - & 15 & 15 & 5 & - & - & - & N/A\\
BDD 100k~\cite{yu2018bdd100k}  & R & 1 & \xmark & \xmark & \cmark & 1 & - & \underline{40} & 19 & \textbf{19} & * & * & - & 10000\\
CamVID~\cite{BROSTOW200988} & R & 1 & \xmark & \xmark & \cmark & 1 & - & 32 & 11 & 11 & - & - & - & 701\\
Cityscapes~\cite{Cordts2016}  & R & 2 & 2\textsuperscript{$\dagger$} & \xmark & \cmark & 1 & - & 35 & 19 & \textbf{19} & - & - & - & 25000\\
DarkZurich~\cite{Sakaridis_2019_ICCV}  & R & 1 & \xmark & \xmark & \xmark & \xmark & - & 19 & 19 & \textbf{19} & - & \textbf{3} & - & 8779\\
DRIV100~\cite{sakashita2021driv100}  & R & 1 & \xmark & \xmark & \xmark & \xmark & - & 19 & 19 & \textbf{19} & - & - & - & N/A\\
IDD~\cite{varma2019idd}  & R & 1 & \xmark & \xmark & \xmark & 1 & - & 34 & 25 & \textbf{19} & - & - & - & 10003\\
KITTI~\cite{Geiger2012CVPR} & R & 2 & 4\textsuperscript{$\dagger$} & 1 & \cmark & \xmark & \xmark & 8 & 8 & 8 & - & - & - & N/A\\
Mapillary~\cite{neuhold2017mapillary}  & R & 1 & \xmark & \xmark & \cmark & 1 & - & \textbf{66} & \textbf{66} & \textbf{19} & * & * & - & 25000\\
NTHU~\cite{chen2017no}  & R & 1 & \xmark & \xmark & \xmark & \xmark & - & 13 & 13 & 13 & - & - & - & 12800\\
Nuscenes~\cite{caesar2020nuscenes}  & R & 6 & \xmark & 1 & \cmark & \xmark & 1 & 23 & 23 & - & - & - & - & 40000\\
RainCouver~\cite{tung2017raincouver}  & R & 1 & \xmark & \xmark & \xmark & \xmark & - & 3 & 3 & - & \underline{1} & \textbf{3} & - & N/A\\
SemanticKITTI~\cite{behley2019iccv}  & R & - & - & 1 & \xmark & - & 1 & 28 & 20 & 15 & - & - & - & 43552\\
Nightcity~\cite{tan2021night}  & R & 1 & \xmark & \xmark & \xmark & 1 & - & 19 & 19 & \textbf{19} & - & 1 & - & 4297\\ \hdashline
\rowcolor{mixed}
CS Fog~\cite{sakaridis2018semantic}  & M & 2 & 2\textsuperscript{$\dagger$} & \xmark & \cmark & 1 & - & 35 & 19 & \textbf{19} & \underline{1} & - & - & 5000\\
\rowcolor{mixed}
CS Rain~\cite{hu2019depth}  & M & 2 & 2\textsuperscript{$\dagger$} & \xmark & \cmark & 1 & - & 34 & 18 & \underline{18} & \underline{1} & - & - & 5000\\ \hdashline
\rowcolor{synth}
GTA5~\cite{Richter2016}  & S & 1 & \xmark & \xmark & \xmark & 1 & - & 35 & 19 & \textbf{19} & - & * & \cmark & 24966\\
\rowcolor{synth}
IDDA~\cite{alberti2020idda}  & S & 1 & 1 & \xmark & \xmark & 1 & - & 24 & 24 & 16 & \textbf{3} & - & \xmark & 16000\\
\rowcolor{synth}
SYNTHIA~\cite{ros2016synthia}  & S & 1 & 1 & \xmark & \xmark & 1 & - & 23 & 16 & 16 & * & * & \xmark & 9400\\ \hdashline
\rowcolor{synth}
SELMA (Ours) & S & \underline{7} & \textbf{7} & \underline{3} & \cmark & \textbf{7} & \textbf{3} & 36 & 19 & \textbf{19} & \textbf{9} & \textbf{3} & \cmark & 30909\\
\hline
\end{tabular}
\label{tab:related}
}
\end{table}

\paragraph{Synthetic datasets} To circumvent the cost involved in the labeling of large-scale datasets, particularly those for \gls{ss}, many synthetically-generated datasets have been proposed over the years.
The first two important benchmarks are GTA5~\cite{Richter2016} and SYNTHIA~\cite{ros2016synthia}, both introduced in 2016. 
The former was generated exploiting the homonymous game, and provides $25\,000$ samples of realistic high-quality images. The semantic labels are provided in the same class set as Cityscapes~\cite{Cordts2016}, although they were inferred by the authors from secondary shader data, and the classes assigned to objects are not always consistent. 
The latter was the first dataset to provide depth ground truth for each of its $9\,000$ samples. The class set is different than that used by Cityscapes, and the overlap is limited to $16$ classes (see \cref{subsec:results_appendix:supervised} for details on the class splits).
A third important synthetic dataset is Virtual KITTI~\cite{gaidon2016virtual, cabon2020vkitti2} which, like its real counterpart, focused heavily on the multimodal aspect. 
It was the first to provide ground truth optical-flow and instance segmentation data, in addition to color, depth and semantic. More recently, the IDDA dataset~\cite{alberti2020idda} was introduced to address the lack of weather conditions variability in available datasets. It was developed using CARLA~\cite{Dosovitskiy17} and includes semantic labels (with an overlap of $16$ classes with Cityscapes), depth and RGB~data.

From \Cref{tab:related} we can see that, among the synthetic datasets, SELMA is the only one to provide labeled data for multiple LiDAR and camera sensors. 
Moreover, it is the only one that provides multiple weather conditions while supporting the full Cityscapes~\cite{Cordts2016} class set, as opposed to IDDA. Even more, it is the only one that provides 3D bounding boxes.
Finally, compared to GTA5~\cite{Richter2016}, i.e., the only competitor to provide anti-aliased color images, SELMA provides more samples, and considers a much higher variability of setups and sensors.

\section{Simulator Setup}
\label{sec:setup}
In this section we describe the CARLA simulator for generating synthetic automotive data (\cref{subsec:carla}, and the changes we introduced for SELMA (\cref{subsec:customization}).

\subsection{The CARLA Simulator}
\label{subsec:carla}
The \gls{carla} simulator is used to generate synthetic data relative to autonomous driving systems. It is designed as an open-source layer over \gls{ue4} to provide high-quality rendering, realistic physics based on the NVIDIA PhysX engine, and basic \gls{npc} logic~\cite{Dosovitskiy17}.
Reproducible and reliable physics simulations, as well as realistic and synchronized sensor data, can be obtained through the \gls{carla} \gls{api}.
Hereby, we briefly report the main characteristics of release 0.9.12, which was the starting point for the customization we made to meet the desired characteristics of the dataset, as detailed in \cref{subsec:customization}.

\paragraph{Unreal Engine Models}
\gls{carla} offers a wide variety of carefully designed \gls{ue4} models for static (e.g., buildings, vegetation, traffic signs) and dynamic objects (e.g., vehicles and pedestrians), sharing a common scale, and with realistic sizes.
In release 0.9.12, the blueprint library includes the model of $24$ cars, $6$ trucks, $4$ motorbikes, $3$ bikes, each with customizable colors, and $41$ pedestrian models of different ethnicity, build, and dressed with a wide variety of clothes.
Furthermore, 8 towns (Town01-07 and Town10HD) were carefully designed by the \gls{carla} team using more than $40$ building models.
Each town has its unique features and landmarks, thus offering 8 simulation environments with diverse visual characteristics. 

\paragraph{Sensors}
Data from the simulated world can be retrieved through a number of different sensors (see \cref{sec:carla_appendix} for a detailed list
of supported sensors), that can be placed at an exact location and a given rotation, and attached to a parent actor, thus following its movements with a rigid or spring-arm-like behavior. 
Sensors data can be collected at each simulation step. When working with multiple, high-resolution sensors, synchronous mode is required to guarantee that the GPU completes the rendering and delivers the data to the client before updating to the next simulation step. Thus, the sensor acquisition rate is the same for all.

\paragraph{Weather Conditions and Daytime}
Leveraging the underlying \gls{ue4} graphics, \gls{carla} offers a variety of \textit{daytime} and \textit{weather} conditions. The combination of daytime and weather will be referred to as \textit{environmental conditions} in the rest of the paper. 
Such conditions differ in the position and color of the sun, and in the intensity and color of diffuse sky radiation (daytime), as well as ambient occlusion, fog, cloudiness, and precipitation (weather). In release 0.9.12, there are $14$ predefined environmental conditions, obtained by the combination of two daytimes (Noon and Sunset), and seven weather conditions (Clear, Cloudy, Wet, WetCloudy\footnote{\emph{Wet} and \emph{WetCloudy} indicate that the road is more reflective and contains puddles. Notably, the former (latter) specifies that observations are acquired in clear (cloudy) sky.}, SoftRain, MidRainy, HardRain).

\begin{figure}[t]
    \centering
    \begin{subfigure}[b]{0.32\columnwidth}
        \centering
        \includegraphics[width=\textwidth]{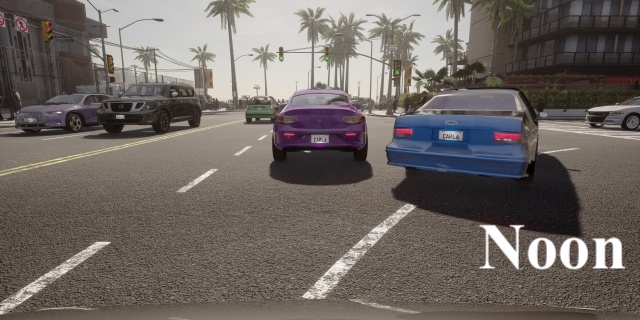}
        \label{fig:noon}
    \end{subfigure}
    \begin{subfigure}[b]{0.32\columnwidth}
        \centering
        \includegraphics[width=\textwidth]{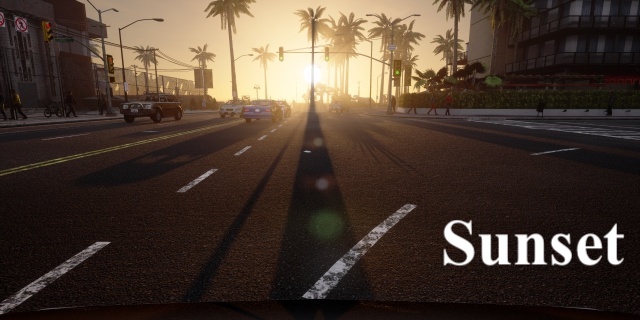}
        \label{fig:sunset}
    \end{subfigure}
    \begin{subfigure}[b]{0.32\columnwidth}
        \centering
        \includegraphics[width=\textwidth]{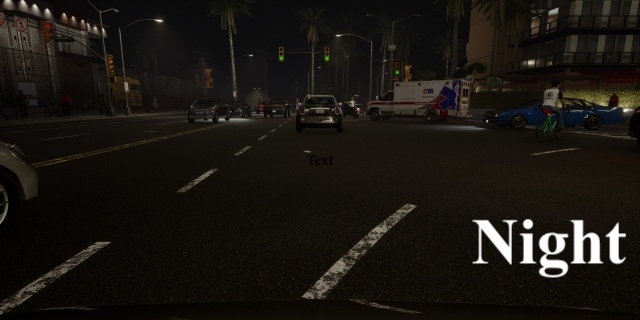}
        \label{fig:night}
    \end{subfigure}
    \caption{Desk view at three different times of the day.}
    \label{fig:hour}
\end{figure}

\newcommand{\sizefiggg}{0.325}
\begin{figure}[tbp]{}
\setlength\tabcolsep{1.2pt} 
\centering
\begin{tabular}{ccc}
  \includegraphics[width=\sizefiggg\columnwidth]{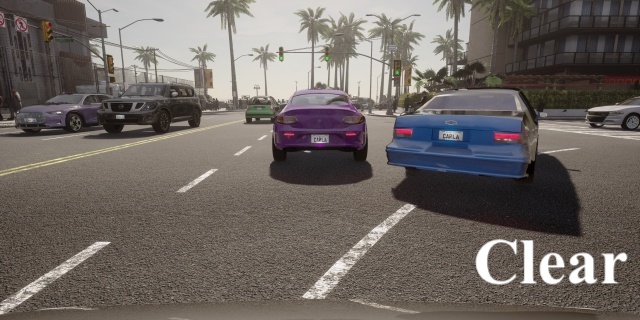} & \includegraphics[width=\sizefiggg\columnwidth]{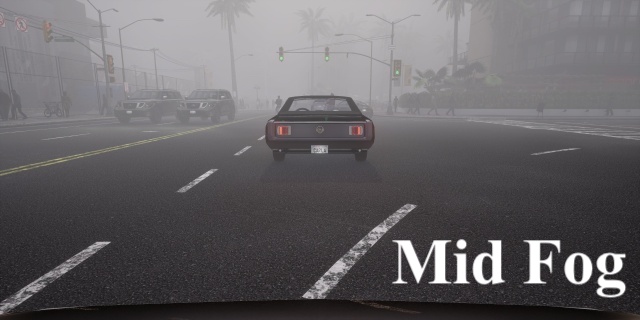} &
\includegraphics[width=\sizefiggg\columnwidth]{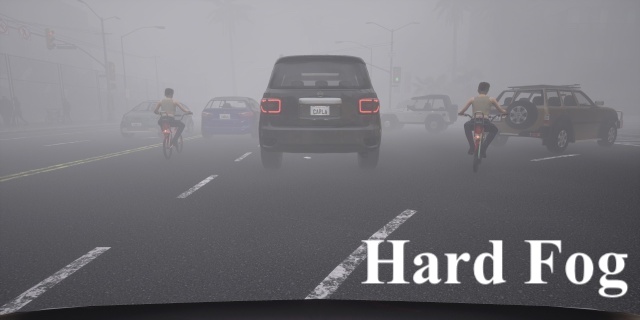}\\
  \includegraphics[width=\sizefiggg\columnwidth]{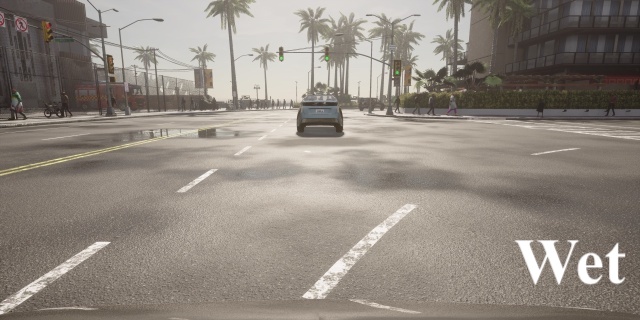} & \includegraphics[width=\sizefiggg\columnwidth]{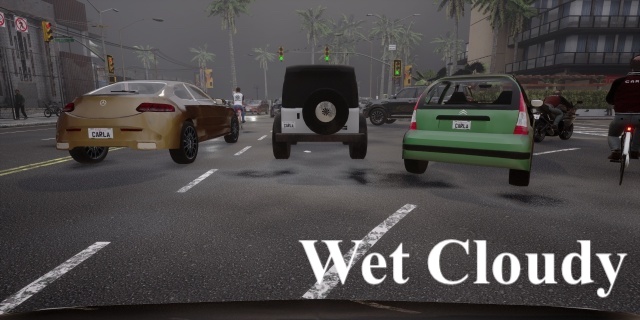} &
\includegraphics[width=\sizefiggg\columnwidth]{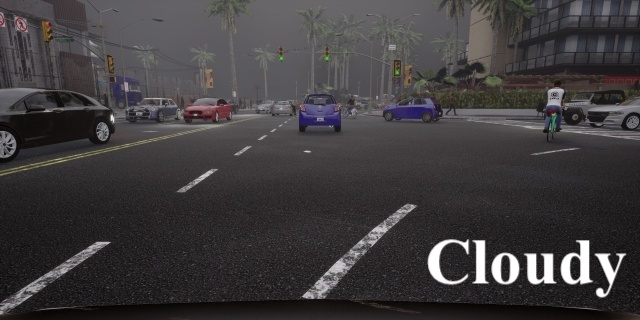}\\
  \includegraphics[width=\sizefiggg\columnwidth]{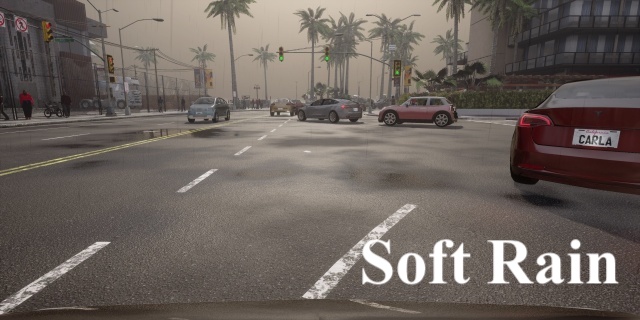} & \includegraphics[width=\sizefiggg\columnwidth]{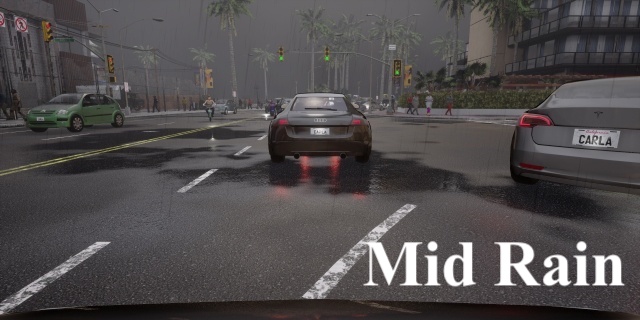} &
\includegraphics[width=\sizefiggg\columnwidth]{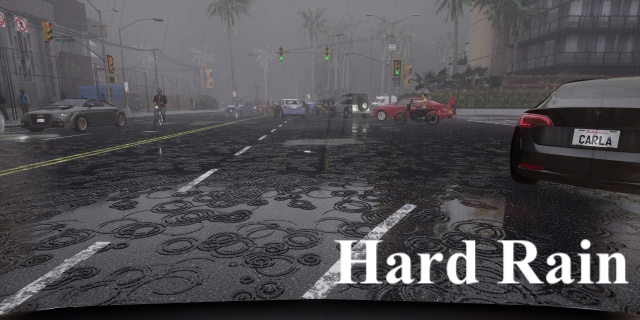}
 \end{tabular}
\caption{Samples in $9$ variable weather conditions at Noon.}
\label{fig:weathers_noon}
\end{figure}

\subsection{Customization}
\label{subsec:customization}
To enhance the quality of the collected data, we customized the source code of \gls{carla}, as detailed in the following.

First, we adjusted the parameters of the predefined environmental conditions, modifying the weather scattering and fog properties, and the position of the sun, to maximize the diversity between the environmental conditions and their photo-realism.
Then, we introduced the Night daytime (\cref{fig:hour}) and the Mid Fog and Hard Fog weather conditions (\cref{fig:weathers_noon}).
Thus, the number of daytimes and weather conditions has been increased from $2$ and $7$ to $3$ and $9$, respectively, bringing the total number of environmental conditions to $27$.

\begin{figure}
    \centering
    \includegraphics[width=.48\textwidth]{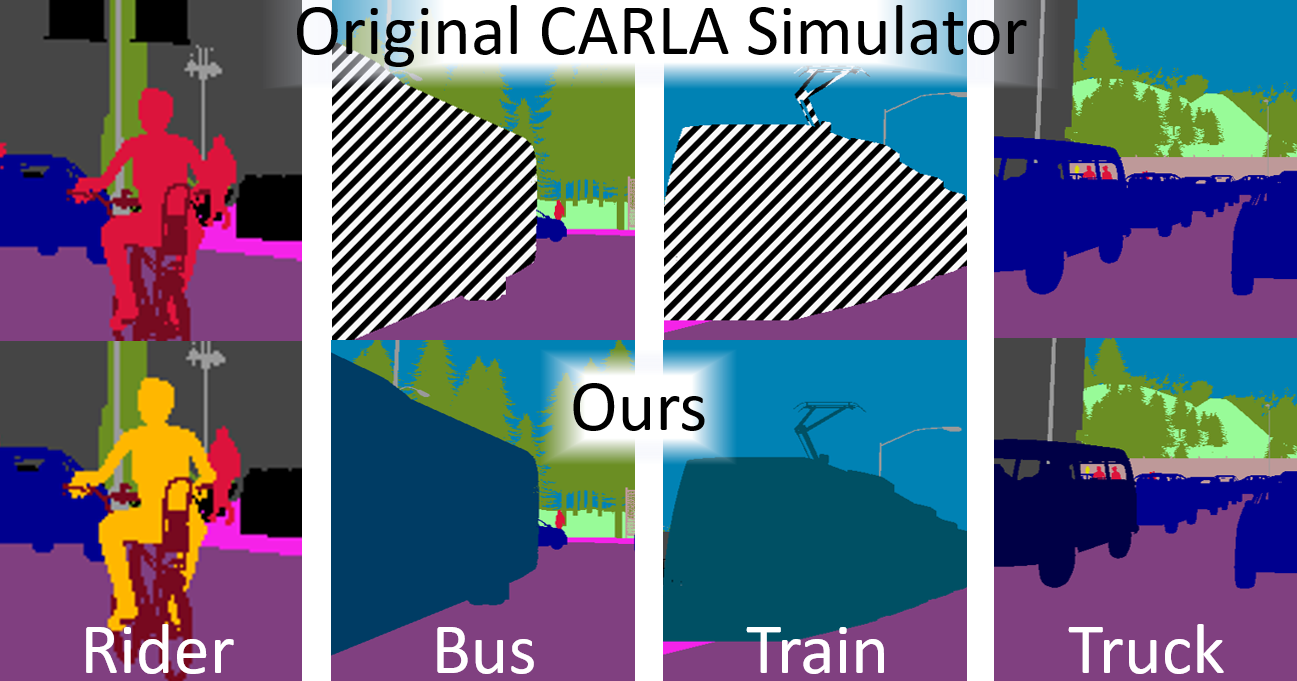}
    \caption{Comparison between the original version of the CARLA simulator and that with our modifications. Dashed regions indicate classes that were originally missing in CARLA. Notice that our implementation now distinguishes riders from people and trucks from cars.}
    \label{fig:added_classes}
\end{figure}

Second, we modified the \gls{carla} semantic classes, to make them compatible with existing benchmark datasets, and added new vehicle models to increase the class diversity. Specifically, the remapping of the classes was done in the source code to affect both the semantic LiDAR and the semantic camera.
We introduced the \textit{Train} class, adding a train and a tram model, and we added two bus and two truck models to the existing classes.
Then, our modifications to the source code allowed us to introduce the \textit{Rider} class, adding the corresponding tag and separating the rider from its bike/motorbike tag. A visual example is reported in \Cref{fig:added_classes}.
Finally, as of release 0.9.12, the parked vehicles were not labeled correctly. Exploiting the \gls{carla} \gls{api}, we removed the corresponding map layer, saving the location information to place vehicles with the correct tag in the exact same position.

\begin{figure}[t]
    \centering
    \includegraphics[trim={0cm 0.8cm 0cm 0cm}, clip, width=\columnwidth]{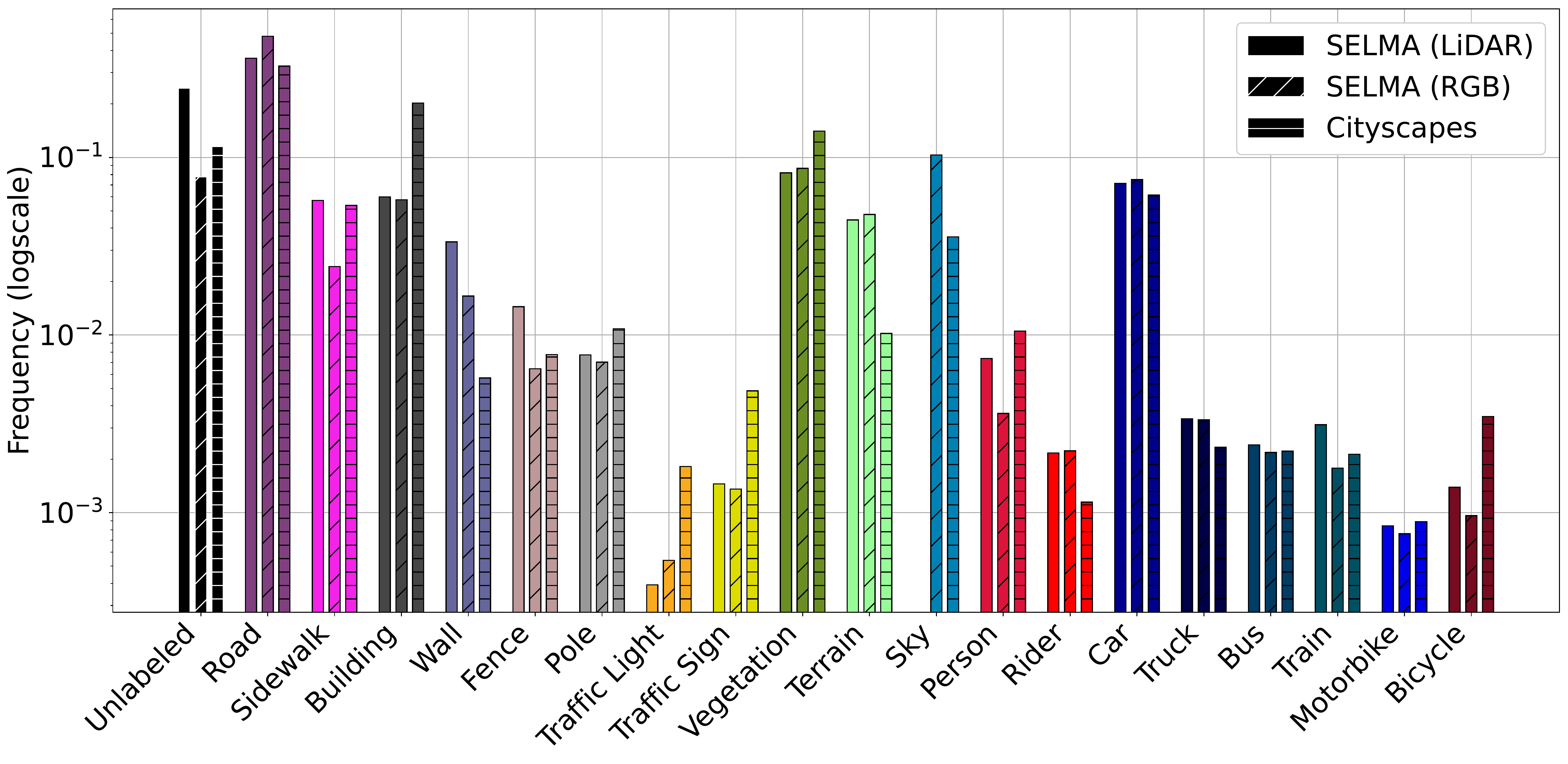}
    \caption{Class distributions in the \gls{selma} dataset.}
    \label{fig:freqs}
\end{figure}

Then, the \gls{ue4} content was modified to meet the strict requirements that we set for the \gls{selma} dataset.
Namely, bikes could only exist along with their rider on board, which prevented parked bikes to be deployed.
Nonetheless, bikes are amongst the main road actors, and \gls{cv} algorithms greatly benefit from visual variability. Indeed, we deemed fundamental for our dataset to include the bike class in all the contexts.
Therefore, we added hundreds of parked bikes into the existing \gls{carla} maps.
Similarly, the number of traffic lights and signs in the default towns did not reflect their distribution in a real setting, thus possibly compromising the learning ability of the algorithms.
Thus, we distributed tens of additional traffic lights and signs in every town. The final class distribution matches that of real-world reference datasets such as Cityscapes, as shown in \cref{fig:freqs}.
The customized simulator is freely available\footnote{Our customized version of CARLA will be made available upon acceptance.}

\graphicspath{{./imgs/design/}}
\section{SELMA Dataset Design}
\label{sec:design}
In this section we present our SELMA dataset, with a focus on the acquisition setup (\cref{ssec:sim_setup}) and splits (\cref{ssec:splits}).

\subsection{Acquisition Setup}
\label{ssec:sim_setup}

\begin{figure}[t]
    \centering
    \begin{subfigure}[b]{0.49\columnwidth}
        \centering
        \includegraphics[width=\columnwidth]{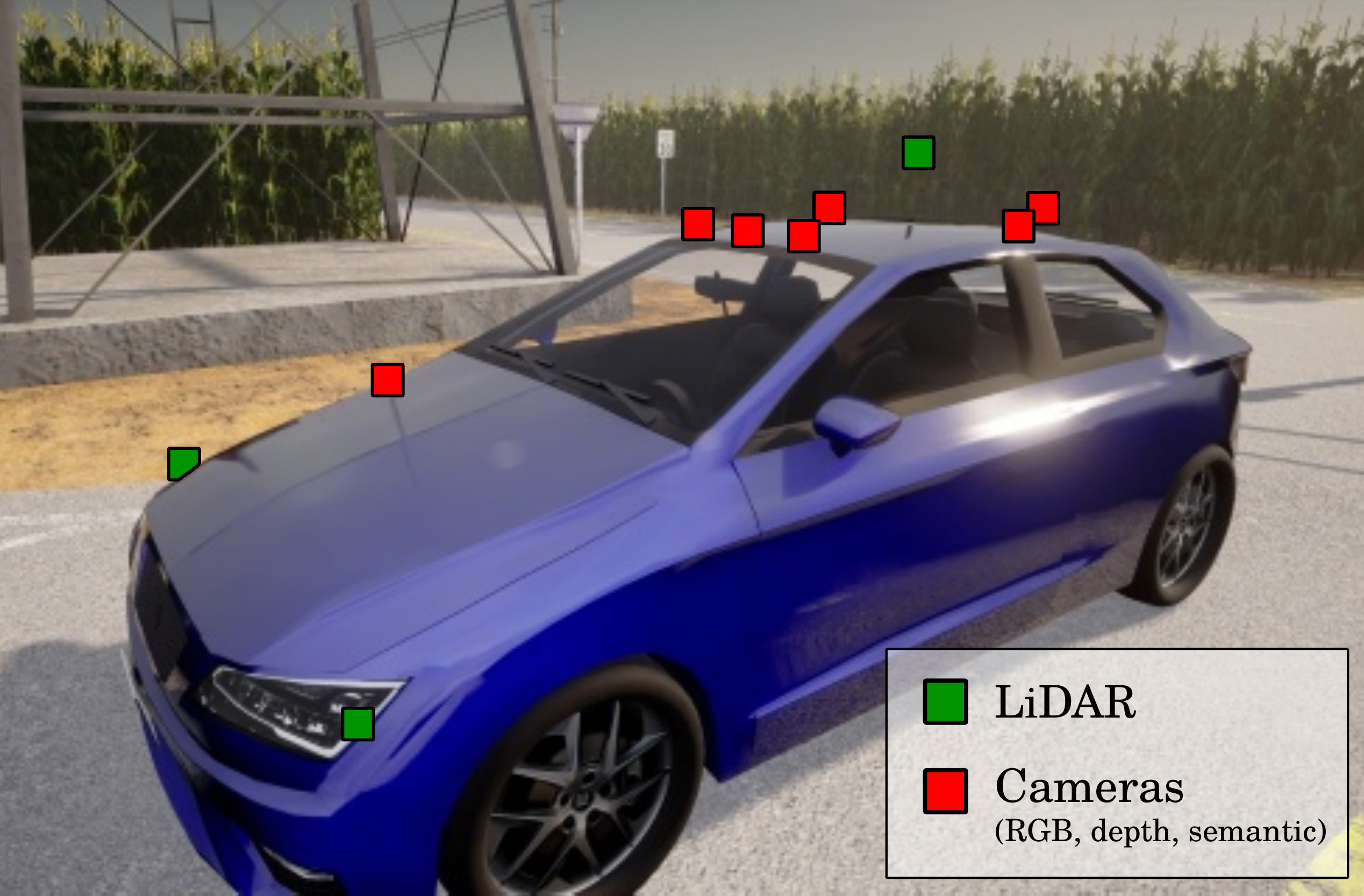}
        \caption*{Side View}
        \label{fig:sensors_latview}
    \end{subfigure}
    \begin{subfigure}[b]{0.49\columnwidth}
        \centering
        \includegraphics[width=\columnwidth]{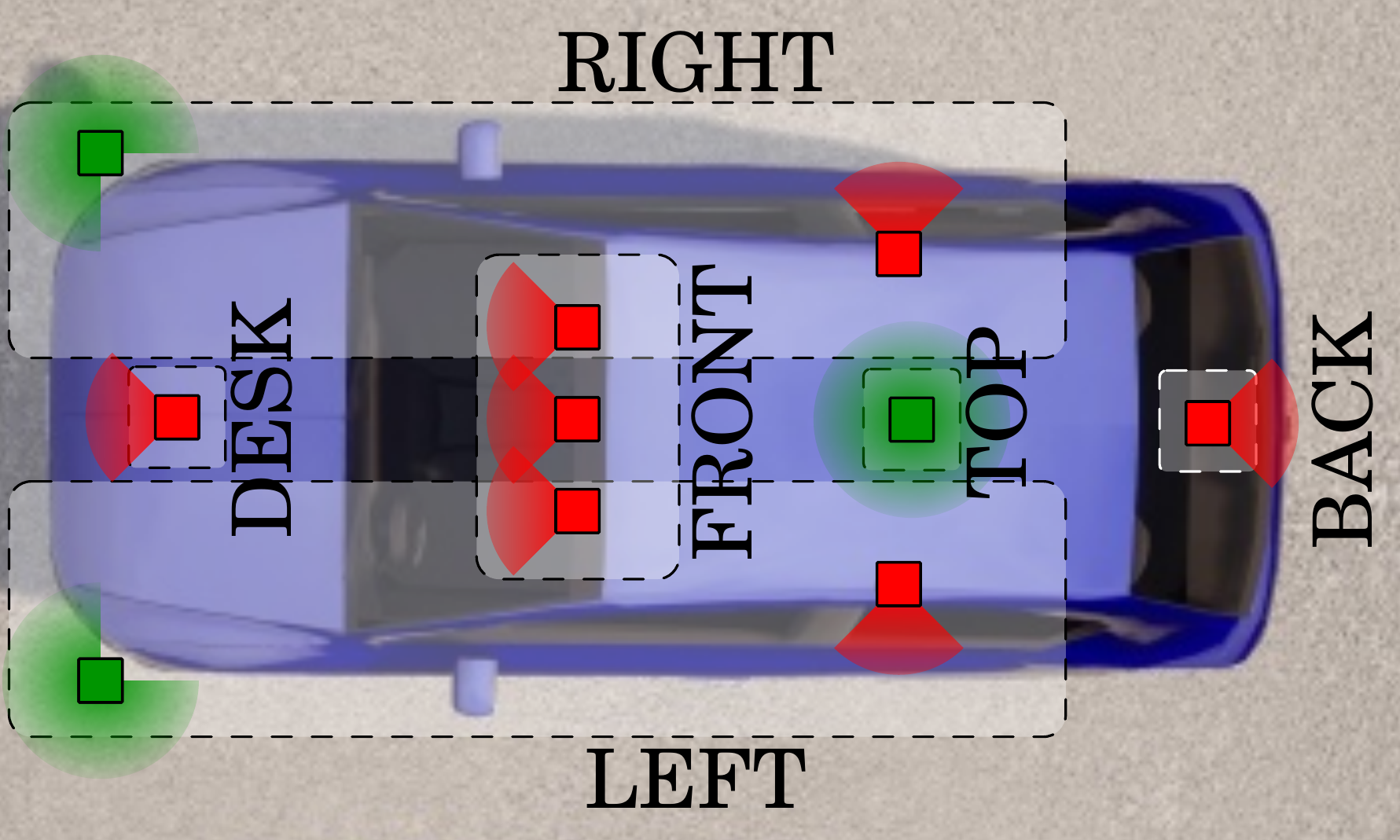}
        \caption*{Top View}
        \label{fig:sensors_topview}
    \end{subfigure}
    \caption{Side and top views of the sensor setup in SELMA. RGB, depth and semantic cameras are co-located in $7$ spots. LiDARs are placed in $3$ locations.}
    \label{fig:sensor_setup}
\end{figure}

\renewcommand{\sizefiggg}{0.118}
\begin{figure*}[tbp]{}
\setlength\tabcolsep{1.pt}
\def\arraystretch{.3}
\centering
\begin{tabular}{ccccccccc}
   & \footnotesize{TOWN 01} & \footnotesize{TOWN 02} & \footnotesize{TOWN 03} & \footnotesize{TOWN 04} & \footnotesize{TOWN 05} & \footnotesize{TOWN 06} & \footnotesize{TOWN 07} & \footnotesize{TOWN 10HD} \\
  \rotatebox{90}{\ \ \footnotesize{front}} & 
  \includegraphics[width=\sizefiggg\linewidth]{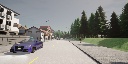} & \includegraphics[width=\sizefiggg\linewidth]{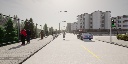} &
\includegraphics[width=\sizefiggg\linewidth]{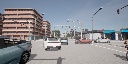} &
\includegraphics[width=\sizefiggg\linewidth]{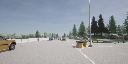} &
\includegraphics[width=\sizefiggg\linewidth]{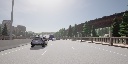} &
\includegraphics[width=\sizefiggg\linewidth]{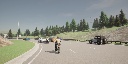} &
\includegraphics[width=\sizefiggg\linewidth]{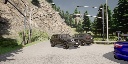} &
\includegraphics[width=\sizefiggg\linewidth]{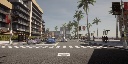}\\

\rotatebox{90}{\ \ \footnotesize{desk}} & 
  \includegraphics[width=\sizefiggg\linewidth]{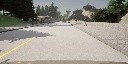} & \includegraphics[width=\sizefiggg\linewidth]{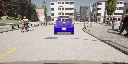} &
\includegraphics[width=\sizefiggg\linewidth]{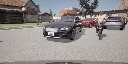} &
\includegraphics[width=\sizefiggg\linewidth]{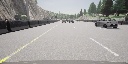} &
\includegraphics[width=\sizefiggg\linewidth]{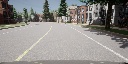} &
\includegraphics[width=\sizefiggg\linewidth]{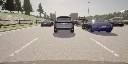} &
\includegraphics[width=\sizefiggg\linewidth]{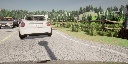} &
\includegraphics[width=\sizefiggg\linewidth]{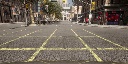}\\

\rotatebox{90}{\ \ \footnotesize{left}} & 
  \includegraphics[width=\sizefiggg\linewidth]{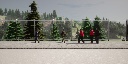} & \includegraphics[width=\sizefiggg\linewidth]{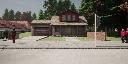} &
\includegraphics[width=\sizefiggg\linewidth]{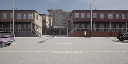} &
\includegraphics[width=\sizefiggg\linewidth]{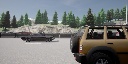} &
\includegraphics[width=\sizefiggg\linewidth]{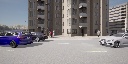} &
\includegraphics[width=\sizefiggg\linewidth]{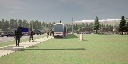} &
\includegraphics[width=\sizefiggg\linewidth]{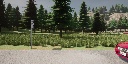} &
\includegraphics[width=\sizefiggg\linewidth]{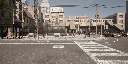}\\

\rotatebox{90}{\ \ \footnotesize{right}} & 
  \includegraphics[width=\sizefiggg\linewidth]{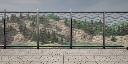} & \includegraphics[width=\sizefiggg\linewidth]{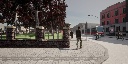} &
\includegraphics[width=\sizefiggg\linewidth]{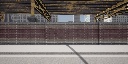} &
\includegraphics[width=\sizefiggg\linewidth]{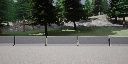} &
\includegraphics[width=\sizefiggg\linewidth]{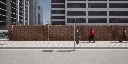} &
\includegraphics[width=\sizefiggg\linewidth]{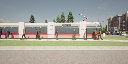} &
\includegraphics[width=\sizefiggg\linewidth]{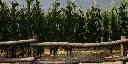} &
\includegraphics[width=\sizefiggg\linewidth]{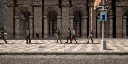}\\

\rotatebox{90}{\ \ \footnotesize{back}} & 
  \includegraphics[width=\sizefiggg\linewidth]{imgs/design/random_samples_ClearNoon/CAM_BACK_Town01_Opt_ClearNoon_4839467463213363345.jpg} & \includegraphics[width=\sizefiggg\linewidth]{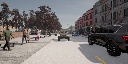} &
\includegraphics[width=\sizefiggg\linewidth]{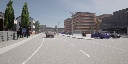} &
\includegraphics[width=\sizefiggg\linewidth]{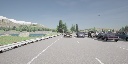} &
\includegraphics[width=\sizefiggg\linewidth]{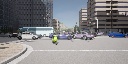} &
\includegraphics[width=\sizefiggg\linewidth]{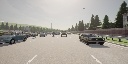} &
\includegraphics[width=\sizefiggg\linewidth]{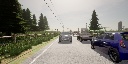} &
\includegraphics[width=\sizefiggg\linewidth]{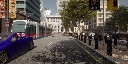}\\
 \end{tabular}
\caption{Randomly sampled images from the SELMA dataset in clear noon setup, demonstrating its diversity. Rows show different cameras, while columns show different (synthetic) towns (thus settings).}
\label{fig:rand_samples}
\end{figure*}

\begin{figure*}[t]
    \centering
    \begin{subfigure}[b]{0.24\textwidth}
        \centering
        \includegraphics[width=\textwidth]{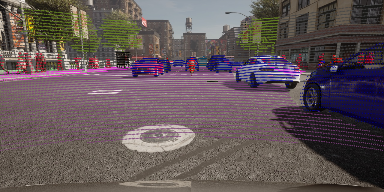}
        \caption*{Semantic LiDAR}
        \label{fig:lidar_left}
    \end{subfigure}
    \begin{subfigure}[b]{0.24\textwidth}
        \centering
        \includegraphics[width=\textwidth]{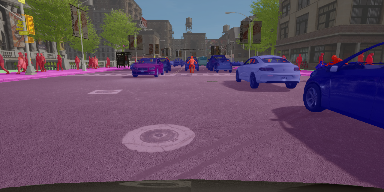}
        \caption*{Semantic Camera}
        \label{fig:seg_front}
    \end{subfigure}
    \begin{subfigure}[b]{0.24\textwidth}
        \centering
        \includegraphics[width=\textwidth]{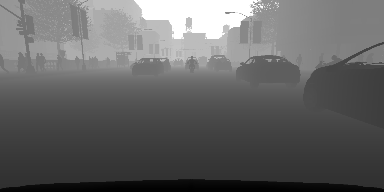}
        \caption*{Depth Camera}
        \label{fig:depth_right}
    \end{subfigure}
    \begin{subfigure}[b]{0.24\textwidth}
        \centering
        \includegraphics[width=\textwidth]{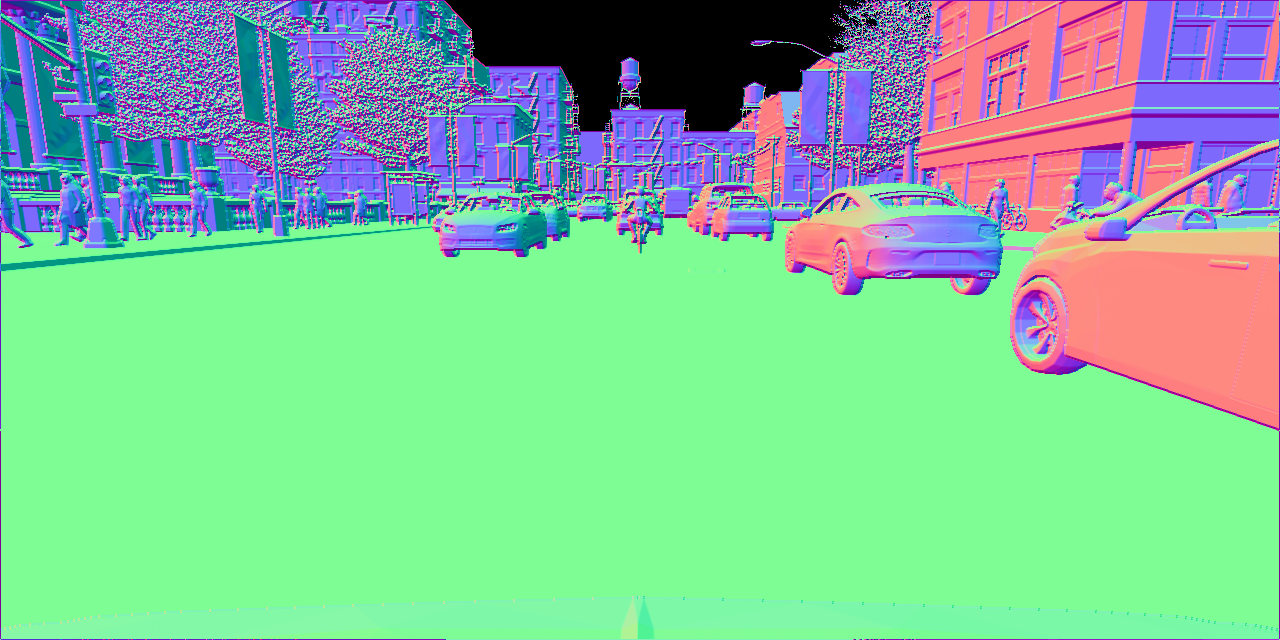}
        \caption*{Surface Normals}
        \label{fig:normals_desk}
    \end{subfigure}
    \caption{Sample acquisitions with semantic LiDAR, semantic camera overlaid to the RGB samples, depth camera, and surface normals.}
    \label{fig:sensor_samples}
\end{figure*}

We designed the acquisition pipeline to exploit the full potential of \gls{carla}, while maximizing the diversity of the acquired data.
Acquisitions were made equipping a vehicle, named the ego vehicle, with a full sensor suite depicted in \cref{fig:sensor_setup}, consisting of:
\begin{itemize}
    \item 7 RGB cameras, with the post-processing effects enabled, a 90-degree horizontal \gls{fov}, and a native resolution of $5120\times 2560$, which is downsampled to $1280\times 640$ to achieve a $\times 4$ anti-aliasing enhancement. The post-processing effects include vignette, grain jitter, bloom, auto exposure, lens flare and depth of field. RGB images are saved in JPEG format.
    \item 7 depth cameras with a 90-degree horizontal \gls{fov} and $1280\times 640$ resolution. For the depth images, anti-aliasing is not required and would compromise depth information. Depth images are saved in PNG format.
    \item 7 semantic cameras, that have the exact same attributes of the depth cameras. 
    \item 3 semantic LiDARs, each with $64$ vertical channels, generating $100\,000$ points per second, with a range of $100$ meters. Point clouds are saved in PLY format.
\end{itemize}
The different camera types are co-registered at $7$ different locations, and are set up to have the same \gls{fov} and resolution, so that their data can be easily matched. 
The variability of viewpoints and maps is shown in \cref{fig:rand_samples}.

\begin{table}[t]
    \centering
    \caption{Number of waypoints (WPs) per town. Data are acquired at each waypoint, independently for all environmental conditions.}
    \resizebox{\columnwidth}{!}{
    \begin{tabular}{|r|c|c|c|c|c|c|c|c||c|}
    \hline
        Town ID & 01 & 02 & 03 & 04 & 05 & 06 & 07 & 10HD & \textbf{Total}\\\hline
        \# of WPs  & 1634 & 756 & 3636 & 8565 & 6250 & 6579 & 1922 & 1568 & \textbf{30909}\\\hline
    \end{tabular}}
    \label{tab:waypoints}
    \vspace{-0.8cm}
\end{table}

Furthermore, we compute the 3D surface normals at each pixel of the image 
acquired by the desk camera for all samples of the default random split (see \cref{ssec:splits}). To do so, we employ a state-of-the-art differential technique as done in \cite{yang2018unsupervised,guizilini2021geometric}. While the result is an approximation of the true normals, the overall precision is very high, as can be seen in the last column of \Cref{fig:sensor_samples}, thanks to the detailed (i.e., synthetic ground truth) and dense depth maps used for the estimation procedure.

Data were acquired in $30\,909$ independent locations, across 8 virtual towns as reported in \cref{tab:waypoints} and in \cref{sec:carla_appendix}. The locations were selected extracting a list of waypoints at a given distance. For our dataset, we selected points on the roads on every lane and junction, at the distance of $4$ meters, which was chosen after several empirical tests as it offered the best trade-off between area coverage and acquisition diversity. The full list of waypoints with their ID is provided with the dataset for each town.
At each position, the ego vehicle is created, traffic is generated around it, and pedestrians are randomly placed on the sidewalks.
After one second of simulation for the transient to end, the sensors are fired simultaneously, and their data retrieved and saved. The server is then reset and the simulation goes on with the following~waypoint.

The same process is repeated in 27 different environmental conditions. These include 3 daytimes (i.e., Noon, Sunset and Night) and 9 weather conditions (i.e., Clear, Cloudy, Wet, Wet and Cloudy, 2 Fog intensities and 3 Rain intensities). 
Traffic and pedestrians are generated randomly at every iteration, thus the same waypoint simulated under different environmental conditions presents different traffic conditions.
We refer to the combinations of environmental conditions and towns as scenes. Our dataset consists of $216$ scenes, obtained considering all the sensors and the complete combinations of the available weather and daytime conditions, viewpoints,~and~towns.

\subsection{Splits}
\label{ssec:splits}
Exploiting the fine-grained control on environmental conditions offered by \gls{selma}, we designed some default splits.
Particularly, we considered 6 different weather distributions: Random (\gls{selma} default), Mostly Clear (MC), Noon, Night, Rain and Fog. For more details on the splits, we refer the interested readers to \cref{sec:thematic_appendix}.

The Random split contains samples from all weather conditions and daytimes, sampled according to the probability distributions reported in \Cref{tab:split_probs}.
\begin{table*}[t]
    \caption{Probability distributions of environmental conditions in the different splits.}
    \label{tab:split_probs}
    \centering
    \setlength{\tabcolsep}{3pt}
    \renewcommand{\arraystretch}{1.1}
    \begin{tabular}{|c|ccc|cccc:cc:ccc|}
        \hline
         Split & Noon & Sunset & Night & Clear & Cloudy & Wet Road & \makecell{Wet Road\\and Cloudy} & \makecell{Mid \\ Fog} & \makecell{Hard \\ Fog} & \makecell{Soft \\ Rain} & \makecell{Mid \\ Rain} & \makecell{Hard \\ Rain} \\ \hline
         SELMA/Noon/Night & \scriptsize{$50$/$100$/$0 \%$}  & \scriptsize{$25$/$0$/$0 \%$} & \scriptsize{$25$/$0$/$100 \%$} & $35 \%$ & $20 \%$ & $10 \%$ & $10 \%$& $3.5 \%$& $3.5 \%$& $6 \%$& $6 \%$& $6 \%$\\
         Mostly Clear/Rain/Fog & $50 \%$& $25 \%$& $25 \%$& \scriptsize{$25$/$0$/$0 \%$} & \scriptsize{$25$/$0$/$0 \%$} & \scriptsize{$25$/$0$/$0 \%$} & \scriptsize{$25$/$0$/$0 \%$} & \scriptsize{$0$/$0$/$50 \%$} & \scriptsize{$0$/$0$/$50 \%$} & \scriptsize{$0$/$34$/$0 \%$} & \scriptsize{$0$/$33$/$0 \%$}  & \scriptsize{$0$/$33$/$0 \%$} \\\hline
    \end{tabular}
\end{table*}
Most of the samples come from high-visibility weather conditions: Clear, Wet (road), Cloudy and WetCloudy make up for 75\% of the~split.

In order to preserve the separation among training, validation and test samples, the splits are provided in CSV format, which allows to easily assign a given weather condition to a sample (and to override it, if needed).
The samples separation was done according to an 80:10:10 split rule for training, validation and test, respectively.

\graphicspath{{./imgs/CV_analyses/}}
\section{Experimental Validation}
\label{sec:experiments}

In this section we carefully analyze and validate our SELMA dataset. 
We start with a series of baseline experiments which serve as a reference  benchmark for future studies (\cref{subsec:baseline}). Then, we analyze the thematic subsets of our dataset (\cref{subsec:thematic}). To conclude, we show how different sensors can be employed jointly to improve the final segmentation accuracy (\cref{subsec:fusion}), and report some experiments exploiting multiple viewpoints (\cref{subsec:multiview}).

\subsection{Baseline Experiments}
\label{subsec:baseline}

\begin{table}[tbp]
  \centering
  \setlength\tabcolsep{3.3pt}
  \caption{mIoU of baseline \gls{ss} architectures on Cityscapes (CS) and subsets of SELMA for both RGB images (first 7 columns) and depth (last column).}
    \begin{tabular}{l|c:cccccc|c}
          & CS & SELMA  & Noon & Night & MC & Fog   & Rain  & Depth \\\hline
    DeeplabV2 \cite{chen2018deeplab} & 67.4 & 68.9  & \textbf{72.3}  & 68.2  & 69.9  & 68.0  & 68.7  & 73.4 \\
    DeeplabV3 \cite{chen2017rethinking} & \textbf{68.2} & \textbf{70.7}  & 71.7  & \textbf{68.4}  & 70.8  & \textbf{69.0}  & 67.8  & 72.7 \\
    FCN \cite{long2015}  & 64.8 & 68.2  & 71.1  & 66.1  & 69.1  & 64.5  & 66.8  &  \textbf{73.7} \\
    PSPNet \cite{zhao2017} & 65.3 & 68.4  & 71.2  & 67.2  & 69.8  & 66.8  & \textbf{69.0} & 73.6\\
    UNet  \cite{ronneberger2015u} & 36.8 & 36.2  & 41.8  & 35.7  & 36.3  & 28.6  & 37.8  & 28.8 \\
    \end{tabular}%
  \label{tab:supervised_backbones}%
\end{table}%
The first set of experiments is designed to provide a series of benchmark results for the \gls{selma} dataset.
The results in Table~\ref{tab:supervised_backbones} show the performance achieved employing different baseline \gls{ss} architectures, i.e., UNet \cite{ronneberger2015u}, FCN \cite{long2015}, PSPNet \cite{zhao2017}, DeepLab-V2 \cite{chen2018deeplab,chen2018encoder}, DeepLab-V3 \cite{chen2017rethinking}. All the networks are trained with SGD with momentum of rate $0.9$. The learning rate was decreased according to a polynomial decay of coefficient $0.9$, starting from $2.5\times10^{-4}$. The batch size was set to $3$ and the weight decay to $10^{-4}$.
The results for the individual classes are reported in \cref{subsec:results_appendix:supervised}.

\paragraph{RGB} Initially, we perform a series of experiments using the RGB images from Cityscapes and from the \gls{selma} desk camera in different environmental conditions.
First, we observe that the UNet architecture achieves poor results, since it is unable to deal with the large visual variability of our SELMA dataset. The other architectures share the same encoder module, i.e., a ResNet-101, and they all achieve similar \gls{miou} performance. 
Overall, DeepLab-V2 and V3 offer the best performance; for the sake of the performance-complexity trade-off we decided to employ DeepLab-V2 in all the following experiments.
The highest accuracy is obtained with the \gls{selma} Noon split, as RGB images are easier to segment. On the contrary, Night, Fog and Rain decrease the accuracy of models.

\begin{figure*}[t]
\centering
\begin{subfigure}[t]{\textwidth}
    \centering
    \begin{subfigure}[b]{.19\textwidth}
        \caption*{RGB}
        \includegraphics[width=\textwidth]{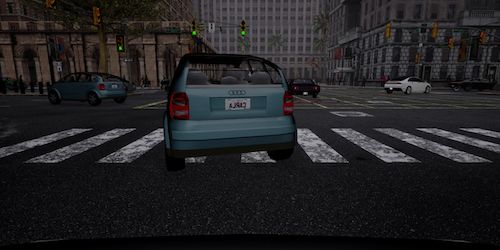}
    \end{subfigure}
    \begin{subfigure}[b]{.19\textwidth}
        \caption*{Depth}
        \includegraphics[width=\textwidth]{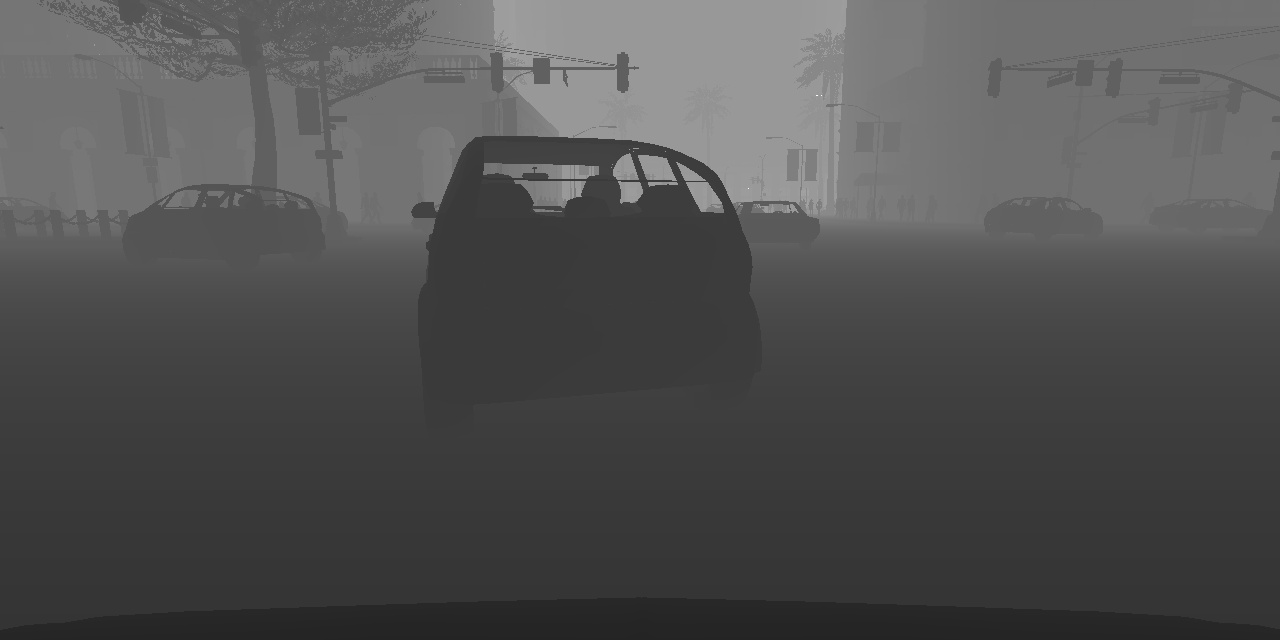}
    \end{subfigure}
    \begin{subfigure}[b]{.19\textwidth}
        \caption*{GT}
        \includegraphics[width=\textwidth]{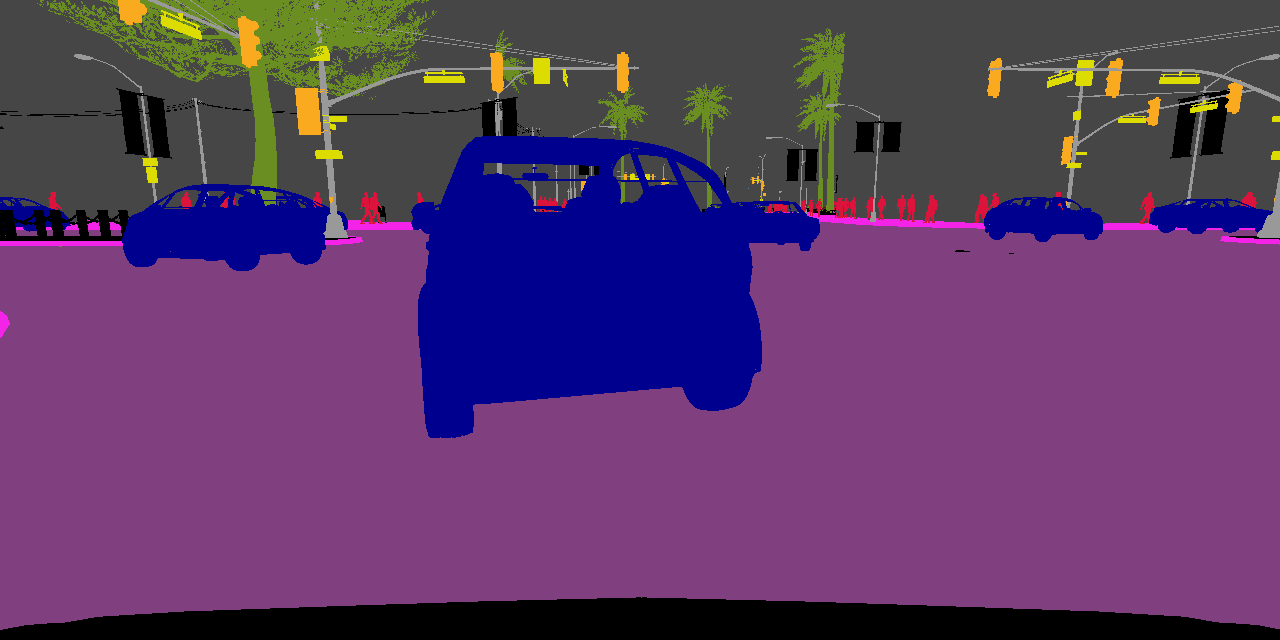}
    \end{subfigure}
    \begin{subfigure}[b]{.19\textwidth}
        \caption*{RGB prediction}
        \includegraphics[width=\textwidth]{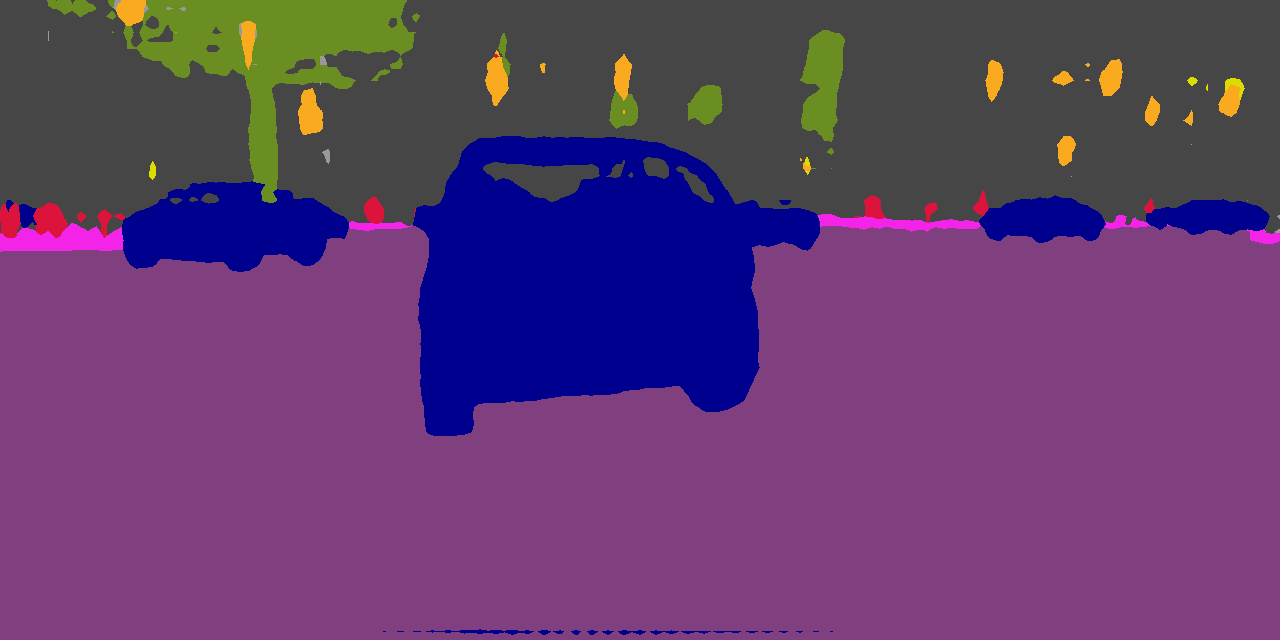}
    \end{subfigure}
    \begin{subfigure}[b]{.19\textwidth}
        \caption*{Depth prediction}
        \includegraphics[width=\textwidth]{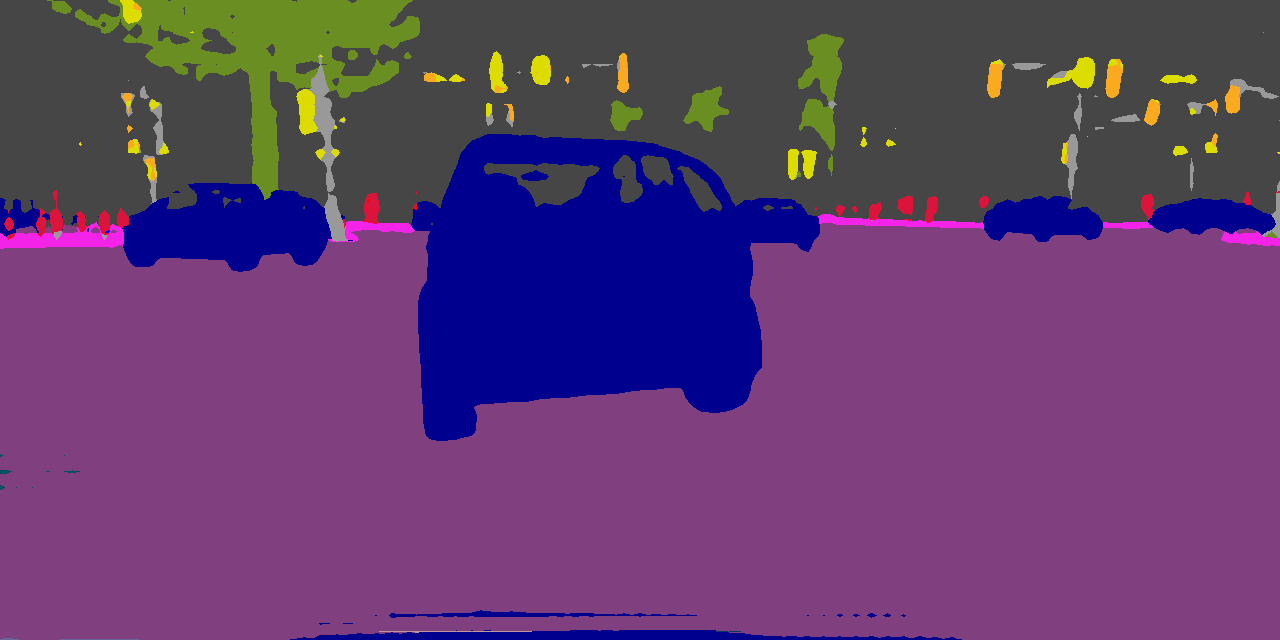}
    \end{subfigure}
\end{subfigure}
\begin{subfigure}{\textwidth}
    \centering
    \begin{subfigure}{.49\textwidth}
        \includegraphics[width=\textwidth]{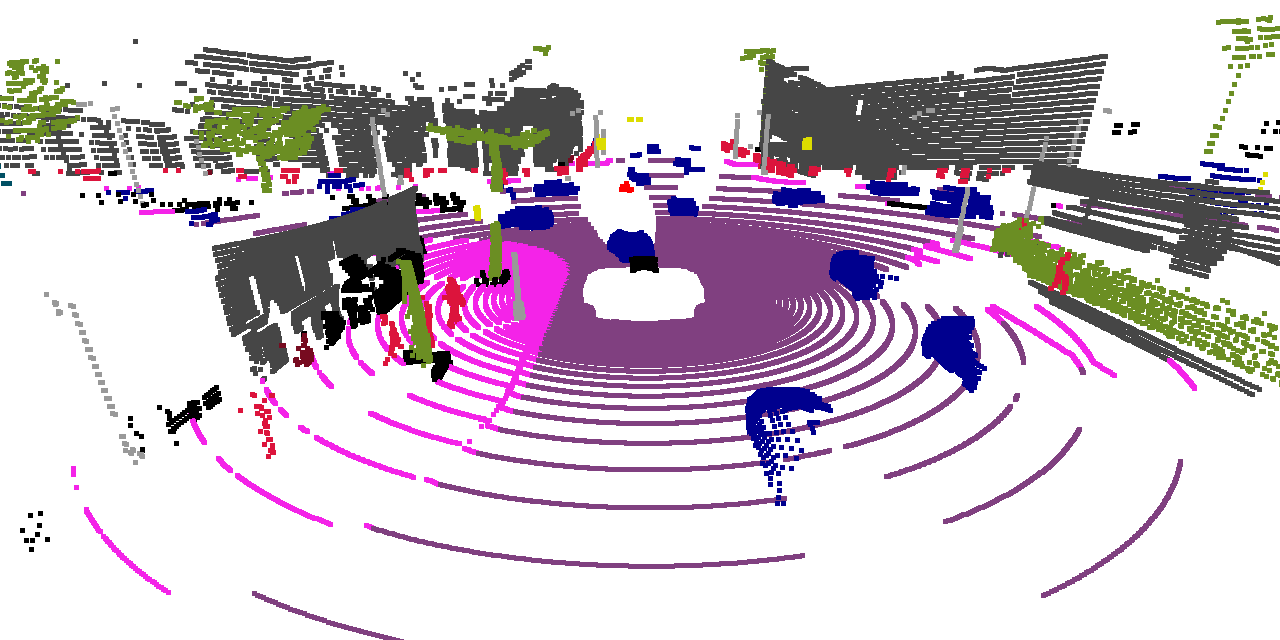}
        \caption*{LiDAR GT}
    \end{subfigure}%
    \begin{subfigure}{.49\textwidth}
        \includegraphics[width=\textwidth]{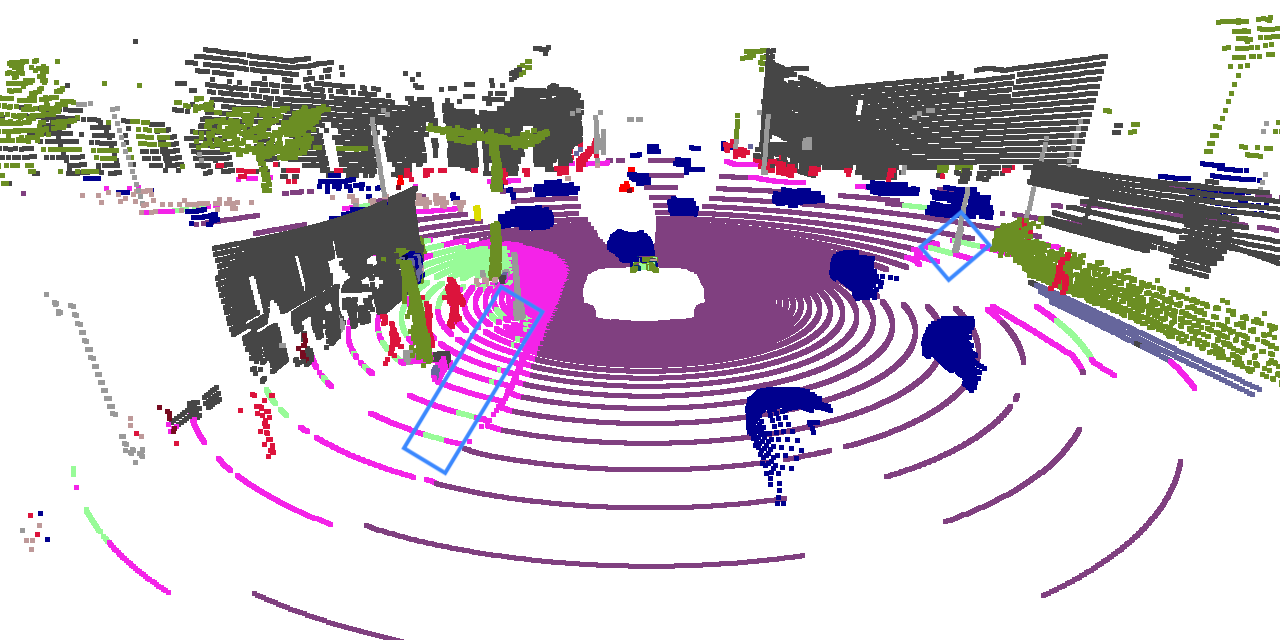}
        \caption*{LiDAR prediction}
    \end{subfigure}
\end{subfigure}
\caption{Qualitative results for the \gls{ss} task from RGB images (DeepLab-V2 \cite{chen2018deeplab}), depth maps (DeepLab-V2 \cite{chen2018deeplab}) or point clouds (Cylinder3D \cite{zhu2021cylindrical}).}
\label{fig:qualitative_baselines}
\end{figure*}

\paragraph{Depth} We run the same experimental evaluation using a single input channel representing the depth of the scene. More precisely, since the range of true values is extremely unbalanced and their distribution is highly skewed, we normalize and rescale the depth values: starting from the original depth produced by the simulator, normalized to $1$, we compute its fourth root to compress the high-distance information and expand the low-distance information. This is necessary as the sky is marked with the maximum distance possible, and overshadows the other pixels. Then, we rescale and shift the values to the $[-1,1]$ range.
Also in this case, the best performing architectures achieve comparable performance, as shown in the rightmost column of \cref{tab:supervised_backbones}.
The models trained on depth images can more easily segment objects of different classes, outperforming the results achieved on RGB samples.

\paragraph{LiDAR} Table~\ref{tab:supervised_backbones_lidar} reports the LiDAR \gls{ss} results obtained with RangeNet++ \cite{milioto2019icra-fiass} with two backbones (SqueezeSeg-V2 \cite{wu2019squeezesegv2} and DarkNet-21 \cite{redmon2018yolov3}) and Cylinder3D \cite{zhu2021cylindrical}. Furthermore, we report the results obtained by flattening the point cloud to an RGB image via spherical projection and employing DeepLab-V2 to segment it.
All the backbones are trained with batch size of $4$ for $40$ epochs with early stopping enabled. The other learning parameters are left to the default values provided in the respective codebases.
We can observe that Cylinder3D outperforms the other architectures, achieving an outstanding \gls{miou} of $80.3$.

\begin{table}[t]
  \begin{minipage}{.6\columnwidth}
      \setlength\tabcolsep{3.3pt}
      \caption{Baseline LiDAR SS methods on \gls{selma} point clouds. Results on the CS label split, removing {sky}.}
        \begin{tabular}{l|c}
              Architecture & \gls{miou} \\\hline
        RangeNet++ \cite{milioto2019icra-fiass} (SqueezeSeg-V2) & 61.9\\
        RangeNet++ \cite{milioto2019icra-fiass} (DarkNet-21) & 67.4\\
        Cylinder3D \cite{zhu2021cylindrical} & \textbf{80.3} \\
        DLV2 \cite{chen2018deeplab} on spherical projections & 57.3 \\ 
        \end{tabular}%
      \label{tab:supervised_backbones_lidar}%
  \end{minipage}
  \qquad
  \begin{minipage}{.3\columnwidth}
  \setlength\tabcolsep{3.3pt}
  \caption{\gls{miou} fusion performance.}
    \begin{tabular}{l|c}
          Method & \gls{miou} \\\hline
          RGB & 68.9 \\
          Grayscale & 68.0 \\
          Depth & 73.4 \\ \hdashline
          RGBD & 72.4\\
          RGBD $@$layer1 & \textbf{74.3}\\
    \end{tabular}%
  \label{tab:fusion}%
\end{minipage}
\end{table}%

\Cref{fig:qualitative_baselines} reports the qualitative results for the best segmentation architectures, i.e., the DeepLab-V2, for the RGB and the depth samples, and Cylinder3D for the point clouds.
Comparing the RGB and depth-based prediction, we can appreciate that the latter offers great improvements in the recognition of far, small and challenging items in the background, such as poles, traffic lights and traffic signs. 
However, the use of geometric information leads to uncertainty in the prediction of traffic lights and signs, which are mixed up in the depth prediction, but not in the RGB one.
On the other hand, looking at the point cloud segmentation, we can appreciate the great overall precision, as expected given the high quantitative score.
Nevertheless, some artifacts are still present due to the prediction based solely on geometric information, e.g., the sidewalk region on the left is partially confused for ground in proximity of the vegetation class. 
Another interesting artifact lies in the geometrical arrangement of the errors.
Due to the intrinsic working principle of Cylinder3D \cite{zhu2021cylindrical}, we observe that most of the errors are propagated along the same radial coordinates.
For instance, we see that the network predicts ground in spite of sidewalk for a few consecutive scans in a couple of regions denoted by light blue rectangles.
Finally, the prediction performance of semantic labels is poor for small classes such as traffic signs or lights, which are often confused for poles.

\subsection{Thematic Subsets}
\label{subsec:thematic}

Then, to highlight the capability of \gls{selma} to incorporate different visual domains, we define $6$ subsets, the so-called \textit{thematic} splits, sampling images with specific daytime or weather condition, as mentioned in \cref{ssec:splits}.

\begin{table}[t]
\centering
    \begin{minipage}{.63\columnwidth}
        \centering
        \includegraphics[trim={0cm 0cm 0cm 0cm}, clip, width=0.83\textwidth]{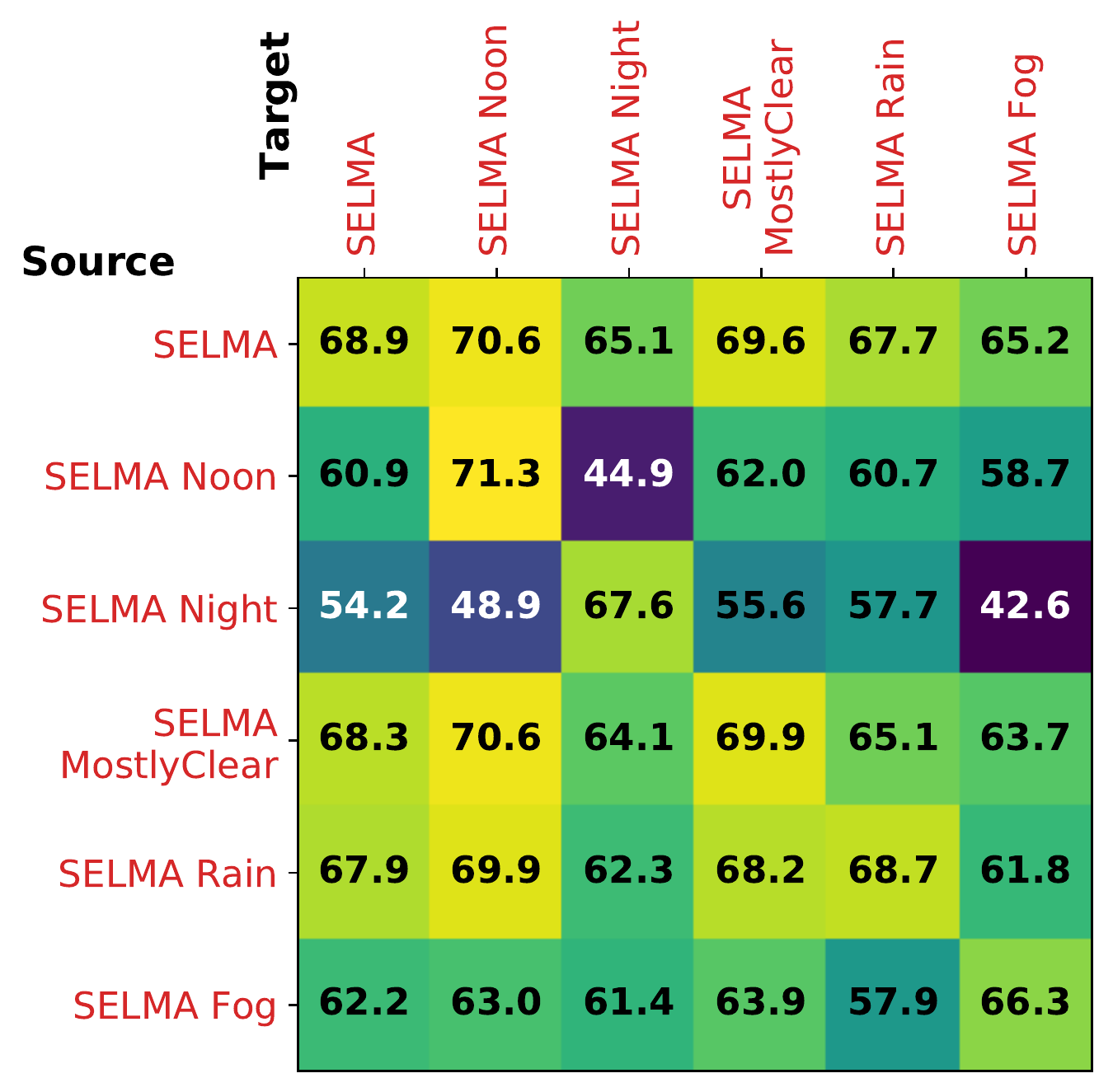}
        \captionof{figure}{\gls{miou} results on thematic splits sub-sampled from the complete SELMA dataset.}
        \label{fig:matrix_thematic}
    \end{minipage}
    \ 
    \begin{minipage}{.33\columnwidth}
          \centering
          \setlength\tabcolsep{3.3pt}
          \caption{\gls{miou} multi-view results. DeepLabV2 \cite{chen2018deeplab} is trained on samples from the desk camera and tested on other cameras.}
            \begin{tabular}{l|c}
                  Target & \gls{miou} \\ \hline
                  Desk & 68.9 \\ \hdashline
                  Front Left & 66.6 \\
                  Front & 66.6 \\
                  Front Right & 66.6 \\
                  Left & 66.2 \\
                  Back & 65.9 \\
                  Right & 62.2
        \end{tabular}%
      \label{tab:desk_to_povs}%
    \end{minipage}
    \vspace{-.7cm}
\end{table}

The \gls{miou} results for the splits are reported in \cref{fig:matrix_thematic}, where we report the supervised accuracy in the diagonal elements and the source-only accuracy (i.e., trained on the source domain and tested on the target domain) on off-diagonal elements.
Here, we can appreciate that training and testing on the same visual domain give almost always the highest \gls{miou} (diagonal elements), except for some cases where the target domain is much easier than the source domain, as is the case for the \gls{selma} Noon as target dataset.

In absolute terms, the hardest subsets (i.e., lowest supervised accuracy) are \gls{selma} Night, Fog, and Rain, respectively.
Adapting the source knowledge acquired from a subset containing a single daytime (e.g., Noon or Night) proves to be less robust to domain variability at test time, rather than adapting knowledge  from a subset containing multiple daytime domains (e.g., Rain or Fog).
Indeed, in the first case we can observe lower off-diagonal accuracy scores compared to the second case. 

\subsection{Fusion Experiments}
\label{subsec:fusion}
To prove the importance of SELMA as a multimodal dataset, we show how we can improve the segmentation quality by coupling acquisitions from different sensors. 
Indeed, different sensors have variable performance depending on the visibility conditions of the scenes, and can be used jointly to leverage understanding scores. To highlight this, we report some experiments in Table~\ref{tab:fusion}.
Notably, using RGB images allows to achieve an \gls{miou} of $68.9$, while using the depth alone could improve the \gls{miou} to $73.4$.
As a comparison, using the grayscale version of the image (single channel) we achieve an \gls{miou} of $68.0$, which is lower than the result on RGB images, as expected.

Building a combined RGB and depth representation at the input level (denoted as RGBD, i.e., an input of $4$ channels), we achieve an \gls{miou} of $72.4$, which is higher than using RGB alone, but lower than using depth alone. Hence, we argue that simply combining the input representations as they are provided is not enough to increase the \gls{ss} accuracy. 
Therefore, we include an additional convolutional layer (followed by batch normalization and ReLU activation function) to extract features from RGB and depth samples separately. Then, we concatenate the outputs and feed the result as the input of the first layer of the ResNet101. We denote this fusion method as ``RGBD $@$layer1" in Table~\ref{tab:fusion}. With this simple provision, we could achieve an \gls{miou} of $74.3$, an improvement of $5.4$ points with respect to using RGB alone, and by $0.9$ points with respect to using depth alone.
We believe that future research could leverage the multimodal design of \gls{selma} to extensively investigate a wide range of fusion strategies for different sensors, such as RGB, depth and~LiDAR.

\subsection{Multi-View Experiments}
\label{subsec:multiview}

As baseline experiments for the multi-view aspect of our dataset, we consider an architecture trained on the desk camera and tested on the available points of view.
This permits to verify the intra-domain shift caused by the variable camera viewpoints, which we expect to be more significant in the cameras facing different directions (like Left, Right and Back). The quantitative results of our experiment are reported in \Cref{tab:desk_to_povs}.
The \gls{miou} scores confirm our expectations, i.e., all front facing cameras have minimal performance degradation ($2.3\%$ \gls{miou}) and are all similar to each other due to their reciprocal proximity (recall \Cref{fig:sensor_setup}). The second best adaptation is achieved by the left-facing camera (with $66.6\%$ \gls{miou}, only $0.4\%$ less than the front cameras), and we argue that this depends on the right-side driving simulated scenario, meaning that the point of view of the left camera is less occluded than its right-facing counterpart, which is much closer to the buildings on the road side. This assumption is verified by looking at the right-facing camera score, which is the lowest among the cameras, with a significant loss of $6.7\%$ \gls{miou} compared to the desk camera.
Finally, the back-facing camera performs similarly to its left-facing counterpart, achieving a score of $65.9\%$ which is only $0.3\%$ lower than the latter.

\section{Experiments on Real-World Datasets}
\label{sec:UDA}
In the last set of experiments we validate our SELMA dataset on different \gls{ss} models. We run an extensive evaluation by training the DeepLab-V2 segmentation architecture on the SELMA dataset and testing it on different real-world datasets.

\subsection{Training with a Mixture of Synthetic and Real Data}
\begin{figure}[t]
    \centering
    \includegraphics[trim={0.7cm 0 0.3cm 1.1cm}, clip, width=\columnwidth]{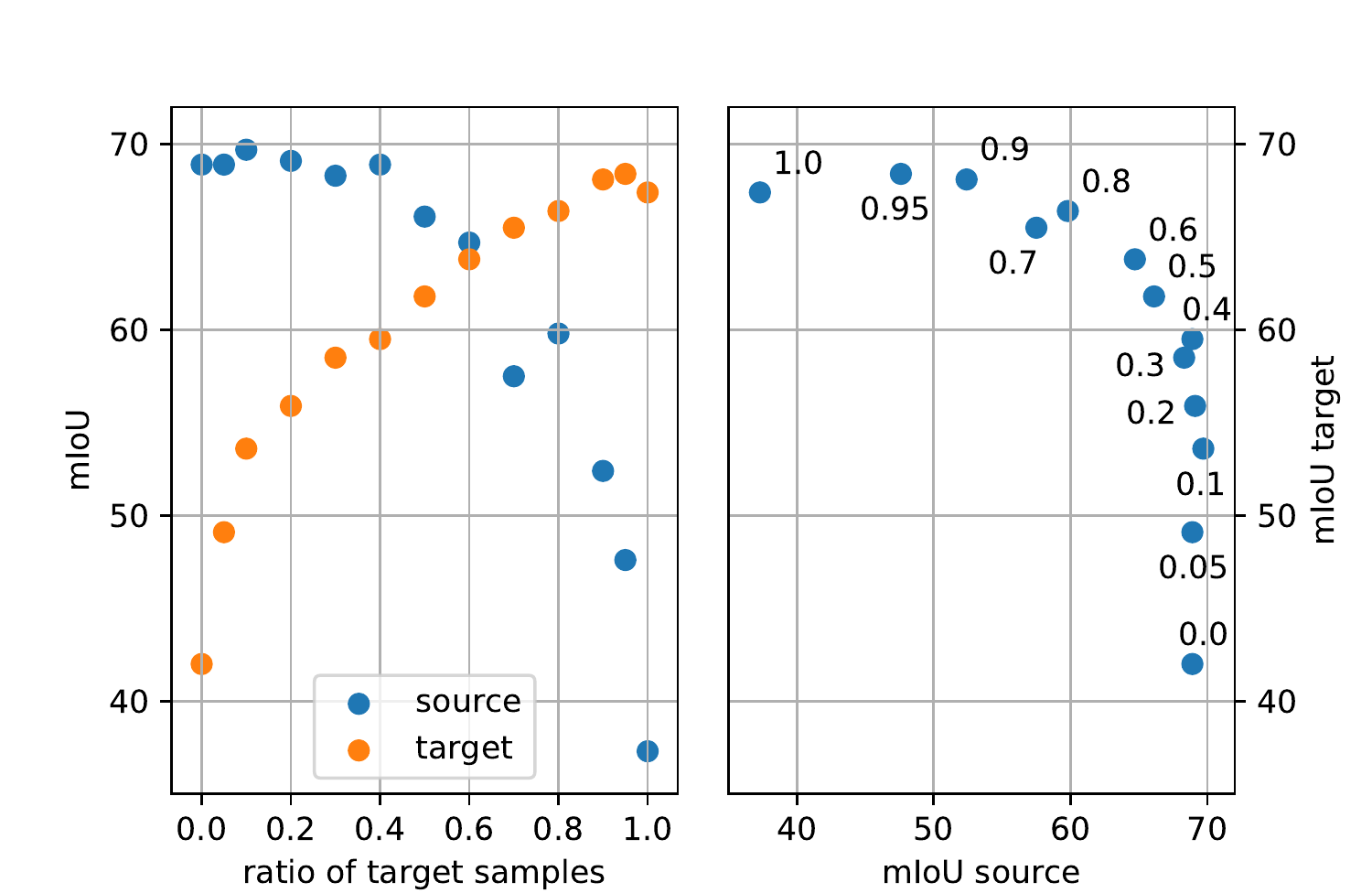}
    \setlength\belowcaptionskip{-.7cm}
    \caption{Accuracy (\gls{miou}) with images sampled either from the Cityscapes dataset (target) or from our SELMA dataset (source).}
    \label{fig:mixed}
\end{figure}

To start, we show in \Cref{fig:mixed} the \gls{miou} accuracy on source and target sets when training the segmentation network on samples drawn either from the target real-world domain, i.e., the Cityscapes dataset (with probability $r$), or from the source domain, i.e., our SELMA dataset (with probability $1-r$). Adding as few as $5\%$ to $10\%$ of data from a different domain improves domain generalization, i.e., the network can perform well on both domains.
Even more, we highlight that a $5\%$ of samples from SELMA can improve the performance on the target domain from $67.4\%$ to $68.4\%$. Similarly, $10\%$ of target samples improve the performance on the source domain from $68.9\%$ to $69.7\%$. The per-class accuracy is reported in the \cref{subsec:results_appendix:supervised}.

To analyze the performance gain when considering imprecise labels, we trained the same architecture using a mixture of SELMA and coarsely-annotated samples from Cityscapes. We achieved an \gls{miou} score of $59.5\%$, which is $7.9\%$ lower than the fully supervised training on Cityscapes, and $2.3\%$ lower than the mixed training score when $r=0.5$. This demonstrates that we can bridge the domain gap with few coarsely-supervised samples in the target domain, thus reducing the cost for accurate labeling.

\subsection{Training with Synthetic Data Only}

\begin{figure}[t]
    \centering
    \includegraphics[trim={0cm 0cm 0cm 0cm}, clip, width=0.9\columnwidth]{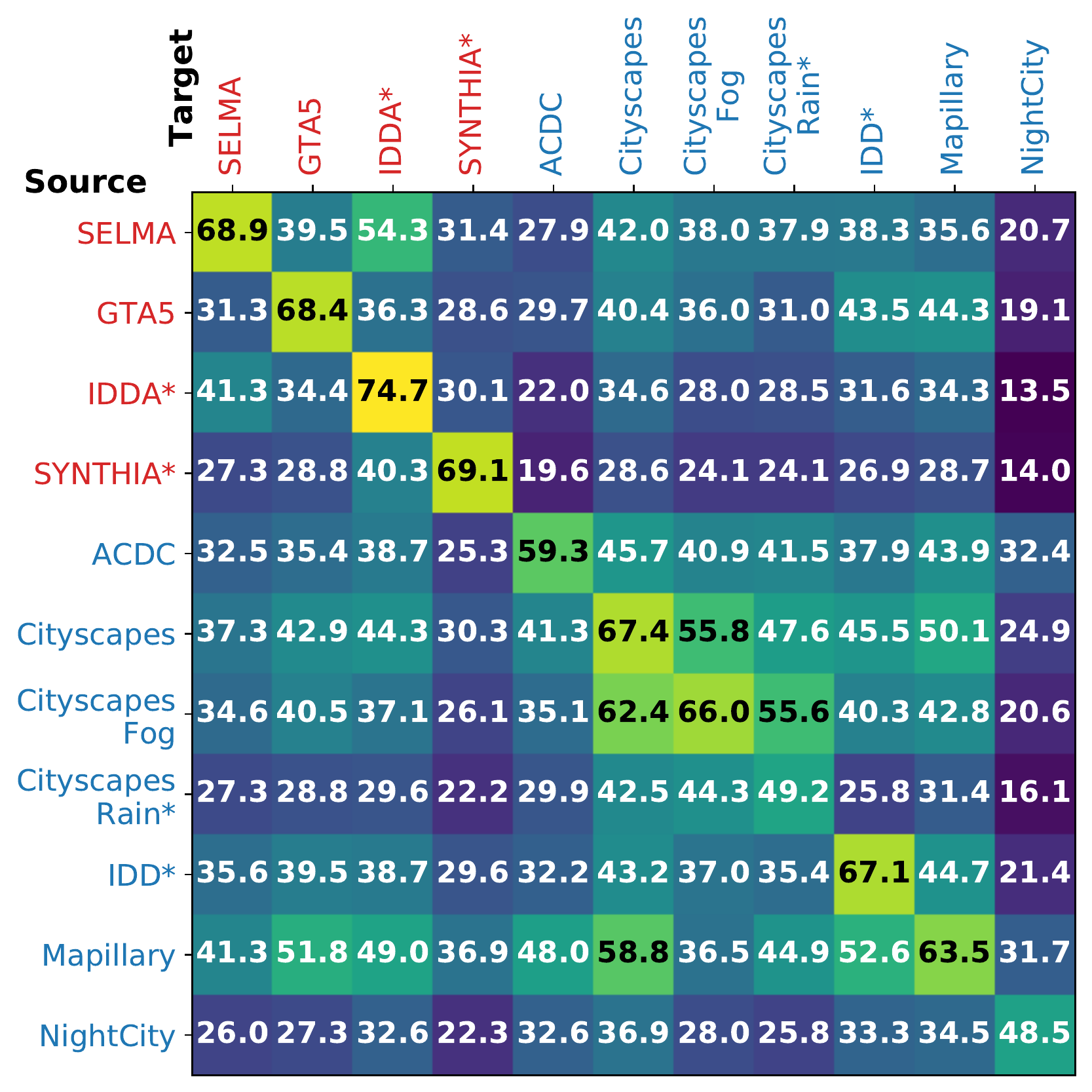}
    \setlength\belowcaptionskip{-.5cm}
    \caption{\gls{miou} performance for different synthetic and real-world datasets. The network is trained on a source domain (rows) and tested on a target domain (columns). Off-diagonal elements correspond to the presence of domain shift. The asterisk (*) indicates that a different label set is employed for testing (see \cref{subsec:results_appendix:supervised} for further details).}
    \label{fig:matrix}
\end{figure}

We then considered an unsupervised setup where no data from the target dataset were used for the training.
We performed an extensive validation in the presence of domain shift, whose results are reported in \Cref{fig:matrix}: we considered four synthetic datasets (in red) and seven real-world datasets (in blue) with extremely variable time and weather conditions. 
We trained a DeepLab-V2 architecture (with ResNet-101 as backbone) on each dataset, and performed the testing on all the domains. 
Values on the diagonal correspond to training and testing performed on the same dataset, i.e., standard supervised training, while values off the diagonal correspond to training on a source domain (on the rows) and testing on a different target domain (on the columns). As expected, the values on the diagonal are larger than the others, since there is no domain shift.

In general, the \gls{miou} performance depends on the source domain where training is performed.
For example, a source training on SELMA performs well on IDDA, and vice-versa, since the rendering engine is common for the two datasets. Also, a source training on Cityscapes performs well on Cityscapes Fog or Cityscapes Rain, and vice-versa, since the city-level domain is common for the two datasets.

On the other side, training on datasets which only account for daytime images, such as Cityscapes, SYNTHIA or IDDA, struggles to generalize to nighttime images, e.g., sampled from the NightCity dataset. Training on statistically variable datasets, such as SELMA or Mapillary, can greatly improve the generalization capabilities in challenging domains.

Furthermore, we can observe that source training on the SELMA dataset outperforms the other synthetic datasets, on the widely-used real-world Cityscapes dataset. Indeed, source knowledge acquired on SELMA transfers well to Cityscapes, achieving $42.0\%$ of \gls{miou}, higher than transferring knowledge from GTA5 ($40.4\%$), IDDA ($22.0\%$) or SYNTHIA ($28.6\%$). Remarkably, SELMA outperforms both GT5A (which is the most popular dataset for this task) by $1.6\%$ of \gls{miou}, and IDDA  (i.e., the most similar dataset based on the same graphic engine) by an outstanding $20\%$ of \gls{miou}.

\begin{figure}[t]
    \centering
    \includegraphics[trim={0cm 0cm 0cm 0cm}, clip, width=\columnwidth]{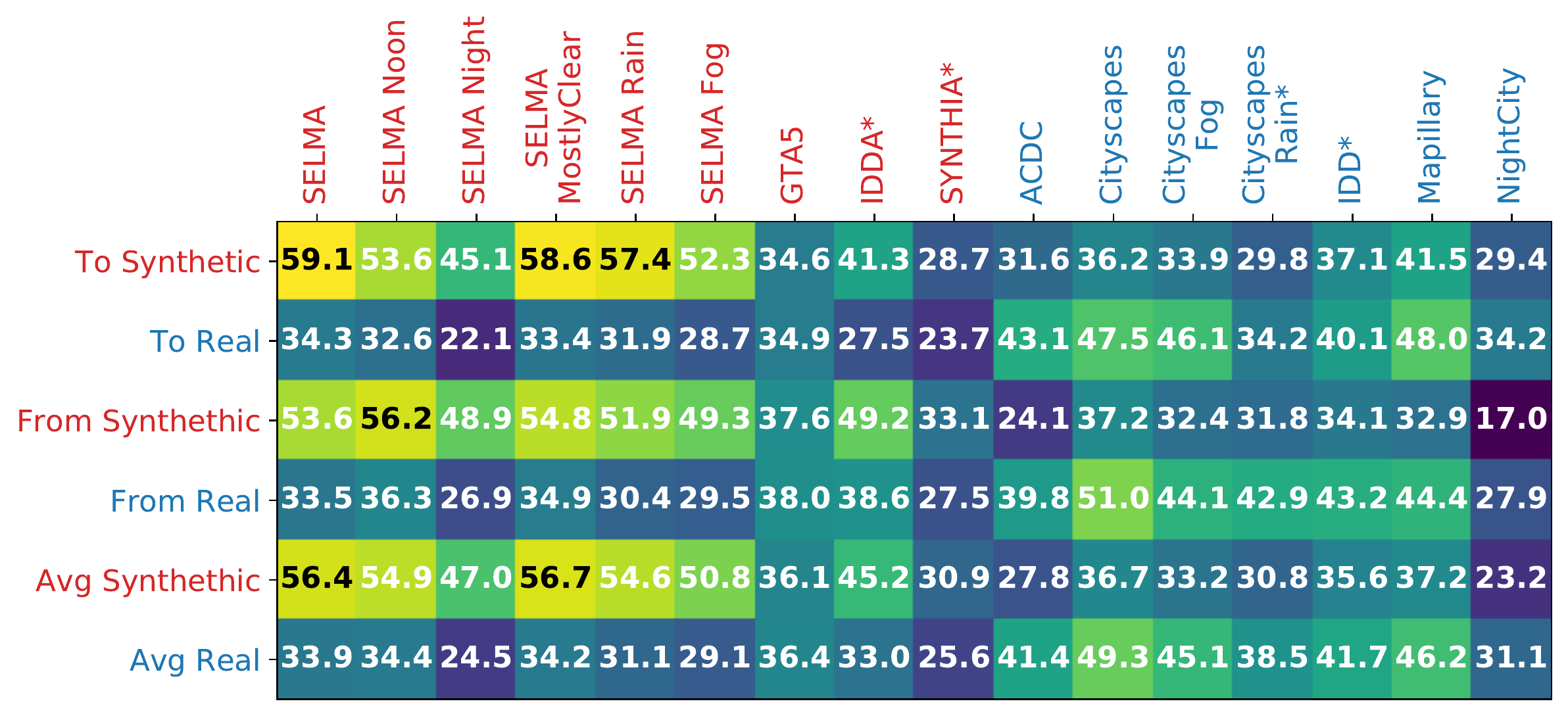}
    \setlength\belowcaptionskip{-.7cm}
    \caption{\gls{miou} performance results taken from the heatmap in \cref{fig:matrix}, and aggregated by synthetic or real domain. The asterisk (*) indicates a different label set (see \cref{subsec:results_appendix:supervised}).}
    \label{fig:matrix_averages}
\end{figure}

Finally, in \Cref{fig:matrix_averages} we show the results averaged according to synthetic or real domains. First, we can appreciate that source domains have better performance on other source domains with respect to target ones, and so do target domains. This is to be attributed to different textures, colors, and brightness rendered by the synthetic graphic engines versus the true target properties of real-world datasets. 
In general, we observe that acquiring source knowledge on our \gls{selma} dataset (or its  subsets) leads to much higher accuracy scores (e.g., $34.3\%$ from \gls{selma}) on both source and target domains, rather than IDDA \cite{alberti2020idda} ($27.5\%$) or SYNTHIA \cite{ros2016synthia} ($23.7\%$) datasets.
Overall, \gls{selma} achieves similar scores than acquiring source knowledge from the baseline GTA5 dataset.

\section{Conclusions}
\label{subsec:conclusions}
In this paper we presented SELMA, a synthetic dataset with driving scenes that contain a large amount of labeled samples acquired considering several different sensors, weather, daytime and viewpoint conditions. 
The experimental evaluation shows that SELMA allows to efficiently train deep learning models for scene understanding tasks in the autonomous driving context, achieving a good generalization to real-world data.
The SELMA dataset is publicly available and can be downloaded for free, in the hope that it will be useful to the scientific community.

The availability of large-scale multimodal acquisitions in variable weather, daytime and viewpoints in SELMA promotes research towards key challenges, for example scene understanding for autonomous driving applications like multimodal sensor exploitation, domain generalization from synthetic datasets to real scenes, and autonomous driving in adverse weather conditions. 

\appendix
In this appendix we include some additional implementation details and some experiments supporting the evaluation of our proposed dataset (SELMA).
In particular, we start by reporting an overview of the CARLA~\cite{Dosovitskiy17} simulator in \cref{sec:carla_appendix}, to better underline the elements that needed modification to support the full scope of SELMA's contributions. Then, \cref{sec:thematic_appendix} present in detail the thematic environmental splits included in SELMA, which are briefly introduced in \cref{ssec:splits}. Finally, we report a more detailed version of the results presented in \cref{sec:experiments}.

\subsection{CARLA Simulator}
\label{sec:carla_appendix}
\paragraph*{Overview}
CARLA is built as a client-server architecture. The server handles the Unreal Engine simulation, i.e., the rendering, the physics computation, the world and actor states. The client offers an \gls{api} (in Python and C++) that handles over part of the control on the simulation to the user.
Specifically, when enabling the synchronous mode, the server waits for a control signal from the client, i.e., a \textit{tick}, before updating the simulation.
Together with a fixed time-step length, this allows to obtain reproducible and reliable physics simulations, as well as realistic and synchronized sensor data.
Namely, the main elements of the Python \gls{api} are
\begin{itemize}
    \item the \textit{Client}, that works as a communication interface to the server. It allows to access and modify the high-level simulation settings, e.g., the loading of the World and of the Traffic Manager, and the connection to the server.
    \item the \textit{World}, that controls the large-scale settings of the environment such as the Weather and the Map, the client-server synchronization, and the life cycles of the Actors.
    \item the \textit{Actors} are \glspl{npc} that can be controlled via the \gls{api}. They include Pedestrians, Vehicles, Sensors, Traffic Signs and Traffic Lights.
\end{itemize}
Specifically, through the \gls{api} it is possible to send commands and meta-commands to the server and receive the sensor readings.
Commands and meta-commands control the Actors (e.g., steering and braking of Vehicles) and the behavior of the server (e.g., resetting the simulation, changing the weather and visibility conditions), respectively.

\renewcommand{\sizefiggg}{0.118}
\begin{figure}[t]
    \centering
    \includegraphics[width=0.95\columnwidth]{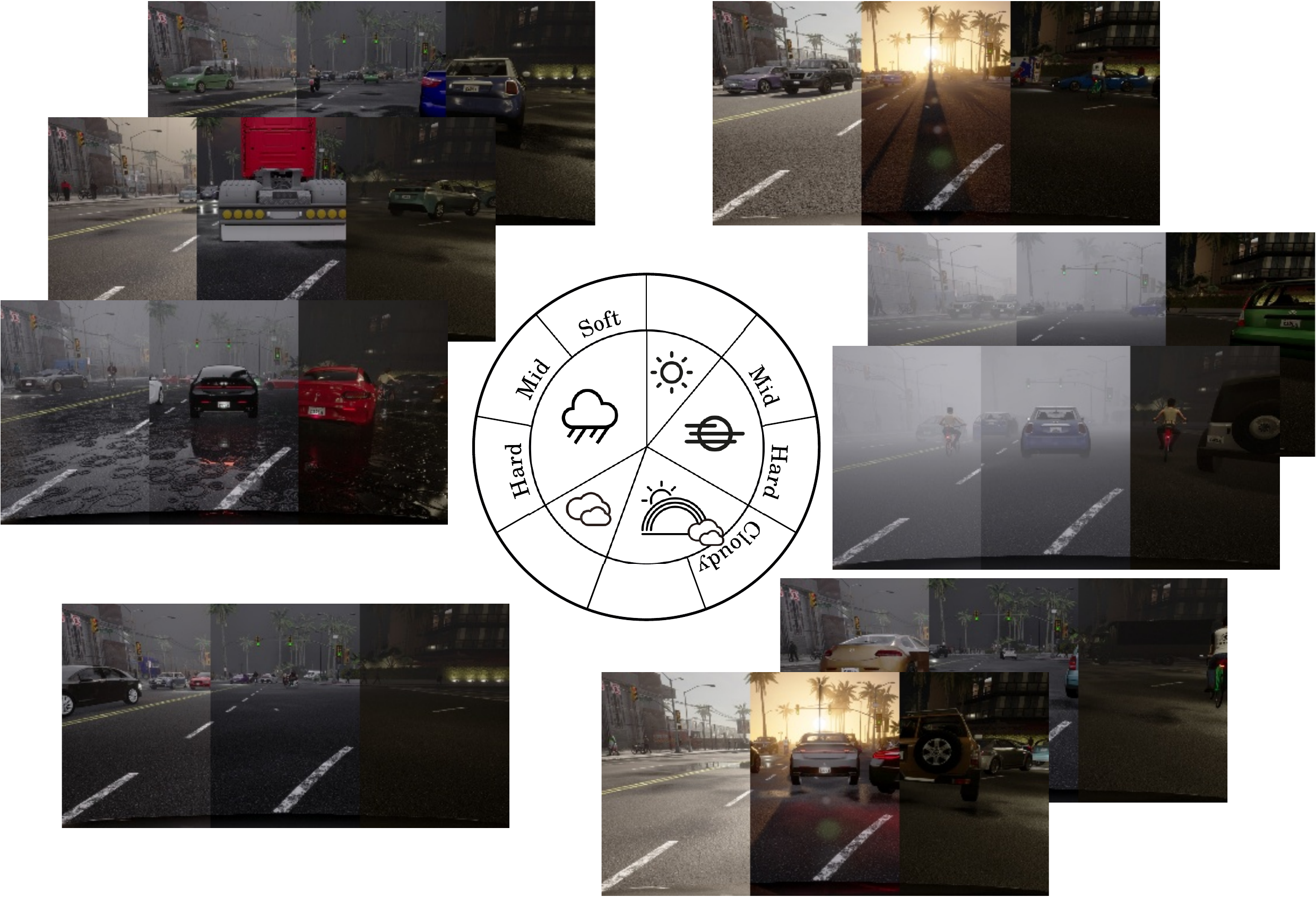}
    \caption{Samples in the $27$ environmental conditions ($9$ weather and $3$ daytime conditions).}
    \label{fig:weather_hour}
\end{figure}

\paragraph*{Sensors}
Namely, there are $14$ sensors available in \gls{carla}: Collision detector, Depth camera, GNSS sensor, IMU sensor, Lane invasion detector, LIDAR sensor, Obstacle Detector, Radar sensor, RGB camera, RSS sensor, Semantic LIDAR sensor, Semantic Segmentation camera, DVS camera, and Optical Flow camera. 

We report here additional details of the sensors used for the creation of the dataset:
\begin{itemize}
    \item \textbf{RGB camera}: the camera offers a classical color view of the environment, as rendered by default in \gls{ue4}. Camera parameters include horizontal field of view, lens aperture, lens distortion, image resolution, gamma value, motion blur, etc. A set of post-processing effects increasing the realism can be enabled with the \texttt{enable\_postprocess\_effects} flag: vignetting, grain jitter, bloom, auto exposure, lens flare and depth of field.
    \item \textbf{Depth camera}: each pixel of the image encodes the distance of the captured object from the camera (also known as depth buffer or z-buffer) to create a depth map of the environment in the \gls{fov} of the sensor.
    \item \textbf{Semantic segmentation camera}: every pixel of the image is labelled with the identifier of the object class it represents. When created, each object is tagged with its class identifier, thus making the segmentation exact and effortless, a key advantage over real datasets. This data is used as ground truth when training the segmentation models.
    \item \textbf{Semantic LiDAR}: LiDARs can produce a \gls{3d} representation of the surrounding environment in the form of a point cloud. Generally, information encoded in each point includes its \gls{3d} coordinates and the back-scattered intensity. The \gls{carla} semantic LIDAR does not capture intensity, but registers the semantic ground-truth of the object hit by the raycast. 
    From the \gls{api}, it is possible to set the number of channels $N_c$, the range, the vertical and horizontal \gls{fov}, the rotation frequency and the number of points per second $P_s$. The rotation is simulated computing the horizontal roatation angle of the LIDAR  in a frame. The point cloud is generated through ray-casting at each time step, thus obtaining a number of points per channel per step $P$ equal to $P = P_s / (\textrm{FPS} * N_c)$. A point cloud contains the points generated in a 1/FPS interval, during which the physics are not updated. Thus, all the points in a point cloud capture the same, static representation of the scene.
\end{itemize}

\paragraph*{Towns}
For convenience, we report here the town description that can be found in the \gls{carla} documentation:
\begin{itemize}
    \item \textbf{Town01}: A basic town layout with "T junctions" between roads.
    \item \textbf{Town02}: Similar to Town01, but smaller.
    \item \textbf{Town03}: The most complex town, with a 5-lane junction, a roundabout, unevenness, a tunnel, and more.
    \item \textbf{Town04}: An infinite loop with a highway and a small town.
    \item \textbf{Town05}: Squared-grid town with cross junctions and a bridge. It has multiple lanes per direction. Useful to deal with lane changes.
    \item \textbf{Town06}: Long highways with many highway entrances and exits. It also has a Michigan left.
    \item \textbf{Town07}: A rural environment with narrow roads, barns and hardly any traffic lights.
    \item \textbf{Town10HD}: A city environment with different environments such as an avenue or promenade, and more realistic textures.
\end{itemize}
Besides the \gls{ue4} model files, the OpenDrive and a pointcloud representation are available. Specifically, the OpenDrive file contains all the information on the road topology, that can be used by the \gls{carla} \gls{api} to generate \textit{Waypoints}, i.e., 3D directed points with attributes associated to OpenDrive properties.

\subsection{Thematic Subsets of SELMA}
\label{sec:thematic_appendix}

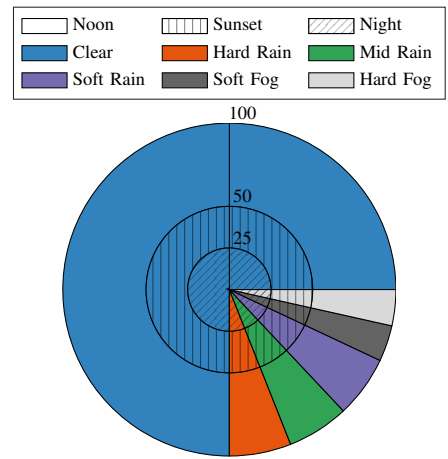
\begin{figure}[t]
    \centering        
    \setlength\fwidth{0.5\columnwidth} 
    \setlength\fheight{0.5\columnwidth}
    \input{imgs/design/split_pie.tex}
    \caption{Distributions of weather conditions and daytime in the SELMA-Rand split. }
    \label{fig:split}
\end{figure}

Six \textit{thematic} SELMA splits were designed sampling images from specific daytime or weather conditions.
The splits are composed as follows:
\begin{itemize}
\item \textbf{SELMA}: with a slight abuse of notation, this split refers to the random split, whose probability densities are defined in \cref{ssec:splits} and summarized in \cref{fig:split}. 
\item \textbf{Noon} and \textbf{Night} splits share the same weather distribution defined for the random split, but the daytime is sampled from Noon and Night samples, respectively.
\item \textbf{MostlyClear} split shares the same daytime distribution of the random split, whereas the weather samples on non-rainy and non-foggy images with a larger skew towards clear weather (${P[\mathrm{Clear}]=0.25,}$ ${P[\mathrm{Cloudy}]=0.25,}$ ${P[\mathrm{WetCloudy}]=0.25,}$ ${P[\mathrm{Wet}]=0.25}$).
\item \textbf{Rain} split shares the same daytime distribution defined for the random split, but the weather is sampled from a rainy scenes (${P[\mathrm{SoftRain}]=0.34,}$ ${P[\mathrm{MidRain}]=0.33,}$ ${P[\mathrm{HardRain}]=0.34}$). 
\item \textbf{Fog} split shares the same daytime distribution defined for the random split, but the weather is sampled from the foggy scenes (${P[\mathrm{MidFog}]=0.50,}$ ${P[\mathrm{HardFog}]=0.5}$). 
\end{itemize}

\subsection{Additional Experimental Results}
\label{sec:results_appendix}


\begin{table*}[htbp]
  \centering
  \setlength\tabcolsep{3.75pt} 
  \caption{Per-class IoU of baseline semantic segmentation architectures on subsets of our dataset for both RGB images (first 6 blocks) or depth (last block).}
    \begin{tabular}{|c|c|ccccccccccccccccccc|c|}
    \hline
     & & \rotatebox{90}{road}  & \rotatebox{90}{swalk} & \rotatebox{90}{building} & \rotatebox{90}{wall}  & \rotatebox{90}{fence} & \rotatebox{90}{pole}  & \rotatebox{90}{tlight} & \rotatebox{90}{tsign} & \rotatebox{90}{veg}   & \rotatebox{90}{terrain} & \rotatebox{90}{sky}   & \rotatebox{90}{person} & \rotatebox{90}{rider} & \rotatebox{90}{car}   & \rotatebox{90}{truck} & \rotatebox{90}{bus}   & \rotatebox{90}{train} & \rotatebox{90}{mbike} & \rotatebox{90}{bike}  & \rotatebox{90}{mIoU} \\\hline     
\multirow{5}[4]{*}{\rotatebox{90}{\quad SELMA}} & DeeplabV2 \cite{chen2018deeplab} & 99.1  & 87.5  & 83.9  & 83.5  & 57.3  & 47.1  & 36.8  & 61.1  & 78.4  & 77.2  & 89.4  & 63.9  & 62.8  & 88.7  & 71.6  & 48.9  & 72.3  & 62.0  & 37.7  & 68.9 \\
          & DeeplabV3 \cite{chen2017rethinking} & 99.2  & 88.1  & 84.3  & 85.6  & 57.7  & 48.0  & 38.4  & 62.0  & 78.6  & 77.7  & 89.5  & 64.2  & 63.7  & 91.4  & 73.4  & 69.8  & 70.9  & 63.0  & 37.2  & 70.7 \\
          & FCN \cite{long2015} & 99.1  & 87.6  & 83.6  & 81.5  & 56.8  & 47.1  & 35.3  & 60.6  & 78.3  & 76.9  & 89.6  & 63.5  & 62.0  & 89.0  & 66.3  & 51.3  & 69.0  & 60.8  & 37.7  & 68.2 \\
          & PSPNet \cite{zhao2017} & 99.1  & 87.6  & 83.6  & 83.3  & 56.8  & 47.0  & 35.7  & 60.7  & 77.9  & 76.6  & 89.5  & 63.4  & 62.2  & 89.4  & 62.3  & 56.8  & 70.7  & 61.0  & 36.9  & 68.4 \\
          & UNet \cite{ronneberger2015u} & 94.8  & 71.5  & 64.2  & 63.8  & 35.5  & 26.6  & 0.0   & 0.0   & 61.5  & 61.6  & 79.3  & 33.1  & 25.3  & 70.0  & 0.0   & 0.0   & 0.0   & 0.0   & 0.0   & 36.2 \\
    \hdashline
    \multirow{5}[2]{*}{\rotatebox{90}{Noon}} & DeeplabV2 \cite{chen2018deeplab} & 99.3  & 88.9  & 85.2  & 86.7  & 59.8  & 49.0  & 40.1  & 62.1  & 82.7  & 83.3  & 92.9  & 65.2  & 63.4  & 90.9  & 73.6  & 70.1  & 76.3  & 66.2  & 37.8  & 72.3 \\
          & DeeplabV3 \cite{chen2017rethinking} & 99.0  & 87.8  & 87.0  & 86.7  & 60.2  & 50.0  & 42.1  & 63.2  & 82.8  & 83.1  & 92.9  & 65.8  & 64.0  & 90.9  & 74.6  & 63.5  & 72.2  & 66.0  & 31.1  & 71.7 \\
          & FCN \cite{long2015} & 99.2  & 88.9  & 86.7  & 86.1  & 59.8  & 49.3  & 38.5  & 62.0  & 82.9  & 83.3  & 93.0  & 65.6  & 62.9  & 90.1  & 70.5  & 57.1  & 71.2  & 65.4  & 37.7  & 71.1 \\
          & PSPNet \cite{zhao2017} & 99.3  & 88.9  & 86.7  & 86.5  & 59.8  & 49.3  & 39.0  & 62.0  & 83.0  & 83.3  & 93.1  & 65.5  & 63.0  & 90.0  & 74.5  & 51.9  & 73.9  & 64.9  & 37.9  & 71.2 \\
          & UNet \cite{ronneberger2015u} & 98.5  & 80.6  & 71.5  & 71.4  & 41.8  & 30.5  & 0.0   & 0.0   & 76.6  & 75.3  & 90.7  & 39.4  & 30.4  & 82.4  & 3.8   & 0.0   & 0.0   & 0.0   & 0.7   & 41.8 \\
    \hdashline
    \multirow{5}[2]{*}{\rotatebox{90}{Night}} & DeeplabV2 \cite{chen2018deeplab} & 99.0  & 86.5  & 79.7  & 82.2  & 54.3  & 46.6  & 35.8  & 61.1  & 70.9  & 67.0  & 84.4  & 62.2  & 62.0  & 90.4  & 78.2  & 67.2  & 71.1  & 60.8  & 35.6  & 68.2 \\
          & DeeplabV3 \cite{chen2017rethinking} & 99.0  & 86.7  & 80.2  & 83.3  & 55.3  & 46.8  & 37.1  & 61.4  & 71.2  & 67.2  & 84.0  & 62.8  & 62.9  & 90.7  & 74.6  & 69.2  & 66.4  & 62.4  & 38.9  & 68.4 \\
          & FCN \cite{long2015} & 99.0  & 86.6  & 79.6  & 81.0  & 54.1  & 45.3  & 34.0  & 60.3  & 70.6  & 67.2  & 84.0  & 62.1  & 61.4  & 89.3  & 67.8  & 51.3  & 66.7  & 59.2  & 35.9  & 66.1 \\
          & PSPNet \cite{zhao2017} & 99.1  & 86.8  & 79.6  & 81.9  & 54.3  & 46.0  & 35.1  & 60.3  & 70.6  & 67.2  & 84.2  & 62.0  & 61.7  & 90.3  & 70.9  & 61.3  & 68.8  & 60.5  & 36.6  & 67.2 \\
          & UNet \cite{ronneberger2015u} & 97.8  & 76.4  & 59.7  & 60.2  & 34.7  & 29.7  & 0.0   & 0.0   & 55.4  & 48.0  & 76.4  & 35.2  & 28.1  & 77.1  & 0.1   & 0.0   & 0.0   & 0.0   & 0.0   & 35.7 \\
    \hdashline
    \multirow{5}[2]{*}{\rotatebox{90}{MostlyClear}} & DeeplabV2 \cite{chen2018deeplab} & 99.2  & 87.8  & 84.8  & 85.3  & 57.3  & 48.2  & 36.3  & 61.2  & 79.5  & 78.6  & 91.1  & 64.7  & 63.5  & 88.8  & 62.6  & 61.2  & 73.4  & 66.8  & 37.9  & 69.9 \\
          & DeeplabV3 \cite{chen2017rethinking} & 99.2  & 88.6  & 85.4  & 85.4  & 58.1  & 49.2  & 38.7  & 61.8  & 80.0  & 79.3  & 90.9  & 65.1  & 63.8  & 90.8  & 61.5  & 70.1  & 72.5  & 67.1  & 38.6  & 70.8 \\
          & FCN \cite{long2015} & 99.2  & 88.0  & 85.1  & 84.1  & 57.3  & 48.0  & 35.4  & 60.9  & 79.9  & 79.2  & 91.2  & 64.6  & 62.5  & 89.0  & 62.2  & 50.8  & 71.8  & 65.9  & 37.9  & 69.1 \\
          & PSPNet \cite{zhao2017} & 99.2  & 88.1  & 85.0  & 84.2  & 57.1  & 47.7  & 36.5  & 60.9  & 79.7  & 78.8  & 91.1  & 64.5  & 63.0  & 89.6  & 63.5  & 62.3  & 72.9  & 65.9  & 36.8  & 69.8 \\
          & UNet \cite{ronneberger2015u} & 94.9  & 75.1  & 65.8  & 58.8  & 34.8  & 27.6  & 0.0   & 0.5   & 61.4  & 62.8  & 77.1  & 35.3  & 25.4  & 66.7  & 2.0   & 0.0   & 0.0   & 0.0   & 0.8   & 36.3 \\
    \hdashline
    \multirow{5}[2]{*}{\rotatebox{90}{Fog}} & DeeplabV2 \cite{chen2018deeplab} & 99.2  & 88.4  & 77.4  & 81.8  & 54.5  & 46.0  & 38.9  & 61.7  & 69.0  & 66.5  & 81.7  & 63.2  & 63.7  & 91.0  & 70.0  & 67.2  & 69.6  & 65.4  & 36.7  & 68.0 \\
          & DeeplabV3 \cite{chen2017rethinking} & 99.2  & 89.0  & 77.3  & 82.7  & 55.7  & 47.2  & 42.1  & 62.0  & 69.6  & 66.0  & 81.0  & 64.1  & 63.9  & 92.0  & 67.7  & 79.6  & 67.2  & 66.1  & 39.3  & 69.0 \\
          & FCN \cite{long2015} & 99.2  & 88.5  & 76.7  & 79.1  & 55.1  & 46.5  & 38.6  & 60.8  & 69.3  & 65.9  & 81.4  & 63.3  & 62.8  & 85.8  & 58.7  & 34.7  & 57.2  & 64.4  & 37.9  & 64.5 \\
          & PSPNet \cite{zhao2017} & 99.2  & 88.5  & 77.3  & 78.4  & 54.6  & 46.7  & 38.8  & 61.3  & 69.2  & 66.0  & 81.7  & 62.8  & 63.0  & 89.6  & 59.5  & 62.5  & 66.9  & 65.5  & 37.4  & 66.8 \\
          & UNet \cite{ronneberger2015u} & 85.6  & 75.4  & 40.8  & 35.6  & 29.8  & 22.0  & 0.0   & 0.7   & 39.7  & 40.1  & 61.8  & 25.8  & 30.0  & 52.8  & 0.1   & 0.0   & 0.0   & 0.0   & 3.2   & 28.6 \\
    \hdashline
    \multirow{5}[2]{*}{\rotatebox{90}{Rain}} & DeeplabV2 \cite{chen2018deeplab} & 99.1  & 86.4  & 82.5  & 84.1  & 57.3  & 46.4  & 38.5  & 62.0  & 77.1  & 73.0  & 88.5  & 64.1  & 63.4  & 89.3  & 68.1  & 55.0  & 70.5  & 64.1  & 36.0  & 68.7 \\
          & DeeplabV3 \cite{chen2017rethinking} & 99.0  & 86.5  & 83.1  & 84.9  & 58.0  & 46.9  & 39.8  & 62.5  & 77.5  & 75.4  & 88.3  & 64.5  & 63.6  & 89.0  & 53.0  & 46.5  & 67.2  & 64.6  & 37.5  & 67.8 \\
          & FCN \cite{long2015} & 99.0  & 86.3  & 82.4  & 83.0  & 57.3  & 46.9  & 37.0  & 61.6  & 77.1  & 74.4  & 88.5  & 64.2  & 62.6  & 86.8  & 44.1  & 52.7  & 64.9  & 63.9  & 36.1  & 66.8 \\
          & PSPNet \cite{zhao2017} & 99.1  & 86.5  & 82.3  & 83.2  & 57.3  & 46.5  & 37.4  & 61.6  & 76.9  & 73.7  & 87.9  & 64.0  & 62.9  & 89.7  & 66.7  & 63.4  & 72.7  & 64.2  & 35.5  & 69.0 \\
          & UNet \cite{ronneberger2015u} & 97.1  & 72.1  & 64.1  & 66.1  & 38.5  & 27.8  & 0.0   & 8.6   & 64.6  & 59.4  & 82.6  & 35.8  & 25.6  & 74.6  & 1.2   & 0.0   & 0.0   & 0.0   & 0.0   & 37.8 \\
    \hline
    \multirow{5}[2]{*}{\rotatebox{90}{Depth}} & DeeplabV2 \cite{chen2018deeplab} & 98.9  & 85.2  & 90.4  & 86.9  & 63.1  & 57.1  & 41.5  & 69.8  & 84.9  & 79.2  & 94.6  & 71.6  & 67.0  & 93.9  & 75.0  & 57.8  & 70.2  & 64.8  & 41.8  & 73.4 \\
          & DeeplabV3 \cite{chen2017rethinking} & 99.3 & 87.7 & 90.7 & 88.2 & 64.3 & 56.9 & 42.3 & 71.4 & 85.1 & 80.9 & 94.8 & 65.5 & 69.0 & 90.5 & 75.8 & 32.2 & 75.9 & 67.2 & 43.4 & 72.7 \\
          & FCN \cite{long2015} & 99.2 & 86.9 & 90.5 & 86.9 & 63.0 & 56.3 & 38.5 & 69.4 & 85.1 & 79.7 & 94.8 & 72.2 & 66.5 & 94.4 & 69.6 & 68.0 & 72.9 & 65.0 & 42.2 & 73.7 \\
          & PSPNet \cite{zhao2017} & 99.2 & 85.4 & 90.4 & 86.7 & 63.2 & 56.8 & 37.2 & 69.2 & 85.1 & 78.7 & 94.6 & 71.9 & 66.6 & 94.4 & 73.3 & 66.7 & 69.8 & 65.7 & 42.8 & 73.6 \\
          & UNet \cite{ronneberger2015u} & 5.8 & 19.2 & 71.2 & 37.3 & 56.3 & 60.8 & 0.5 & 46.8 & 54.0 & 19.8 & 21.2 & 55.7 & 44.6 & 16.0 & 3.8 & 0.4 & 13.6 & 8.6 & 12.0 & 28.8 \\
    \hline
    \end{tabular}%
  \label{tab:supervised_backbones_per_class}%
\end{table*}%

\begin{table*}[htbp]
\setlength\tabcolsep{1.75pt} 
  \centering
  \caption{Per-class and mean IoU results for the supervised training and testing on the same domain with the DeepLab-V2 architecture. The mean is computed over different label sets.}
    \begin{tabular}{|l|ccccccccccccccccccc|cccccc|}
    \hline
          & \multicolumn{19}{c|}{per-class IoUs} & \multicolumn{6}{c|}{mIoUs on different label sets} \\\hline
          & \rotatebox{90}{road}  & \rotatebox{90}{swalk} & \rotatebox{90}{building} & \rotatebox{90}{wall}  & \rotatebox{90}{fence} & \rotatebox{90}{pole}  & \rotatebox{90}{tlight} & \rotatebox{90}{tsign} & \rotatebox{90}{veg}   & \rotatebox{90}{terrain} & \rotatebox{90}{sky}   & \rotatebox{90}{person} & \rotatebox{90}{rider} & \rotatebox{90}{car}   & \rotatebox{90}{truck} & \rotatebox{90}{bus}   & \rotatebox{90}{train} & \rotatebox{90}{mbike} & \rotatebox{90}{bike}  & \rotatebox{90}{city19} & \rotatebox{90}{idda16} & \rotatebox{90}{synthia16} & \rotatebox{90}{synthia13} & \rotatebox{90}{idda-synthia-15} & \rotatebox{90}{idda-synthia-12} \\\hline
    SELMA & 99.1  & 87.5  & 83.9  & 83.5  & 57.3  & 47.1  & 36.8  & 61.1  & 78.4  & 77.2  & 89.4  & 63.9  & 62.8  & 88.7  & 71.6  & 48.9  & 72.3  & 62.0  & 37.7  & 68.9  & 69.8  & 68.0  & 69.2  & 69.3  & 70.9 \\
    Noon & 99.3  & 88.9  & 86.7  & 86.7  & 59.9  & 49.1  & 40.1  & 61.8  & 82.8  & 83.3  & 92.8  & 65.5  & 63.3  & 89.5  & 76.8  & 48.8  & 75.0  & 65.7  & 37.7  & 71.2  & 72.1  & 69.9  & 71.0  & 71.3  & 72.8 \\
    Night & 99.0  & 86.3  & 80.0  & 82.2  & 53.6  & 45.9  & 36.0  & 60.8  & 71.1  & 67.4  & 84.1  & 62.0  & 62.1  & 90.1  & 78.0  & 57.7  & 71.3  & 60.9  & 34.9  & 67.5  & 67.3  & 66.7  & 68.1  & 67.3  & 68.9 \\
    Mostly Clear & 99.2  & 88.1  & 85.3  & 85.1  & 57.7  & 48.0  & 37.4  & 61.1  & 79.9  & 79.3  & 91.0  & 64.7  & 63.5  & 89.0  & 65.5  & 54.6  & 73.5  & 66.8  & 38.5  & 69.9  & 70.9  & 69.4  & 70.7  & 70.4  & 72.0 \\
    Rain & 99.0  & 86.3  & 82.9  & 84.0  & 57.5  & 46.6  & 38.6  & 62.0  & 77.5  & 74.8  & 88.0  & 64.5  & 63.5  & 88.7  & 65.1  & 55.1  & 71.8  & 64.0  & 36.2  & 68.7  & 69.6  & 68.4  & 69.7  & 69.3  & 70.9 \\
    Fog & 99.2  & 88.6  & 77.3  & 80.6  & 55.0  & 46.4  & 40.0  & 61.6  & 69.7  & 66.9  & 81.8  & 63.6  & 63.4  & 88.3  & 57.4  & 47.5  & 69.6  & 65.3  & 37.8  & 66.3  & 67.8  & 66.6  & 68.0  & 67.9  & 69.7 \\\hdashline
    GTA5~\cite{Richter2016}& 95.8  & 83.1  & 86.9  & 59.4  & 45.4  & 52.1  & 51.7  & 50.3  & 80.8  & 68.6  & 95.1  & 67.9  & 44.1  & 88.6  & 85.2  & 84.6  & 72.8  & 56.3  & 31.6  & 68.4  & 66.1  & 67.1  & 70.5  & 65.9  & 69.4 \\
    IDDA~\cite{alberti2020idda}  & 98.7  & 92.1  & 93.2  & 85.3  & 68.8  & 59.2  & 65.0  & 60.2  & 84.8  & 84.3  & 97.4  & 57.9  & 54.2  & 97.8  & -     & -     & -     & 65.3  & 30.6  & 62.9  & 74.7  & 69.4  & 69.0  & 74.0  & 74.8 \\
    SYNTHIA~\cite{ros2016}& 92.2  & 90.9  & 92.3  & 79.6  & 59.0  & 54.4  & 25.0  & 42.1  & 79.6  & -     & 94.6  & 74.6  & 49.8  & 87.2  & -     & 87.5  & -     & 63.4  & 33.0  & 58.2  & 63.6  & 69.1  & 70.2  & 67.8  & 68.7 \\\hline
    ACDC~\cite{sakaridis2021acdc}  & 92.7  & 71.3  & 82.6  & 46.5  & 37.2  & 50.5  & 61.1  & 49.3  & 83.9  & 44.5  & 94.6  & 49.0  & 17.9  & 80.8  & 41.0  & 71.1  & 81.7  & 29.5  & 41.7  & 59.3  & 58.3  & 60.0  & 63.5  & 59.2  & 62.9 \\
    CityScapes~\cite{Cordts2016}& 97.1  & 77.4  & 89.2  & 50.1  & 46.2  & 45.2  & 48.9  & 61.4  & 89.5  & 55.2  & 92.2  & 69.9  & 46.5  & 91.4  & 66.9  & 75.2  & 60.3  & 52.2  & 65.8  & 67.4  & 67.4  & 68.6  & 73.6  & 68.2  & 73.5 \\
    Cityscapes Fog~\cite{sakaridis2018semantic}& 97.1  & 77.2  & 87.9  & 47.7  & 46.5  & 44.9  & 50.4  & 59.7  & 88.5  & 54.3  & 86.6  & 69.2  & 47.4  & 90.6  & 59.2  & 73.0  & 59.7  & 50.3  & 64.6  & 66.0  & 66.4  & 67.6  & 72.5  & 67.2  & 72.5 \\
    Cityscapes Rain~\cite{hu2019depth}& 97.6  & 77.3  & 85.1  & 25.9  & 5.1   & 29.4  & 25.4  & 35.9  & 87.4  & 61.1  & 93.9  & 49.7  & 33.4  & 85.1  & 0.2   & 25.2  & -     & 8.3   & 60.2  & 46.6  & 53.8  & 51.6  & 58.8  & 53.3  & 61.6 \\
    IDD~\cite{varma2019idd}& 96.8  & 72.2  & 73.4  & 70.0  & 35.2  & 45.2  & 9.9   & 70.4  & 87.8  & -     & 94.9  & 63.8  & 69.1  & 88.5  & 83.5  & 87.5  & -     & 65.2  & 26.8  & 60.0  & 60.6  & 66.0  & 69.7  & 64.6  & 68.2 \\
    Mapillary~\cite{neuhold2017mapillary}& 92.8  & 59.7  & 85.5  & 47.7  & 52.5  & 49.4  & 53.7  & 65.3  & 88.3  & 64.2  & 97.7  & 66.9  & 44.3  & 88.6  & 58.8  & 63.4  & 21.5  & 50.2  & 56.7  & 63.5  & 66.5  & 66.4  & 70.2  & 66.6  & 70.8 \\
    NightCity~\cite{tan2021night}& 89.7  & 42.2  & 81.3  & 44.2  & 49.0  & 25.7  & 20.4  & 47.5  & 58.5  & 25.2  & 86.3  & 47.5  & 27.1  & 78.6  & 52.4  & 62.1  & 27.6  & 25.2  & 30.1  & 48.5  & 48.7  & 51.0  & 53.6  & 50.2  & 52.9 \\\hline
    \end{tabular}%
  \label{tab:supervised_deeplabv2}%
\end{table*}%

\subsubsection{Supervised Learning}
\label{subsec:results_appendix:supervised}
In this paragraph, we report the per-class IoU results of the baseline semantic segmentation architectures.

\textbf{RGB.} Table~\ref{tab:supervised_backbones_per_class} reports the per-class performance of the considered baseline semantic segmentation architectures: UNet \cite{ronneberger2015u}, FCN \cite{long2015}, PSPNet \cite{zhao2017}, DeepLab-V2 \cite{chen2018deeplab,chen2018encoder}, DeepLab-V3 \cite{chen2017rethinking}.
The highest accuracy is obtained with the SELMA Noon split, as the RGB images acquired under favorable weather conditions are easier to segment. On the contrary, the presence of night, fog and rain, degrades the accuracy of the models. 
As we can observe, our results are more balanced across the classes with respect to the Cityscapes split.

Overall, DeepLab-V2 and DeepLab-V3 achieve the best performance.
For the sake of the perfomance-complexity trade-off we decided to employ DeepLab-V2 for all the other experiments.
In Table~\ref{tab:supervised_deeplabv2} we show the per-class IoU scores training DeepLab-V2 on our datasets (first group), on common synthetic benchmarks (second group), and on  real-world datasets (last group). For each dataset, we evaluate the trained model on the labels splits most commonly considered in the literature. Table~\ref{tab:label_sets} reports the names of the classes of each label set. The most widely used label set is City19, which is also the most complete.
For the sake of fairness with respect of all the datasets, in Table~\ref{tab:supervised_deeplabv2} we test our models on all the possible label sets. 
Our SELMA demonstrates par, or often higher, performance compared to other synthetic and real benchmarks due to its extreme variability in daytime and weather conditions, as well as its large-scale property with more than $30,000$ unique (labeled) samples.

\textbf{Depth.} Then, we run the same experimental evaluation using a single input channel representing the depth of the scene. We observe that the best performing architectures offer comparable performance and that the final mIoU is higher than mIoU obtained on RGB samples, since we have true (synthetic) depth maps at disposal. The results are reported as last block of \cref{tab:supervised_backbones_per_class}.

\begin{table*}[t]
  \setlength\tabcolsep{3pt}  
  \centering
  \caption{Per-class and mean IoU results for the supervised training and testing on the same domain with multiple different LiDAR segmentation techniques.}
  \label{tab:supervised_backbones_lidar_appendix}
    \begin{tabular}{|c|ccccccccccccccccccc|c|}
    \hline
     & \rotatebox{90}{road}  & \rotatebox{90}{swalk} & \rotatebox{90}{building} & \rotatebox{90}{wall}  & \rotatebox{90}{fence} & \rotatebox{90}{pole}  & \rotatebox{90}{tlight} & \rotatebox{90}{tsign} & \rotatebox{90}{veg}   & \rotatebox{90}{terrain} & \rotatebox{90}{sky}   & \rotatebox{90}{person} & \rotatebox{90}{rider} & \rotatebox{90}{car}   & \rotatebox{90}{truck} & \rotatebox{90}{bus}   & \rotatebox{90}{train} & \rotatebox{90}{mbike} & \rotatebox{90}{bike}  & \rotatebox{90}{mIoU} \\\hline     
    RangeNet++ \cite{milioto2019icra-fiass} (SqueezeSeg-V2) & 95.3 & 78.5 & 82.0 & 86.7 & 66.8 & 54.4 & 23.3 & 41.1 & 81.6 & 66.5 & - & 61.6 & 41.5 & 90.5 & 49.5 & 67.7 & 62.8 & 39.2 & 25.3 & 61.9\\
    RangeNet++ \cite{milioto2019icra-fiass} (DarkNet-21) & 96.6 & 82.2 & 88.0 & 91.9 & 79.4 & 59.3 & 31.5 & 51.9 & 86.8 & 77.3 & - & 66.0 & 50.4 & 90.0 & 43.5 & 62.4 & 78.1 & 42.9 & 34.2 & 67.4\\
    Cylinder3D \cite{zhu2021cylindrical} & 96.7 & 65.8 & 85.7 & 79.0 & 87.5 & 75.6 & 87.9 & 91.2 & 88.5 & 48.9 & - & 95.0 & 89.5 & 98.9 & 84.5 & 55.3 & 42.2 & 88.7 & 84.6 & 80.3\\
    DLv2 \cite{chen2018deeplab} on spherical projections & 91.7 & 62.9 & 77.9 & 69.3 & 49.2 & 44.2 & 23.2 & 46.0 & 80.8 & 55.0 & - & 56.6 & 43.3 & 83.8 & 61.2 & 61.7 & 57.3 & 46.5 & 20.0 & 57.3 \\\hline
    \end{tabular}
\end{table*}

\begin{table}[t]
\setlength\tabcolsep{1.3pt} 
  \centering
  \caption{Detailed view of the label sets considered.}
    \begin{tabular}{|l|ccccccccccccccccccc|c|}
    \hline
          & \rotatebox{90}{road}  & \rotatebox{90}{swalk} & \rotatebox{90}{building} & \rotatebox{90}{wall}  & \rotatebox{90}{fence} & \rotatebox{90}{pole}  & \rotatebox{90}{tlight} & \rotatebox{90}{tsign} & \rotatebox{90}{veg}   & \rotatebox{90}{terrain} & \rotatebox{90}{sky}   & \rotatebox{90}{person} & \rotatebox{90}{rider} & \rotatebox{90}{car}   & \rotatebox{90}{truck} & \rotatebox{90}{bus}   & \rotatebox{90}{train} & \rotatebox{90}{mbike} & \rotatebox{90}{bike}  & \rotatebox{90}{Count} \\\hline
    City19 & \checkmark & \checkmark & \checkmark & \checkmark & \checkmark & \checkmark & \checkmark & \checkmark & \checkmark & \checkmark & \checkmark & \checkmark & \checkmark & \checkmark & \checkmark & \checkmark & \checkmark & \checkmark & \checkmark & 19 \\
    Idd17 & \checkmark & \checkmark & \checkmark & \checkmark & \checkmark & \checkmark & \checkmark & \checkmark & \checkmark & & \checkmark & \checkmark & \checkmark & \checkmark & \checkmark & \checkmark &  & \checkmark & \checkmark & 17 \\
    Synthia16 & \checkmark & \checkmark & \checkmark & \checkmark & \checkmark & \checkmark & \checkmark & \checkmark & \checkmark & & \checkmark & \checkmark & \checkmark & \checkmark &  & \checkmark &  & \checkmark & \checkmark & 16 \\
    Idda16 & \checkmark & \checkmark & \checkmark & \checkmark & \checkmark & \checkmark & \checkmark & \checkmark & \checkmark & \checkmark & \checkmark & \checkmark & \checkmark & \checkmark &  &  &  & \checkmark & \checkmark & 16 \\
    Idda-Synthia-15 & \checkmark & \checkmark & \checkmark & \checkmark & \checkmark & \checkmark & \checkmark & \checkmark & \checkmark &  & \checkmark & \checkmark & \checkmark & \checkmark &  &  &  & \checkmark & \checkmark & 15 \\
    Synthia-13 & \checkmark & \checkmark & \checkmark &  &  &  & \checkmark & \checkmark & \checkmark &  & \checkmark & \checkmark & \checkmark & \checkmark &  &  &  & \checkmark & \checkmark & 13 \\
    Idda-Synthia-12 & \checkmark & \checkmark & \checkmark &  &  &  & \checkmark & \checkmark & \checkmark &  & \checkmark & \checkmark & \checkmark &  &  &  &  & \checkmark & \checkmark & 12 \\ \hline
    \end{tabular}%
  \label{tab:label_sets}%
\end{table}%

\textbf{LiDAR.} Table~\ref{tab:supervised_backbones_lidar_appendix} presents the results obtained with two  LiDAR  semantic  segmentation  models, i.e., RangeNet and Cylinder3D~\cite{zhu2021cylindrical}, and with DeepLab-V2~\cite{chen2018deeplab} trained on LiDAR samples converted to spherical images. Note that the \emph{sky} class is not included in the LiDAR samples, as no signal is backscattered from the sky.
Cylinder3D~\cite{zhu2021cylindrical} scores the best result, with an impressive $80.3\%$ mIoU.
Its performance is consistent among classes, with the exception of \emph{terrain}, \emph{bus} and \emph{train}, lower than $60\%$.

\begin{table}[htbp]
  \centering
    \caption{Accuracy (mIoU) with images sampled either from the target Cityscapes dataset ($r$) or from our source dataset ($1-r$).}
    \begin{tabular}{|c|c|c|}
    \hline
    $r$     & \hspace{0.2cm}SELMA\hspace{0.2cm} & Cityscapes \\\hline
    $0.0$   & $68.9$  & $42.0$ \\
    $0.05$  & $68.9$  & $49.1$ \\
    $0.1$   & $\mathbf{69.7}$ & $53.6$ \\
    $0.2$   & $69.1$  & $55.9$ \\
    $0.3$   & $68.3$  & $58.5$ \\
    $0.4$   & $68.9$  & $59.5$ \\
    $0.5$   & $66.1$  & $61.8$ \\
    $0.6$   & $64.7$  & $63.8$ \\
    $0.7$   & $57.5$  & $65.5$ \\
    $0.8$   & $59.8$  & $66.4$ \\
    $0.9$   & $52.4$  & $68.1$ \\
    $0.95$  & $47.6$  & $\mathbf{68.4}$ \\
    $1.0$   & $37.3$  & $67.4$ \\\hline
    \end{tabular}%
  \label{tab:mixed}%
\end{table}%

\subsubsection{Domain Generalization}
\label{subsec:results_appendix:domainadaptation}
To assess the domain generalization properties from  of the proposed dataset, we run an extensive comparison with different datasets and using DeepLab-V2 as segmentation architecture to tackle real datasets either with labeled sample from the real-world domain or without.

In Table~\ref{tab:mixed} we show the mIoU accuracy on source and target sets when training the segmentation network on samples drawn either from the target domain (with probability $r$) or from the source domain (with probability $1-r$).
Adding as few as $5\%$ to $10\%$ of data from a different domain greatly helps in improving domain generalization, i.e., the network can perform well on both domains.
Even more, we highlight that a $5\%$ of source samples can significantly improve the performance on the target domain from $67.4\%$ to $68.4\%$. Similarly, $10\%$ of target samples improve the performance on the source domain from $68.9\%$ to $69.7\%$. 

In~\Cref{fig:matrix}, we reported an extensive evaluation to investigate the domain generalization properties of models trained on a source domain with respect to a target domain. 
We show in Table~\ref{tab:sourceonly_deeplabv2} the per-class IoU results for the source only training on each domain and testing on the Cityscapes dataset.

Finally, we report in Table~\ref{tab:cameras} the per-class IoU results for the supervised training on desk camera and testing on the different points of views.

\begin{table*}[htbp]
\setlength\tabcolsep{1pt} 
  \centering
  \caption{Per-class IoUs for the source only training on each domain and testing on Cityscapes.}
  \centering
    \begin{tabular}{|l|ccccccccccccccccccc|cccccc|}
          \hline
          & \multicolumn{19}{c|}{per-class IoUs} & \multicolumn{6}{c|}{mIoUs on different label sets} \\\hline
          & \rotatebox{90}{road}  & \rotatebox{90}{swalk} & \rotatebox{90}{building} & \rotatebox{90}{wall}  & \rotatebox{90}{fence} & \rotatebox{90}{pole}  & \rotatebox{90}{tlight} & \rotatebox{90}{tsign} & \rotatebox{90}{veg}   & \rotatebox{90}{terrain} & \rotatebox{90}{sky}   & \rotatebox{90}{person} & \rotatebox{90}{rider} & \rotatebox{90}{car}   & \rotatebox{90}{truck} & \rotatebox{90}{bus}   & \rotatebox{90}{train} & \rotatebox{90}{mbike} & \rotatebox{90}{bike}  & \rotatebox{90}{city19} & \rotatebox{90}{idda16} & \rotatebox{90}{synthia16} & \rotatebox{90}{synthia13} & \rotatebox{90}{idda-synthia-15} & \rotatebox{90}{idda-synthia-12} \\\hline
    Cityscapes~\cite{Cordts2016}& 97.1  & 77.4  & 89.2  & 50.1  & 46.2  & 45.2  & 48.9  & 61.4  & 89.5  & 55.2  & 92.2  & 69.9  & 46.5  & 91.4  & 66.9  & 75.2  & 60.3  & 52.2  & 65.8  & 67.4  & 67.4  & 68.6  & 73.6  & 68.2  & 73.5 \\
    \hline
    SELMA & 91.1  & 50.2  & 80.4  & 24.8  & 11.5  & 32.0  & 13.1  & 31.3  & 82.3  & 24.9  & 76.5  & 51.9  & 28.0  & 76.5  & 17.6  & 21.9  & 17.7  & 29.3  & 36.9  & 42.0  & 46.3  & 46.1  & 51.5  & 47.7  & 54.0 \\
    Noon & 88.5  & 52.1  & 81.9  & 21.8  & 14.3  & 29.2  & 16.4  & 31.2  & 83.0  & 26.0  & 78.7  & 54.7  & 27.6  & 69.2  & 15.5  & 9.1   & 21.9  & 32.0  & 35.3  & 41.5  & 46.4  & 45.3  & 50.7  & 47.7  & 54.2 \\
    Night & 88.1  & 38.7  & 66.5  & 8.5   & 9.0   & 30.7  & 5.5   & 26.7  & 70.0  & 14.5  & 8.1   & 41.4  & 24.1  & 65.0  & 7.0   & 13.8  & 10.2  & 24.3  & 33.5  & 30.8  & 34.7  & 34.6  & 38.9  & 36.0  & 41.0 \\
    MostlyClear & 87.8  & 52.7  & 80.8  & 20.2  & 11.6  & 32.3  & 14.7  & 30.6  & 82.7  & 27.4  & 77.7  & 53.1  & 29.9  & 69.5  & 12.8  & 20.7  & 17.4  & 28.6  & 38.8  & 41.5  & 46.2  & 45.7  & 51.4  & 47.4  & 53.9 \\
    Rain & 89.0  & 44.6  & 79.4  & 21.2  & 9.5   & 30.7  & 13.0  & 31.3  & 81.4  & 21.5  & 75.6  & 49.8  & 26.6  & 72.1  & 9.9   & 19.5  & 13.0  & 27.3  & 36.1  & 39.6  & 44.3  & 44.2  & 49.7  & 45.8  & 52.2 \\
    Fog & 77.9  & 47.9  & 75.2  & 11.6  & 9.9   & 27.9  & 12.1  & 28.9  & 79.1  & 18.2  & 67.1  & 47.8  & 23.5  & 60.1  & 3.9   & 11.3  & 11.8  & 25.5  & 35.2  & 35.5  & 40.5  & 40.1  & 45.5  & 42.0  & 48.4 \\\hdashline
    GTA5~\cite{Richter2016}& 70.3  & 21.9  & 82.5  & 29.3  & 25.4  & 31.7  & 32.5  & 16.6  & 84.4  & 41.1  & 82.6  & 52.1  & 7.9   & 78.3  & 33.2  & 34.0  & 7.9   & 25.9  & 9.9   & 40.4  & 43.3  & 42.8  & 46.1  & 43.4  & 47.1 \\
    IDDA~\cite{alberti2020idda}  & 88.4  & 49.8  & 78.0  & 15.8  & 8.7   & 35.2  & 19.8  & 9.1   & 81.3  & 21.8  & 69.5  & 48.8  & 23.4  & 71.3  & -     & -     & -     & 15.3  & 21.8  & 34.6  & 41.1  & 39.8  & 44.3  & 42.4  & 48.0 \\
    SYNTHIA~\cite{ros2016}& 43.3  & 18.6  & 78.0  & 14.0  & 0.1   & 32.7  & 5.4   & 15.9  & 81.2  & -     & 80.0  & 55.7  & 19.2  & 63.5  & -     & 0.0   & -     & 8.0   & 23.3  & 28.4  & 33.7  & 33.7  & 37.9  & 35.9  & 41.0 \\\hline
    ACDC~\cite{sakaridis2021acdc}  & 82.8  & 46.1  & 81.0  & 26.9  & 22.1  & 33.6  & 27.6  & 41.9  & 85.1  & 37.9  & 80.8  & 53.8  & 25.5  & 76.9  & 30.2  & 35.8  & 13.7  & 16.8  & 49.5  & 45.7  & 49.3  & 49.1  & 54.1  & 50.0  & 55.7 \\
    Cityscapes Fog~\cite{sakaridis2018semantic}& 96.9  & 75.5  & 87.0  & 43.5  & 37.5  & 41.4  & 39.1  & 56.4  & 87.2  & 52.0  & 88.9  & 65.6  & 42.6  & 89.5  & 57.9  & 71.6  & 48.4  & 44.0  & 60.7  & 62.4  & 63.0  & 64.2  & 69.6  & 63.7  & 69.5 \\
    Cityscapes Rain~\cite{hu2019depth}& 93.4  & 58.9  & 87.3  & 9.1   & 9.6   & 34.9  & 42.7  & 57.8  & 89.2  & 61.1  & 91.8  & 63.0  & 48.1  & 87.1  & 20.4  & 75.3  & 0.0   & 5.1   & 67.3  & 52.7  & 56.7  & 57.5  & 66.7  & 56.4  & 66.0 \\
    IDD~\cite{varma2019idd}& 88.9  & 21.9  & 84.6  & 30.5  & 24.9  & 37.5  & 6.1   & 30.7  & 86.8  & -     & 90.6  & 61.6  & 34.8  & 83.6  & 43.0  & 44.1  & -     & 15.3  & 36.1  & 43.2  & 45.9  & 48.6  & 52.7  & 48.9  & 53.4 \\
    Mapillary~\cite{neuhold2017mapillary}& 93.9  & 60.5  & 87.2  & 48.4  & 39.5  & 42.3  & 41.0  & 53.3  & 88.4  & 46.4  & 90.7  & 64.0  & 36.2  & 89.9  & 54.8  & 64.4  & 13.5  & 43.7  & 59.0  & 58.8  & 61.5  & 62.7  & 67.1  & 62.5  & 67.3 \\
    NightCity~\cite{tan2021night}& 92.3  & 48.9  & 73.4  & 22.7  & 14.2  & 22.4  & 6.7   & 16.8  & 77.6  & 23.7  & 52.1  & 47.9  & 14.6  & 81.3  & 23.8  & 40.3  & 5.7   & 7.2   & 30.0  & 36.9  & 39.5  & 40.5  & 45.3  & 40.5  & 45.7 \\\hline
    \end{tabular}%
  \label{tab:sourceonly_deeplabv2}%
\end{table*}%

\begin{table*}[htbp]
  \setlength\tabcolsep{3pt}  
  \centering
  \caption{Per-class IoUs for the supervised training on the desk camera and testing on the different point of views.}
  \label{tab:desk_to_povs_appendix}
    \begin{tabular}{|c|ccccccccccccccccccc|c|}
    \hline
     & \rotatebox{90}{road}  & \rotatebox{90}{swalk} & \rotatebox{90}{building} & \rotatebox{90}{wall}  & \rotatebox{90}{fence} & \rotatebox{90}{pole}  & \rotatebox{90}{tlight} & \rotatebox{90}{tsign} & \rotatebox{90}{veg}   & \rotatebox{90}{terrain} & \rotatebox{90}{sky}   & \rotatebox{90}{person} & \rotatebox{90}{rider} & \rotatebox{90}{car}   & \rotatebox{90}{truck} & \rotatebox{90}{bus}   & \rotatebox{90}{train} & \rotatebox{90}{mbike} & \rotatebox{90}{bike}  & \rotatebox{90}{mIoU} \\\hline     
    Desk & 99.1 & 87.5 & 83.9 & 83.5 & 57.3 & 47.1 & 36.8 & 61.1 & 78.4 & 77.2 & 89.4 & 63.9 & 62.8 & 88.7 & 71.6 & 48.9 & 72.3 & 62.0 & 37.7 & 68.9 \\ \hdashline
    Front Left & 96.7 & 76.7 & 84.3 & 78.8 & 53.8 & 45.4 & 43.7 & 58.0 & 76.9 & 73.8 & 92.8 & 62.8 & 56.4 & 89.6 & 65.1 & 48.5 & 69.2 & 58.7 & 33.6 & 66.6\\
    Front & 96.4 & 75.1 & 84.2 & 79.0 & 54.1 & 45.3 & 43.9 & 58.0 & 77.0 & 73.8 & 92.8 & 63.5 & 56.9 & 89.7 & 65.8 & 48.8 & 69.3 & 58.5 & 33.5 & 66.6\\
    Front Right & 96.1 & 73.5 & 84.1 & 79.0 & 54.5 & 45.2 & 44.6 & 57.6 & 77.0 & 73.7 & 92.8 & 64.0 & 56.9 & 89.6 & 65.8 & 48.9 & 70.1 & 58.5 & 33.4 & 66.6\\ \hdashline
    Left & 94.8 & 76.9 & 87.0 & 82.7 & 46.4 & 54.7 & 41.3 & 48.6 & 81.4 & 78.9 & 91.5 & 66.6 & 56.0 & 86.1 & 51.0 & 46.5 & 71.1 & 64.1 & 32.8 & 66.2\\
    Back & 97.3 & 80.9 & 84.6 & 82.8 & 53.9 & 46.7 & 34.1 & 46.6 & 78.6 & 76.7 & 93.7 & 64.4 & 55.6 & 88.1 & 56.7 & 49.6 & 69.3 & 60.3 & 31.7 & 65.9\\
    Right & 81.1 & 53.0 & 84.6 & 81.2 & 61.7 & 55.4 & 38.2 & 45.6 & 83.7 & 78.2 & 89.6 & 77.2 & 52.7 & 83.0 & 28.0 & 26.2 & 68.5 & 62.5 & 31.7 & 62.2\\ \hline
    \end{tabular}
    \label{tab:cameras}%
\end{table*}

\bibliographystyle{IEEEtran}
\bibliography{strings_short,refs}
%
%
%
%
%
%
\begin{IEEEbiography}[{\includegraphics[width=1in,height=1.25in,clip,keepaspectratio]{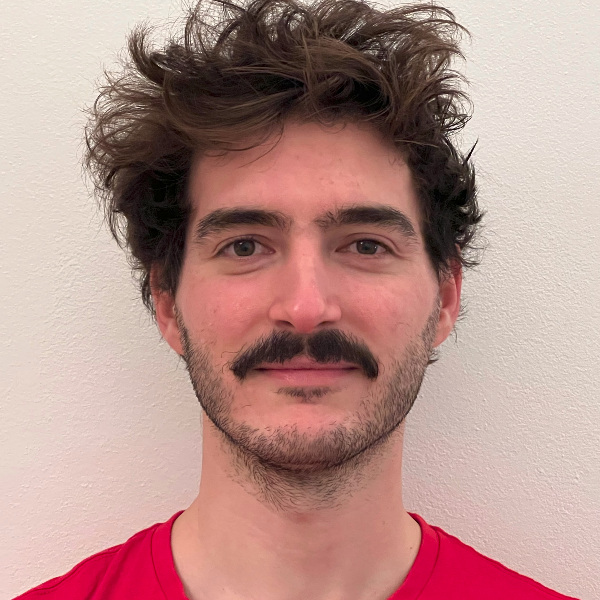}}]{Paolo Testolina}
received the B.Sc. in Information Engineering in 2016 and the M.Sc. with Honors in Telecommunication Engineering in 2018 from the University of Padova. He is currently working towards his Ph.D. in Information Engineering at the same university.
His research focuses mainly on mmWave networks, from channel modeling to link layer simulation, traffic modeling, and vehicular networks.
\end{IEEEbiography}
\vspace{-4.5em}
\begin{IEEEbiography}[{\includegraphics[width=1in,height=1.25in,clip,keepaspectratio]{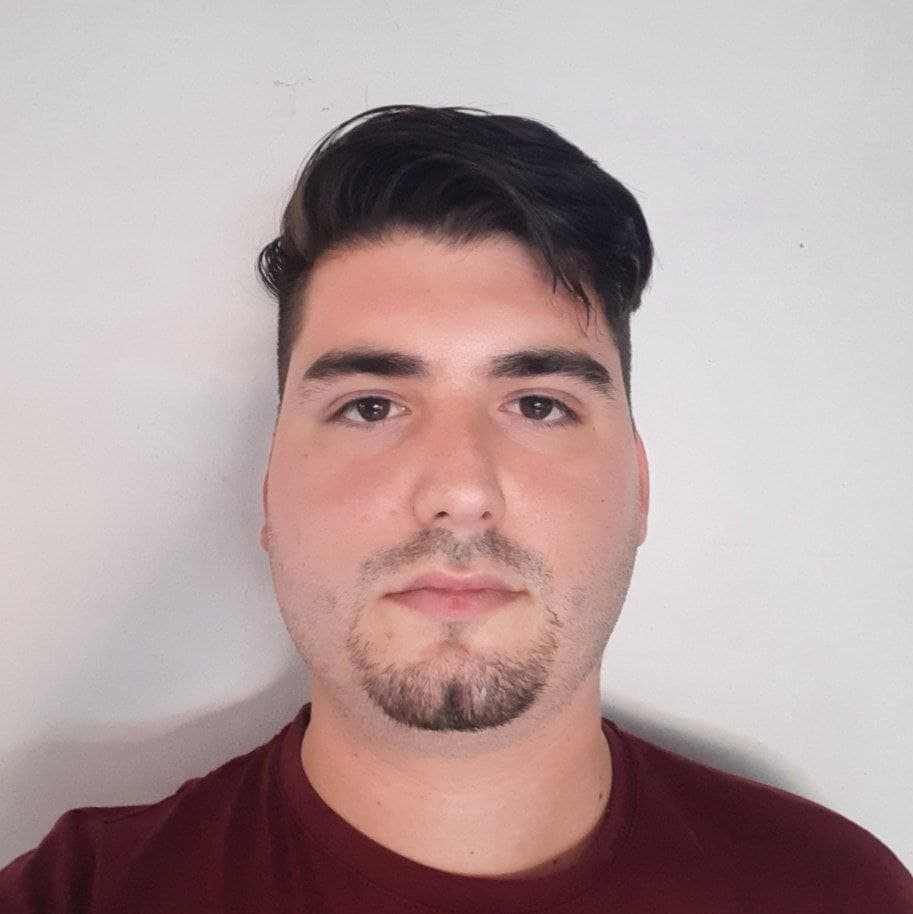}}]{Francesco Barbato}
received the M.Sc.\ degree in Telecommunication Engineering with Honors from the University of Padova in 2020. He is currently a Ph.D.\ student in the same University. His research focuses on unsupervised domain adaptation, multi-modal learning and continual learning applied to computer vision tasks, particularly to \gls{ss} for autonomous vehicles.
\end{IEEEbiography}
\vspace{-4.5em}
\begin{IEEEbiography}[{\includegraphics[width=1in,height=1.25in,clip,keepaspectratio]{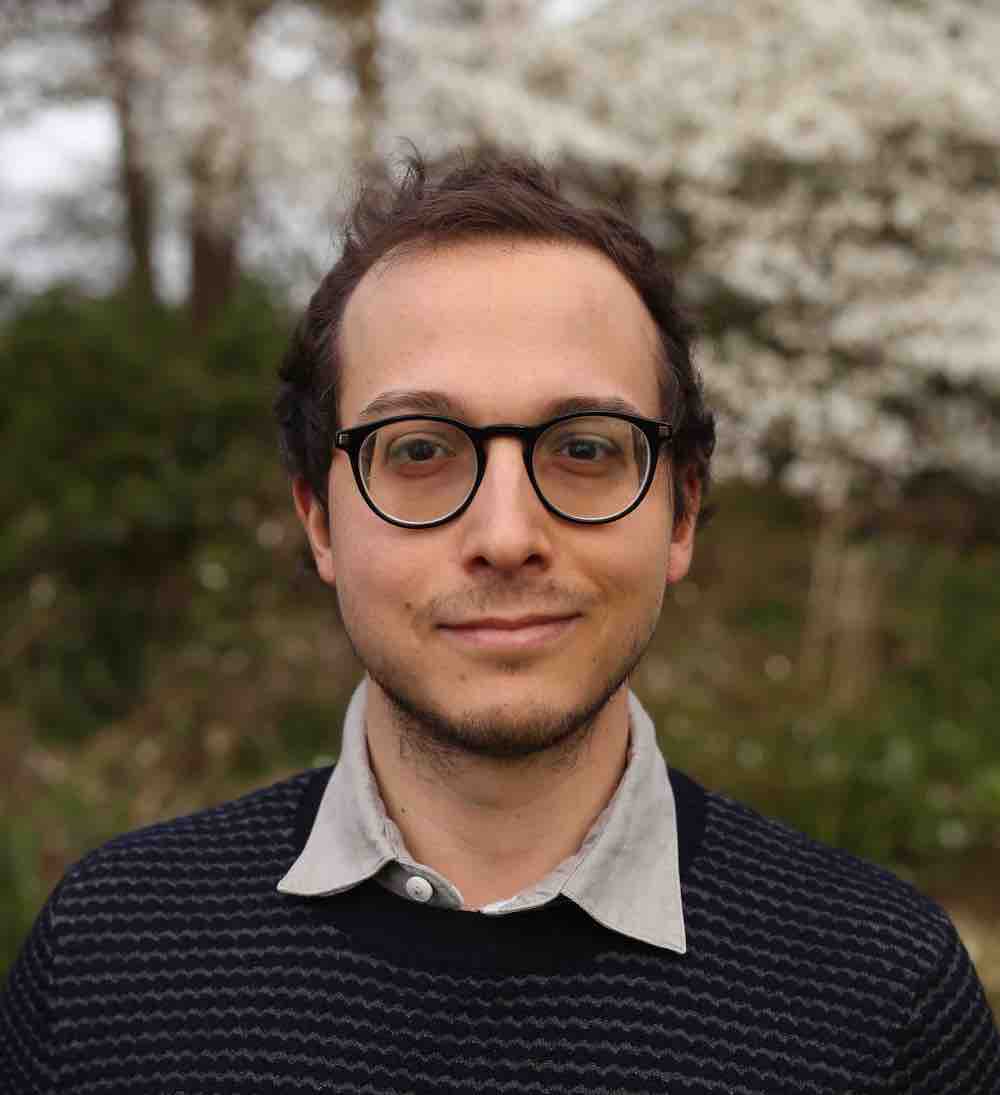}}]{Umberto Michieli}
received his Ph.D.\ in Information Engineering from the University of Padova in 2021. Currently, he is a Postoctoral Researcher and Adjunct Professor at the same University. He spent research periods at Technische Universit\"at Dresden and Samsung Research UK. His research lies at the intersection of foundation AI problems applied to semantic understanding. In particular, he focuses on domain adaptation, continual learning, coarse-to-fine learning and federated learning.
\end{IEEEbiography}
\vspace{-4.5em}
\begin{IEEEbiography}[{\includegraphics[width=1in,height=1.25in,clip,keepaspectratio]{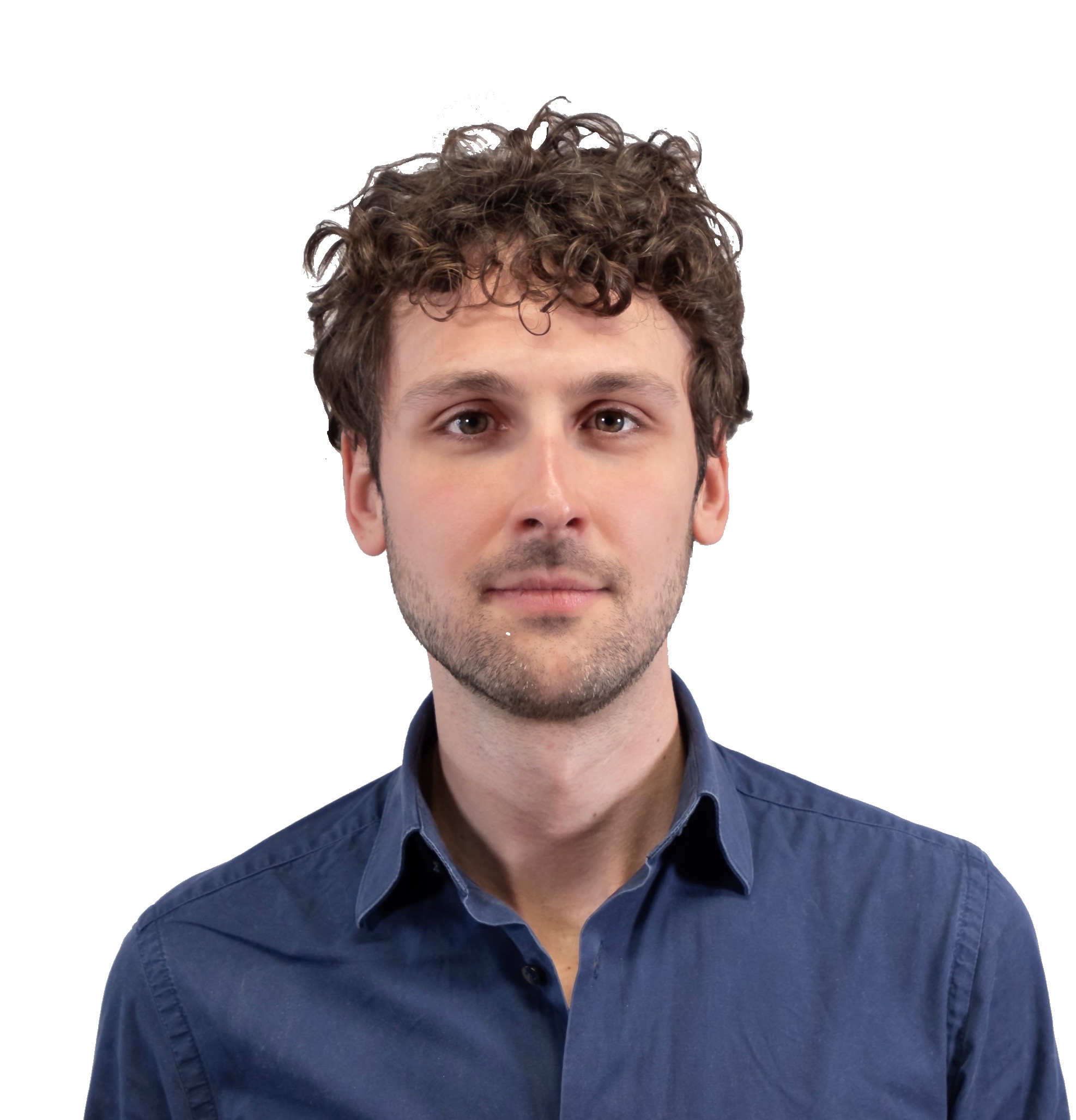}}]{Marco Giordani}
received his Ph.D. in Information Engineering in 2020 from the University of Padova, Italy, where he is now a Postdoc and Adjunct Professor. He visited NYU and TOYOTA, and was awarded with the IEEE VTS “Daniel E. Noble Fellowship Award” in 2018 and the IEEE ComSoc EMEA Outstanding Young Researcher Award in 2021. His research focuses on protocol design for 5G/6G networks.
%
\end{IEEEbiography}
\begin{IEEEbiography}[{\includegraphics[width=1in,height=1.25in,clip,keepaspectratio]{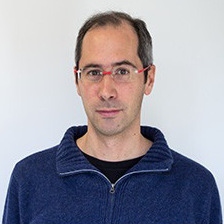}}]{Pietro Zanuttigh}
received his Ph.D. at the University of Padova in 2007. Currently he is an associate professor at the Department of Information Engineering. He works in the computer vision and machine learning fields, with a special  focus on domain adaptation and incremental learning in semantic segmentation, 3D acquisition with ToF sensors, depth data processing, sensor fusion and hand gesture recognition.
\end{IEEEbiography}
\begin{IEEEbiography}[{\includegraphics[width=1in,height=1.25in,clip,keepaspectratio]{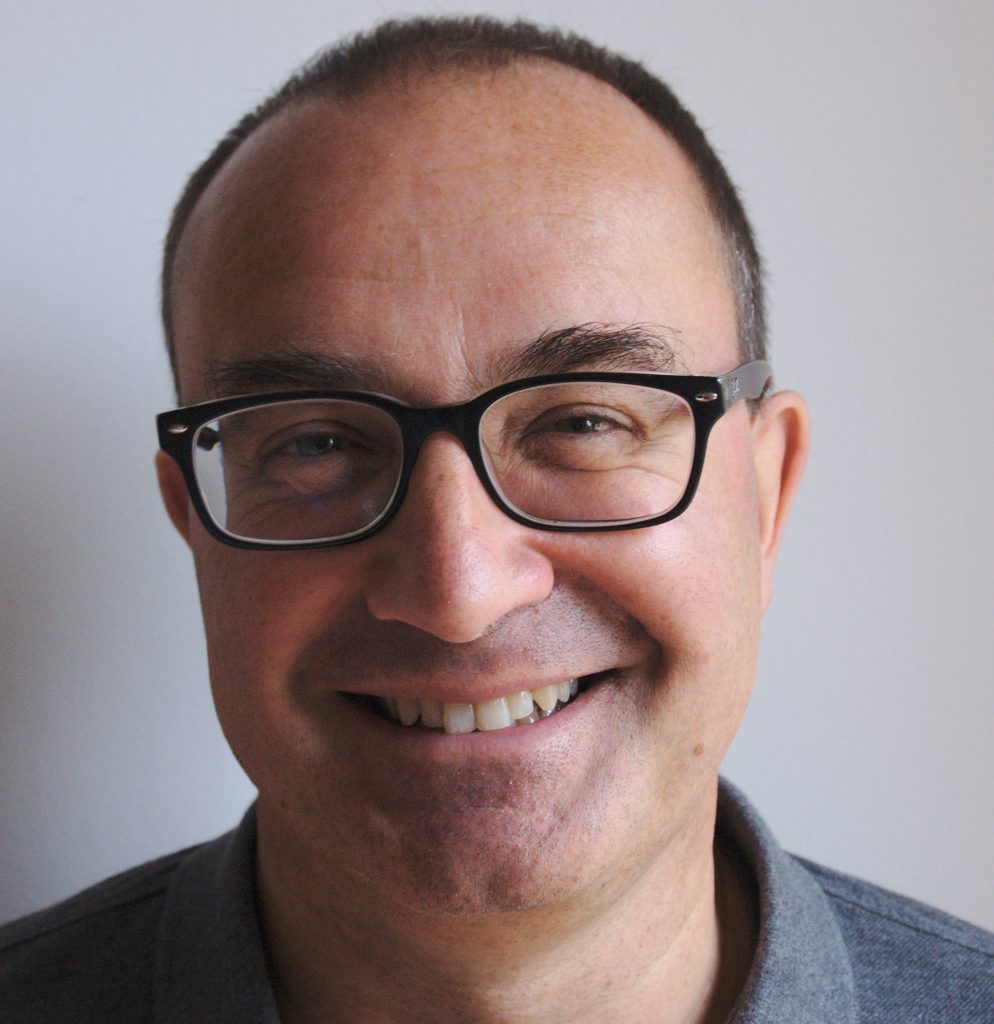}}]{Michele Zorzi}
 is with the Information Engineering Department of the University of Padova, focusing on wireless communications. He was Editor-in-Chief of IEEE Wireless Communications from 2003 to 2005, IEEE Trans. on Communications from 2008 to 2011, and IEEE Trans. on Cognitive Communications and Networking from 2014 to 2018. He served ComSoc as a Member-at-Large of the Board of Governors from 2009 to 2011 and from 2021 to 2023, as Director of Education and Training from 2014 to 2015, and as Director of Journals from 2020 to 2021.
\end{IEEEbiography}



\vfill

\end{document}

%% file: acronymsSorted.tex
\newacronym{yolo}{YOLO}{You Only Look Once}
\newacronym{ml}{ML}{machine learning}
\newacronym{dl}{DL}{deep learning}
\newacronym{ai}{AI}{artificial intelligence}
\newacronym{fov}{FoV}{field of view}
\newacronym{api}{API}{Application Programming Interface}
\newacronym{crs}{CRS}{Cell Reference Signal}
\newacronym{v2v}{V2V}{Vehicle-to-Vehicle}
\newacronym{v2i}{V2I}{Vehicle-to-Infrastructure}
\newacronym{v2n}{V2N}{Vehicle-to-Network}
\newacronym{v2x}{V2X}{Vehicle-to-Everything}
\newacronym{vn}{VN}{Vehicular Node}
\newacronym{dsrc}{DSRC}{Dedicated Short Range Communication}
\newacronym{ci}{CI}{context information}
\newacronym{voi}{VoI}{value of information}
\newacronym{gps}{GPS}{Global Positioning System}
\newacronym{qos}{QoS}{Quality of Service}
\newacronym{qoe}{QoE}{Quality of Experience}
\newacronym{ahp}{AHP}{Analytic Hierarchy Process}
\newacronym{lidar}{LiDAR}{Light Detection and Ranging}
\newacronym{sumo}{SUMO}{Simulation of Urban MObility}
\newacronym{wave}{WAVE}{Wireless Access in Vehicular Environment}
\newacronym{c-its}{C-ITS}{Connected Intelligent Transportation System}
\newacronym{dash}{DASH}{Dynamic Adaptive Streaming over HTTP}
\newacronym{http}{HTTP}{HyperText Transfer Protocol}
\newacronym{3d}{3D}{thee-dimensional}
\newacronym{dnn}{DNN}{Deep Neural Network}
\newacronym{gpcc}{G-PCC}{geometry-based point cloud compression}
\newacronym{hsc}{HSC}{Hybrid Semantic Compression}
\newacronym{bpp}{BPP}{bits per point}
\newacronym{cpu}{CPU}{Central Processing Unit}
\newacronym{iou}{IoU}{Intersection over Union}
\newacronym{npc}{NPC}{Non-Player Character}
\newacronym{ue4}{UE4}{Unreal Engine 4}
\newacronym{ss}{SS}{semantic segmentation}
\newacronym{uda}{UDA}{Unsupervised Domain Adaptation}
\newacronym{carla}{CARLA}{CAR Learning to Act}
\newacronym{fps}{FPS}{frame per second}
\newacronym{selma}{SELMA}{SEmantic Large-scale Multimodal Acquisitions in Variable Weather, Daytime and Viewpoints}
\newacronym{cv}{CV}{computer vision}
\newacronym{miou}{mIoU}{mean Intersection over Union}
\newacronym{pov}{PoV}{Point-of-View}

%% file: imgs/design/split_pie.tex
\begin{tikzpicture}

\definecolor{color0}{rgb}{0.192156862745098,0.509803921568627,0.741176470588235}
\definecolor{color1}{rgb}{0.901960784313726,0.333333333333333,0.0509803921568627}
\definecolor{color2}{rgb}{0.192156862745098,0.63921568627451,0.329411764705882}
\definecolor{color3}{rgb}{0.458823529411765,0.419607843137255,0.694117647058824}
\definecolor{color4}{rgb}{0.419607843137255,0.682352941176471,0.83921568627451}
\definecolor{color5}{rgb}{0.992156862745098,0.552941176470588,0.235294117647059}
\definecolor{color6}{rgb}{0.454901960784314,0.768627450980392,0.462745098039216}
\definecolor{color7}{rgb}{0.619607843137255,0.603921568627451,0.784313725490196}
\definecolor{color8}{rgb}{0.619607843137255,0.792156862745098,0.882352941176471}
\definecolor{color9}{rgb}{0.992156862745098,0.682352941176471,0.419607843137255}
\definecolor{color10}{rgb}{0.631372549019608,0.850980392156863,0.607843137254902}
\definecolor{color11}{rgb}{0.737254901960784,0.741176470588235,0.862745098039216}

\pgfplotsset{
tick label style={font=\scriptsize},
label style={font=\scriptsize},
legend  style={font=\scriptsize}
}

\begin{axis}[
axis lines=none,
width=\fwidth,
height=\fheight,
at={(0\fwidth,0\fheight)},
scale only axis,
legend cell align={left},
legend style={legend cell align=left, align=left, draw=white!15!black, at={(0.5,1.35)},/tikz/every even column/.append style={column sep=0.15cm},
  anchor=north ,legend columns=3},
xmin=-1., xmax=1.,
ymin=-1., ymax=1.
]
\addlegendimage{circle,area legend,fill=white, draw=black};
\addlegendentry{Noon}

\addlegendimage{circle,area legend,fill=white, draw=black, postaction={pattern=vertical lines, opacity=0.5}};
\addlegendentry{Sunset}

\addlegendimage{circle,area legend,fill=white, draw=black, postaction={pattern=north east lines, opacity=0.5}};
\addlegendentry{Night}

\addlegendimage{circle,area legend,fill=color0, draw=black};
\addlegendentry{Clear}

\addlegendimage{circle,area legend,fill=color1, draw=black};
\addlegendentry{Hard Rain}

\addlegendimage{circle,area legend,fill=color2, draw=black};
\addlegendentry{Mid Rain}

\addlegendimage{circle,area legend,fill=color3, draw=black};
\addlegendentry{Soft Rain}

\addlegendimage{circle,area legend,fill=white!38.8235294117647!black, draw=black};
\addlegendentry{Soft Fog}

\addlegendimage{circle,area legend,fill=white!85.0980392156863!black, draw=black};
\addlegendentry{Hard Fog}

\path [draw=black, fill=color0]
(axis cs:1,0)
.. controls (axis cs:1,0.197757642866966) and (axis cs:0.941337872031553,0.391140762375141) .. (axis cs:0.831469612302545,0.555570233019602)
.. controls (axis cs:0.721601352573538,0.719999703664063) and (axis cs:0.565387671007556,0.848200958962527) .. (axis cs:0.38268343236509,0.923879532511287)
.. controls (axis cs:0.199979193722624,0.999558106060047) and (axis cs:-0.0011325368049695,1.0193658826313) .. (axis cs:-0.195090322016128,0.98078528040323)
.. controls (axis cs:-0.389048107227287,0.942204678175164) and (axis cs:-0.567271010883849,0.846942551489246) .. (axis cs:-0.707106781186547,0.707106781186548)
.. controls (axis cs:-0.846942551489246,0.567271010883849) and (axis cs:-0.942204678175164,0.389048107227287) .. (axis cs:-0.98078528040323,0.195090322016129)
.. controls (axis cs:-1.0193658826313,0.00113253680496991) and (axis cs:-0.999558106060047,-0.199979193722624) .. (axis cs:-0.923879532511287,-0.38268343236509)
.. controls (axis cs:-0.848200958962527,-0.565387671007556) and (axis cs:-0.719999703664063,-0.721601352573538) .. (axis cs:-0.555570233019602,-0.831469612302545)
.. controls (axis cs:-0.391140762375141,-0.941337872031553) and (axis cs:-0.197757642866966,-1) .. (axis cs:-1.83697019872103e-16,-1)
--(axis cs:-9.18485099360515e-17,-0.5)
.. controls (axis cs:-0.0988788214334829,-0.5) and (axis cs:-0.195570381187571,-0.470668936015776) .. (axis cs:-0.277785116509801,-0.415734806151273)
.. controls (axis cs:-0.359999851832032,-0.360800676286769) and (axis cs:-0.424100479481263,-0.282693835503778) .. (axis cs:-0.461939766255643,-0.191341716182545)
.. controls (axis cs:-0.499779053030023,-0.0999895968613118) and (axis cs:-0.509682941315649,0.000566268402484957) .. (axis cs:-0.490392640201615,0.0975451610080643)
.. controls (axis cs:-0.471102339087582,0.194524053613644) and (axis cs:-0.423471275744623,0.283635505441924) .. (axis cs:-0.353553390593274,0.353553390593274)
.. controls (axis cs:-0.283635505441924,0.423471275744623) and (axis cs:-0.194524053613643,0.471102339087582) .. (axis cs:-0.0975451610080641,0.490392640201615)
.. controls (axis cs:-0.000566268402484749,0.509682941315649) and (axis cs:0.0999895968613119,0.499779053030023) .. (axis cs:0.191341716182545,0.461939766255643)
.. controls (axis cs:0.282693835503778,0.424100479481263) and (axis cs:0.360800676286769,0.359999851832032) .. (axis cs:0.415734806151273,0.277785116509801)
.. controls (axis cs:0.470668936015776,0.195570381187571) and (axis cs:0.5,0.0988788214334828) .. (axis cs:0.5,0)
--(axis cs:1,0)
--cycle;
\path [draw=black, fill=color1]
(axis cs:-1.83697019872103e-16,-1)
.. controls (axis cs:0.0628783247670458,-1) and (axis cs:0.12561673363166,-0.994069474411735) .. (axis cs:0.187381310447148,-0.982287251518164)
.. controls (axis cs:0.249145887262636,-0.970505028624594) and (axis cs:0.309661756814511,-0.952923543676019) .. (axis cs:0.368124544850001,-0.929776488990219)
--(axis cs:0.184062272425001,-0.46488824449511)
.. controls (axis cs:0.154830878407255,-0.476461771838009) and (axis cs:0.124572943631318,-0.485252514312297) .. (axis cs:0.0936906552235738,-0.491143625759082)
.. controls (axis cs:0.0628083668158298,-0.497034737205867) and (axis cs:0.0314391623835229,-0.5) .. (axis cs:-9.18485099360515e-17,-0.5)
--(axis cs:-1.83697019872103e-16,-1)
--cycle;
\path [draw=black, fill=color2]
(axis cs:0.368124544850001,-0.929776488990219)
.. controls (axis cs:0.426587332885492,-0.90662943430442) and (axis cs:0.482736858371565,-0.878019822837196) .. (axis cs:0.535826784307019,-0.844327932274658)
.. controls (axis cs:0.588916710242474,-0.810636041712119) and (axis cs:0.638710766818124,-0.772011813430412) .. (axis cs:0.684547093643514,-0.728968638957958)
--(axis cs:0.342273546821757,-0.364484319478979)
.. controls (axis cs:0.319355383409062,-0.386005906715206) and (axis cs:0.294458355121237,-0.40531802085606) .. (axis cs:0.26791339215351,-0.422163966137329)
.. controls (axis cs:0.241368429185782,-0.439009911418598) and (axis cs:0.213293666442746,-0.45331471715221) .. (axis cs:0.184062272425001,-0.46488824449511)
--(axis cs:0.368124544850001,-0.929776488990219)
--cycle;
\path [draw=black, fill=color3]
(axis cs:0.684547093643514,-0.728968638957958)
.. controls (axis cs:0.730383420468904,-0.685925464485504) and (axis cs:0.772058028934104,-0.638654901871031) .. (axis cs:0.809016981992651,-0.587785269335241)
.. controls (axis cs:0.845975935051198,-0.536915636799452) and (axis cs:0.878054751690301,-0.482673323024583) .. (axis cs:0.904827041702649,-0.425779314438399)
--(axis cs:0.452413520851324,-0.212889657219199)
.. controls (axis cs:0.439027375845151,-0.241336661512292) and (axis cs:0.422987967525599,-0.268457818399726) .. (axis cs:0.404508490996326,-0.293892634667621)
.. controls (axis cs:0.386029014467052,-0.319327450935515) and (axis cs:0.365191710234452,-0.342962732242752) .. (axis cs:0.342273546821757,-0.364484319478979)
--(axis cs:0.684547093643514,-0.728968638957958)
--cycle;
\path [draw=black, fill=white!38.8235294117647!black]
(axis cs:0.904827041702649,-0.425779314438399)
.. controls (axis cs:0.920436598421575,-0.39260732005228) and (axis cs:0.934210166847938,-0.358602012864486) .. (axis cs:0.9460853507908,-0.323917441671559)
.. controls (axis cs:0.957960534733663,-0.289232870478632) and (axis cs:0.967919374992536,-0.25392148980851) .. (axis cs:0.975916756628496,-0.218143265153242)
--(axis cs:0.487958378314248,-0.109071632576621)
.. controls (axis cs:0.483959687496268,-0.126960744904255) and (axis cs:0.478980267366831,-0.144616435239316) .. (axis cs:0.4730426753954,-0.16195872083578)
.. controls (axis cs:0.467105083423969,-0.179301006432243) and (axis cs:0.460218299210788,-0.19630366002614) .. (axis cs:0.452413520851324,-0.212889657219199)
--(axis cs:0.904827041702649,-0.425779314438399)
--cycle;
\path [draw=black, fill=white!85.0980392156863!black]
(axis cs:0.975916756628496,-0.218143265153242)
.. controls (axis cs:0.983914138264456,-0.182365040497974) and (axis cs:0.989937966594939,-0.146174080350765) .. (axis cs:0.993960952835292,-0.109734334821687)
.. controls (axis cs:0.997983939075646,-0.0732945892926087) and (axis cs:0.999999999141884,-0.0366611673129293) .. (axis cs:1,-2.3406689276376e-08)
--(axis cs:0.5,-1.1703344638188e-08)
.. controls (axis cs:0.499999999570942,-0.0183305836564646) and (axis cs:0.498991969537823,-0.0366472946463043) .. (axis cs:0.496980476417646,-0.0548671674108434)
.. controls (axis cs:0.494968983297469,-0.0730870401753824) and (axis cs:0.491957069132228,-0.0911825202489869) .. (axis cs:0.487958378314248,-0.109071632576621)
--(axis cs:0.975916756628496,-0.218143265153242)
--cycle;
\path [draw=black, fill=color0, postaction={pattern=vertical lines, opacity=0.5}]
(axis cs:0.5,0)
.. controls (axis cs:0.5,0.0988788214334828) and (axis cs:0.470668936015776,0.195570381187571) .. (axis cs:0.415734806151273,0.277785116509801)
.. controls (axis cs:0.360800676286769,0.359999851832032) and (axis cs:0.282693835503778,0.424100479481263) .. (axis cs:0.191341716182545,0.461939766255643)
.. controls (axis cs:0.0999895968613119,0.499779053030023) and (axis cs:-0.000566268402484749,0.509682941315649) .. (axis cs:-0.0975451610080641,0.490392640201615)
.. controls (axis cs:-0.194524053613643,0.471102339087582) and (axis cs:-0.283635505441924,0.423471275744623) .. (axis cs:-0.353553390593274,0.353553390593274)
.. controls (axis cs:-0.423471275744623,0.283635505441924) and (axis cs:-0.471102339087582,0.194524053613644) .. (axis cs:-0.490392640201615,0.0975451610080643)
.. controls (axis cs:-0.509682941315649,0.000566268402484957) and (axis cs:-0.499779053030023,-0.0999895968613118) .. (axis cs:-0.461939766255643,-0.191341716182545)
.. controls (axis cs:-0.424100479481263,-0.282693835503778) and (axis cs:-0.359999851832032,-0.360800676286769) .. (axis cs:-0.277785116509801,-0.415734806151273)
.. controls (axis cs:-0.195570381187571,-0.470668936015776) and (axis cs:-0.0988788214334829,-0.5) .. (axis cs:-9.18485099360515e-17,-0.5)
--(axis cs:-4.59242549680257e-17,-0.25)
.. controls (axis cs:-0.0494394107167414,-0.25) and (axis cs:-0.0977851905937853,-0.235334468007888) .. (axis cs:-0.138892558254901,-0.207867403075636)
.. controls (axis cs:-0.179999925916016,-0.180400338143384) and (axis cs:-0.212050239740632,-0.141346917751889) .. (axis cs:-0.230969883127822,-0.0956708580912724)
.. controls (axis cs:-0.249889526515012,-0.0499947984306559) and (axis cs:-0.254841470657824,0.000283134201242478) .. (axis cs:-0.245196320100808,0.0487725805040322)
.. controls (axis cs:-0.235551169543791,0.0972620268068218) and (axis cs:-0.211735637872312,0.141817752720962) .. (axis cs:-0.176776695296637,0.176776695296637)
.. controls (axis cs:-0.141817752720962,0.211735637872312) and (axis cs:-0.0972620268068217,0.235551169543791) .. (axis cs:-0.048772580504032,0.245196320100808)
.. controls (axis cs:-0.000283134201242374,0.254841470657824) and (axis cs:0.0499947984306559,0.249889526515012) .. (axis cs:0.0956708580912725,0.230969883127822)
.. controls (axis cs:0.141346917751889,0.212050239740632) and (axis cs:0.180400338143384,0.179999925916016) .. (axis cs:0.207867403075636,0.138892558254901)
.. controls (axis cs:0.235334468007888,0.0977851905937853) and (axis cs:0.25,0.0494394107167414) .. (axis cs:0.25,0)
--(axis cs:0.5,0)
--cycle;
\path [draw=black, fill=color1, postaction={pattern=vertical lines, opacity=0.5}]
(axis cs:-9.18485099360515e-17,-0.5)
.. controls (axis cs:0.0314391623835229,-0.5) and (axis cs:0.0628083668158298,-0.497034737205867) .. (axis cs:0.0936906552235738,-0.491143625759082)
.. controls (axis cs:0.124572943631318,-0.485252514312297) and (axis cs:0.154830878407255,-0.476461771838009) .. (axis cs:0.184062272425001,-0.46488824449511)
--(axis cs:0.0920311362125004,-0.232444122247555)
.. controls (axis cs:0.0774154392036277,-0.238230885919005) and (axis cs:0.062286471815659,-0.242626257156148) .. (axis cs:0.0468453276117869,-0.245571812879541)
.. controls (axis cs:0.0314041834079149,-0.248517368602934) and (axis cs:0.0157195811917615,-0.25) .. (axis cs:-4.59242549680257e-17,-0.25)
--(axis cs:-9.18485099360515e-17,-0.5)
--cycle;
\path [draw=black, fill=color2, postaction={pattern=vertical lines, opacity=0.5}]
(axis cs:0.184062272425001,-0.46488824449511)
.. controls (axis cs:0.213293666442746,-0.45331471715221) and (axis cs:0.241368429185782,-0.439009911418598) .. (axis cs:0.26791339215351,-0.422163966137329)
.. controls (axis cs:0.294458355121237,-0.40531802085606) and (axis cs:0.319355383409062,-0.386005906715206) .. (axis cs:0.342273546821757,-0.364484319478979)
--(axis cs:0.171136773410879,-0.182242159739489)
.. controls (axis cs:0.159677691704531,-0.193002953357603) and (axis cs:0.147229177560618,-0.20265901042803) .. (axis cs:0.133956696076755,-0.211081983068664)
.. controls (axis cs:0.120684214592891,-0.219504955709299) and (axis cs:0.106646833221373,-0.226657358576105) .. (axis cs:0.0920311362125004,-0.232444122247555)
--(axis cs:0.184062272425001,-0.46488824449511)
--cycle;
\path [draw=black, fill=color3, postaction={pattern=vertical lines, opacity=0.5}]
(axis cs:0.342273546821757,-0.364484319478979)
.. controls (axis cs:0.365191710234452,-0.342962732242752) and (axis cs:0.386029014467052,-0.319327450935515) .. (axis cs:0.404508490996326,-0.293892634667621)
.. controls (axis cs:0.422987967525599,-0.268457818399726) and (axis cs:0.439027375845151,-0.241336661512292) .. (axis cs:0.452413520851324,-0.212889657219199)
--(axis cs:0.226206760425662,-0.1064448286096)
.. controls (axis cs:0.219513687922575,-0.120668330756146) and (axis cs:0.2114939837628,-0.134228909199863) .. (axis cs:0.202254245498163,-0.14694631733381)
.. controls (axis cs:0.193014507233526,-0.159663725467758) and (axis cs:0.182595855117226,-0.171481366121376) .. (axis cs:0.171136773410879,-0.182242159739489)
--(axis cs:0.342273546821757,-0.364484319478979)
--cycle;
\path [draw=black, fill=white!38.8235294117647!black, postaction={pattern=vertical lines, opacity=0.5}]
(axis cs:0.452413520851324,-0.212889657219199)
.. controls (axis cs:0.460218299210788,-0.19630366002614) and (axis cs:0.467105083423969,-0.179301006432243) .. (axis cs:0.4730426753954,-0.16195872083578)
.. controls (axis cs:0.478980267366831,-0.144616435239316) and (axis cs:0.483959687496268,-0.126960744904255) .. (axis cs:0.487958378314248,-0.109071632576621)
--(axis cs:0.243979189157124,-0.0545358162883105)
.. controls (axis cs:0.241979843748134,-0.0634803724521276) and (axis cs:0.239490133683416,-0.0723082176196581) .. (axis cs:0.2365213376977,-0.0809793604178898)
.. controls (axis cs:0.233552541711985,-0.0896505032161216) and (axis cs:0.230109149605394,-0.0981518300130701) .. (axis cs:0.226206760425662,-0.1064448286096)
--(axis cs:0.452413520851324,-0.212889657219199)
--cycle;
\path [draw=black, fill=white!85.0980392156863!black, postaction={pattern=vertical lines, opacity=0.5}]
(axis cs:0.487958378314248,-0.109071632576621)
.. controls (axis cs:0.491957069132228,-0.0911825202489869) and (axis cs:0.494968983297469,-0.0730870401753824) .. (axis cs:0.496980476417646,-0.0548671674108434)
.. controls (axis cs:0.498991969537823,-0.0366472946463043) and (axis cs:0.499999999570942,-0.0183305836564646) .. (axis cs:0.5,-1.1703344638188e-08)
--(axis cs:0.25,-5.85167231909399e-09)
.. controls (axis cs:0.249999999785471,-0.00916529182823232) and (axis cs:0.249495984768911,-0.0183236473231522) .. (axis cs:0.248490238208823,-0.0274335837054217)
.. controls (axis cs:0.247484491648735,-0.0365435200876912) and (axis cs:0.245978534566114,-0.0455912601244935) .. (axis cs:0.243979189157124,-0.0545358162883105)
--(axis cs:0.487958378314248,-0.109071632576621)
--cycle;
\path [draw=black, fill=color0, postaction={pattern=north east lines, opacity=0.5}]
(axis cs:0.25,0)
.. controls (axis cs:0.25,0.0494394107167414) and (axis cs:0.235334468007888,0.0977851905937853) .. (axis cs:0.207867403075636,0.138892558254901)
.. controls (axis cs:0.180400338143384,0.179999925916016) and (axis cs:0.141346917751889,0.212050239740632) .. (axis cs:0.0956708580912725,0.230969883127822)
.. controls (axis cs:0.0499947984306559,0.249889526515012) and (axis cs:-0.000283134201242374,0.254841470657824) .. (axis cs:-0.048772580504032,0.245196320100808)
.. controls (axis cs:-0.0972620268068217,0.235551169543791) and (axis cs:-0.141817752720962,0.211735637872312) .. (axis cs:-0.176776695296637,0.176776695296637)
.. controls (axis cs:-0.211735637872312,0.141817752720962) and (axis cs:-0.235551169543791,0.0972620268068218) .. (axis cs:-0.245196320100808,0.0487725805040322)
.. controls (axis cs:-0.254841470657824,0.000283134201242478) and (axis cs:-0.249889526515012,-0.0499947984306559) .. (axis cs:-0.230969883127822,-0.0956708580912724)
.. controls (axis cs:-0.212050239740632,-0.141346917751889) and (axis cs:-0.179999925916016,-0.180400338143384) .. (axis cs:-0.138892558254901,-0.207867403075636)
.. controls (axis cs:-0.0977851905937853,-0.235334468007888) and (axis cs:-0.0494394107167414,-0.25) .. (axis cs:-4.59242549680257e-17,-0.25)
--(axis cs:0,0)
.. controls (axis cs:0,0) and (axis cs:0,0) .. (axis cs:0,0)
.. controls (axis cs:0,0) and (axis cs:0,0) .. (axis cs:0,0)
.. controls (axis cs:0,0) and (axis cs:0,0) .. (axis cs:0,0)
.. controls (axis cs:0,0) and (axis cs:0,0) .. (axis cs:0,0)
.. controls (axis cs:0,0) and (axis cs:0,0) .. (axis cs:0,0)
.. controls (axis cs:0,0) and (axis cs:0,0) .. (axis cs:0,0)
.. controls (axis cs:0,0) and (axis cs:0,0) .. (axis cs:0,0)
.. controls (axis cs:0,0) and (axis cs:0,0) .. (axis cs:0,0)
--(axis cs:0.25,0)
--cycle;
\path [draw=black, fill=color1, postaction={pattern=north east lines, opacity=0.5}]
(axis cs:-4.59242549680257e-17,-0.25)
.. controls (axis cs:0.0157195811917615,-0.25) and (axis cs:0.0314041834079149,-0.248517368602934) .. (axis cs:0.0468453276117869,-0.245571812879541)
.. controls (axis cs:0.062286471815659,-0.242626257156148) and (axis cs:0.0774154392036277,-0.238230885919005) .. (axis cs:0.0920311362125004,-0.232444122247555)
--(axis cs:0,0)
.. controls (axis cs:0,0) and (axis cs:0,0) .. (axis cs:0,0)
.. controls (axis cs:0,0) and (axis cs:0,0) .. (axis cs:0,0)
--(axis cs:-4.59242549680257e-17,-0.25)
--cycle;
\path [draw=black, fill=color2, postaction={pattern=north east lines, opacity=0.5}]
(axis cs:0.0920311362125004,-0.232444122247555)
.. controls (axis cs:0.106646833221373,-0.226657358576105) and (axis cs:0.120684214592891,-0.219504955709299) .. (axis cs:0.133956696076755,-0.211081983068664)
.. controls (axis cs:0.147229177560618,-0.20265901042803) and (axis cs:0.159677691704531,-0.193002953357603) .. (axis cs:0.171136773410879,-0.182242159739489)
--(axis cs:0,0)
.. controls (axis cs:0,0) and (axis cs:0,0) .. (axis cs:0,0)
.. controls (axis cs:0,0) and (axis cs:0,0) .. (axis cs:0,0)
--(axis cs:0.0920311362125004,-0.232444122247555)
--cycle;
\path [draw=black, fill=color3, postaction={pattern=north east lines, opacity=0.5}]
(axis cs:0.171136773410879,-0.182242159739489)
.. controls (axis cs:0.182595855117226,-0.171481366121376) and (axis cs:0.193014507233526,-0.159663725467758) .. (axis cs:0.202254245498163,-0.14694631733381)
.. controls (axis cs:0.2114939837628,-0.134228909199863) and (axis cs:0.219513687922575,-0.120668330756146) .. (axis cs:0.226206760425662,-0.1064448286096)
--(axis cs:0,0)
.. controls (axis cs:0,0) and (axis cs:0,0) .. (axis cs:0,0)
.. controls (axis cs:0,0) and (axis cs:0,0) .. (axis cs:0,0)
--(axis cs:0.171136773410879,-0.182242159739489)
--cycle;
\path [draw=black, fill=white!38.8235294117647!black, postaction={pattern=north east lines, opacity=0.5}]
(axis cs:0.226206760425662,-0.1064448286096)
.. controls (axis cs:0.230109149605394,-0.0981518300130701) and (axis cs:0.233552541711985,-0.0896505032161216) .. (axis cs:0.2365213376977,-0.0809793604178898)
.. controls (axis cs:0.239490133683416,-0.0723082176196581) and (axis cs:0.241979843748134,-0.0634803724521276) .. (axis cs:0.243979189157124,-0.0545358162883105)
--(axis cs:0,0)
.. controls (axis cs:0,0) and (axis cs:0,0) .. (axis cs:0,0)
.. controls (axis cs:0,0) and (axis cs:0,0) .. (axis cs:0,0)
--(axis cs:0.226206760425662,-0.1064448286096)
--cycle;
\path [draw=black, fill=white!85.0980392156863!black, postaction={pattern=north east lines, opacity=0.5}]
(axis cs:0.243979189157124,-0.0545358162883105)
.. controls (axis cs:0.245978534566114,-0.0455912601244935) and (axis cs:0.247484491648735,-0.0365435200876912) .. (axis cs:0.248490238208823,-0.0274335837054217)
.. controls (axis cs:0.249495984768911,-0.0183236473231522) and (axis cs:0.249999999785471,-0.00916529182823232) .. (axis cs:0.25,-5.85167231909399e-09)
--(axis cs:0,0)
.. controls (axis cs:0,0) and (axis cs:0,0) .. (axis cs:0,0)
.. controls (axis cs:0,0) and (axis cs:0,0) .. (axis cs:0,0)
--(axis cs:0.243979189157124,-0.0545358162883105)
--cycle;
\end{axis}

\draw (\fwidth/2,\fheight/2) -- (\fwidth/2,\fheight);	
\node at (\fwidth/2+5.,\fheight+4.) {\scriptsize 100};
\node at (\fwidth/2+5.,3\fheight/4+4.) {\scriptsize 50};
\node at (\fwidth/2+5.,5\fheight/8+4.) {\scriptsize 25};

\end{tikzpicture}